\documentclass{article}
\usepackage{iclr_like,times}

\usepackage{hyperref}
\usepackage{url}

\usepackage[utf8]{inputenc}
\usepackage[T1]{fontenc}
\usepackage{hyperref}
\usepackage{url}
\usepackage{booktabs}
\usepackage{amsfonts}
\usepackage{nicefrac}
\usepackage{microtype}
\usepackage{xcolor}

\usepackage{amsmath}
\usepackage{amssymb}
\usepackage{bm}
\usepackage{url}
\usepackage{stmaryrd}
\usepackage{mathtools}
\usepackage{calc}
\usepackage{booktabs}
\usepackage{nccmath}

\usepackage{algorithm}
\usepackage{algorithmic}

\usepackage{math_commands}
\usepackage{multirow}
\usepackage{mathtools}
\usepackage{enumerate}
\usepackage{amsthm}
\usepackage{wrapfig}
\usepackage{colortbl}
\usepackage[safe]{tipa}
\newtheorem{thm}{Theorem}

\newtheorem{conj}{Conjecture}

\newtheorem{dfn}{Definition}
\newtheorem{alg}{Algorithm}

\usepackage[subrefformat=parens]{subcaption}

\def\fst{\bar{\rvf}_{\sharp}}
\def\fdet{\bar{\rvf}_{\flat}}
\def\lst{{L}_{\sharp}}
\def\gst{{G}_{\sharp}}
\def\ldet{{L}_{\flat}}
\newcommand{\diff}[1]{\partial_{#1}}
\renewcommand{\diff}[2][]{\frac{\partial^{#1}}{\partial {#2}^{#1}}}
\renewcommand{\diff}[2][]{\partial_{#2}^{#1}}
\def\scorefn{\mathbf{S}_{\theta}}

\newcommand{\myparagraph}[1]{$\blacksquare$ \noindent\textit{#1}:~}

\usepackage{cancel}

\definecolor{bluekeywords}{rgb}{0.13,0.13,1}
\definecolor{greencomments}{rgb}{0,0.5,0}
\definecolor{redstrings}{rgb}{0.9,0,0}
\usepackage{listings}
\title{Quasi-Taylor Samplers for Diffusion Generative Models based on Ideal Derivatives}

\author{Hideyuki Tachibana, Mocho Go, Muneyoshi Inahara, Yotaro Katayama \& Yotaro Watanabe \thanks{
   HT is also with Asia University, Musashino City, Tokyo, Japan.
   } \\
PKSHA Technology, Inc., Bunkyo, Tokyo, Japan\\
\texttt{\{h\_tachibana, m\_go, m\_inahara, y\_katayama, y\_watanabe\}@pkshatech.com}
}

\begin{document}

\maketitle

\begin{abstract}
   Diffusion generative models have emerged as a
   new challenger to popular deep neural generative models such as GANs,
   but have the drawback that they often require a huge number of neural function evaluations (NFEs)
   during synthesis unless some sophisticated sampling strategies are employed.
   This paper proposes new efficient samplers based on the numerical schemes
   derived by the familiar Taylor expansion, which directly solves the ODE/SDE of interest.
   In general, it is not easy to compute the derivatives
   that are required in higher-order Taylor schemes, but in the case of diffusion models,
   this difficulty is alleviated by the trick that the authors call ``ideal derivative substitution,''
   in which the higher-order derivatives are replaced by
   tractable ones.
   To derive ideal derivatives, the authors argue the ``single point approximation,''
   in which the true score function is approximated by a conditional one,
   holds in many cases,
   and considered the derivatives of this approximation.
   Applying thus obtained new quasi-Taylor samplers to image generation tasks,
   the authors experimentally confirmed that the proposed samplers could synthesize plausible images
   in small number of NFEs,
   and that the performance was better or at the same level as DDIM and Runge-Kutta methods.
   The paper also argues the relevance of the proposed samplers to the existing ones mentioned above.
\end{abstract}

\def\mystyle{\displaystyle}
\section{Introduction}\label{label:000}
    Generative modeling based on deep neural networks is an important research
    subject for both fundamental and applied purposes,
    and has been a major trend in machine learning studies for several years.
    To date, various types of neural generative models have been studied including
    GANs~\citep{goodfellow2014generative}, VAEs~\citep{kingma2021variational,kingma2019introduction},
    normalizing flows~\citep{rezende2015variational},
    and autoregressive models~\citep{van2016pixel,oord2016wavenet}.
    In addition to these popular models,
    a class of novel generative models
    based on the idea of iteratively refinement using the diffusion process
    has been rapidly gaining attention recently as a challenger that rivals
    the classics above~\citep{sohl2015deep,song2019generative,song2020score,song2020improved,ho2020denoising,dhariwal2021diffusion}.
    The diffusion-based generative models
    have recently been showing impressive results in many fields including
    image~\citep{ho2020denoising,vahdat2021score,saharia2021image,ho2021cascaded,sasaki2021unit},
    video~\citep{ho2022video},
    text-to-image~\citep{nichol2021glide,ramesh2022hierarchical},
    speech~\citep{chen2020wavegrad,chen2021wavegrad,kong2020diffwave,popov2021grad,kameoka2020voicegrad},
    symbolic music~\citep{mittal2021symbolic},
    natural language~\citep{hoogeboom2021argmax,austin2021structured},
    chemoinformatics~\citep{xu2022geodiff}, etc.

    However, while the diffusion models have good synthesis quality,
    it has been said that they have a fatal drawback that they often require
    a very large number of iterations (refinement steps) during synthesis,
    ranging from hundreds to a thousand.
    In particular, the increase in refinement steps critically
    reduces the synthesis speed, as each step involves at least one neural function evaluation (NFE).
    Therefore, it has been a common research question
    how to establish a systematic method to stably generate good data from diffusion models in a relatively
    small number of refinement steps, or NFEs in particular.
    From this motivation, there have already been some studies aiming
    at reducing the NFEs (See \secref{label:001}).
    Among these, Probability Flow ODE (PF-ODE)~\citep{song2020score}
    enable efficient and deterministic sampling, and is gaining attention.
    This framework has the merit of deriving a simple ODE by a straightforward conceptual manipulation
    of diffusion process.
    However, the ODE is eventually solved by using a black-box Runge-Kutta solver in the original paper,
    which requires several NFEs per step and is clearly costly.
    Another PF-ODE solver includes DDIM~\citep{song2020denoising}, and is also commonly used.
    It is certainly efficient and can generate plausible images.
    However, it was not originally formulated as a PF-ODE solver,
    and the relationship between DDIM and PF-ODE is not straightforward.

    From these motivations, we provide another sampler to solve the same ODE,
    which performs better than or on par with DDIM.
    The derivation outline is simple and intuitive:
    (1) consider the Taylor expansion of the given system,
    and (2) replace the derivatives in the Taylor series with appropriate functions;
    that's all.

    The contribution of this paper would be as follows:
    (1)
        We propose novel samplers for diffusion models based on Taylor expansion of PF-ODE.
        They outperformed, or were on par with DDIM,
        while its derivation would be more elementary.
    (2)
        We also show that the same technique is applicable to derive a stochastic solver
        for reverse-time SDE argued in~\citep{song2020score},
        which we call R-SDE in this paper.
    (3)
        To derive these algorithms,
        we show that
        the derivatives of score function
        can be approximated by simple functions.
        We call this technique the \textit{ideal derivative substitution}.
    (4)
        This idea is also used to derive a Lipschitz-continuity-aware noise schedule.
    (5)
        The fact that proposed methods work well could be an evidence that the assumptions
        behind our methods are largely valid.
        This will promote further understanding of the fundamentals of diffusion generative models.

\section{Background and Related Work}\label{label:001}

    \paragraph{Diffusion Process to draw a new data from a target density:}
        Let us first briefly summarize the framework of the diffusion-based generative models.
        Following~\citet{song2020score}, we describe the mechanisms using the language
        of continuous-time diffusion process for later convenience.
        Let us consider ``particles'' $\{\rvx_t\}$ moving in a $d$-dim space
        obeying the following It\^o diffusion,
        \begin{align}
            \text{SDE:}\quad &
                d\rvx_t = \rvf(\rvx_t, t) dt + g(\rvx_t, t) d\mathbf{B}_t,
            \label{label:002}
        \end{align}
        where $\mathbf{B}_t$ is the $d$-dim Brownian motion whose temporal increments obeys the standard Gaussian.
        The drift $\rvf(\cdot, \cdot)$ is $d$-dim vector, and the diffusion coefficient $g(\cdot, \cdot)$
        is scalar.
        The SDE describes the microscopic dynamics of each particle.
        On the other hand,
        the ``population'' of the particles obeying the above SDE,
        i.e.\ density function $p(\rvx_t, t \mid \rvx_s, s), (t > s)$,
        follows the following PDEs,
        which are known as Kolmogorov's forward and backward equations;
        the former is also known as the Fokker-Planck equation
        (see \secref{label:065}),
        \begin{align}
            &\text{FPE:}&
                \diff{t} p(\rvx_t, t \mid \rvx_s, s)
                &=
                - \nabla_{\rvx_t} \cdot \rvf(\rvx_t, t)  p(\rvx_t, t \mid \rvx_s, s)
                + \Delta_{\rvx_t} \frac{g(\rvx_t, t)^2}{2} p(\rvx_t, t \mid \rvx_s, s) ,
                \label{label:003}\\
            &\text{KBE:}&
                - \diff{s} p(\rvx_t, t \mid \rvx_s, s)
                &=
                \rvf(\rvx_t, t) \cdot \nabla_{\rvx_s} p(\rvx_t, t \mid \rvx_s, s)
                + \frac{g(\rvx_t, t)^2}{2} \Delta_{\rvx_s} p(\rvx_t, t \mid \rvx_s, s),
                \label{label:004}
        \end{align}
        where $\Delta_{\rvx} \coloneqq \nabla_{\rvx}^2$ is Laplacian.
        (FPE also holds for $p(\rvx_t, t)$; consider the expectation $\mathbb{E}_{p(\rvx_s, s)}[\cdot]$.)
        These PDEs enables us to understand the macroscopic behavior of the particle ensemble.
        For example, if $\rvf(\rvx, t) = - \nabla U(\rvx), g(\rvx, t) = \sqrt{2 D}$,
        where $U(\rvx)$ a certain potential and $D$ a constant,
        then we may verify that the stationary solution of FPE is
        $p(\rvx) \propto e^{-U(\rvx) / D}$.
        It means that we may draw a sample $\rvx$ that follows the stationary density
        by evolving the SDE over time.
        This technique is often referred to as the Langevin Monte Carlo method~\citep{rossky1978brownian,roberts1996exponential}.
        Some of the diffusion generative models are based on this framework,
        e.g.~\citep{song2019generative,song2020improved},
        in which the potential gradient $\nabla U(\rvx)$ is approximated by a neural network.

        Another systematic approach is considering the reverse-time dynamics~\citep{song2020score} based on KBE.
        Roughly speaking, FPE gives information about the future from the initial density,
        while KBE gives information about what the past states were likely to be from the terminal density.
        Here, instead of using KBE directly, it is useful to consider a variant of it which is
        transformed into the form of FPE, because it has an associated SDE that enables
        the particle-wise backward sampling~\citep{stratonovich1965conditional,anderson1982reverse}.
        \begin{align}
            &\text{\!\!\!R-FPE:\!\!}&
                - \diff{s} p(\rvx_s, s \mid \rvx_t, t)
                &=
                \nabla_{\rvx_s} \cdot \bar{\rvf}(\rvx_s, s) p(\rvx_s, s \mid \rvx_t, t)
                + \Delta_{\rvx_s} \frac{\bar{g}(\rvx_s, s)^2}{2} p(\rvx_s, s \mid \rvx_t, t)
                \label{label:005} \\
            &\text{\!\!\!R-SDE:\!\!}&
                d\rvx_s &= -\bar{\rvf}(\rvx_s, s) (-ds) + \bar{g}(\rvx_s, s) d\bar{\mathbf{B}}_s. \label{label:006}
        \end{align}
        Hereafter, let $g(x_t, t) = g(t)$ for simplicity.
        Then the specific forms of drift and diffusion coefficients are written as follows,
        \begin{equation}
            \bar{\rvf}(\rvx_t, t) = \fst (\rvx_t, t) \coloneqq \rvf(\rvx_t, t) - g(t)^2 \nabla_{\rvx_t} \log p(\rvx_t, t),
            \quad \bar{g}(t) = g(t).
            \label{label:007}
        \end{equation}
        Starting from a certain random variable $\rvx_T$,
        then by evolving the R-SDE reverse in time, we may obtain a
        $\hat{\rvx}_0$ which follows $p(\rvx_0, 0 \mid \rvx_T, T)$ (i.e.\ the solution of R-FPE \eqref{label:005}).
        Therefore, if the initial density $p(\rvx_0, 0)$ of the forward dynamics \eqref{label:003} is the true density,
        then we may utilize this mechanism as a generative model to draw a new sample $\hat{\rvx}_0$ from it.

        Exactly the same initial density is achieved by evolving a deterministic system
        where the diffusion coefficient is exactly zero, i.e.\ $\bar{g}(t)=0$ in \eqref{label:006}.
        This is called the Probability Flow ODE (PF-ODE)~\citep{song2020score},
        which is an example of neural ODEs~\citep{chen2018neural}.
        The PF-ODE is derived by formally eliminating the diffusion term of FPE \eqref{label:003}.
        The coefficients of PF-ODE are written as follows. See also \secref{label:069}.
        \begin{equation}
            \bar{\rvf}(\rvx_t, t) = \fdet (\rvx_t, t) \coloneqq \rvf(\rvx_t, t) - \frac{1}{2}g(t)^2 \nabla_{\rvx_t} \log p(\rvx_t, t).
            \quad \bar{g}(t) = 0.
            \label{label:008}
        \end{equation}

        Some extensions of this framework include as follows.
        \citet{dockhorn2021score} introduced the velocity variable
        considering the Hamiltonian dynamics.
        Another extension is the introduction of a conditioning parameter,
        and guidance techniques using it~\citep{dhariwal2021diffusion,ho2021classifier,choi2021ilvr}
        to promote the dynamics to go to a specific class of images,
        which has achieved remarkable results in text-to-image tasks~\citep{nichol2021glide,ramesh2022hierarchical}.

    \paragraph{Variance-Preserving Model (VP-SDE Model):}
        The solution of unconditioned FPE is written as the
        convolution with the initial density $p(\rvx_0, 0)$
        and the \textit{fundamental solution}, or the \textit{heat kernel}, $p(\rvx_t, t \mid \rvx_0, 0)$,
        which is the solution of the conditional FPE
        under the assumption that the initial density was delta function, $p(\rvx_0, 0) = \delta(\rvx_0 - \rvx_0^*)$.
        Although it is still intractable to solve this problem in general,
        a well-known exception is the (time-dependent) Ornstein-Uhlenbeck process
        where $\rvf(\rvx_t, t) = -\frac{1}{2}\beta_t \rvx_t $ and $g(\rvx_t, t) = \sqrt{\beta_t}$;
        $\beta_t = \beta(t)$ is a non-negative continuous function.
        In this case, the heat kernel is simply written as follows,
        \begin{equation}
            p(\rvx_t, t \mid \rvx_0, 0) = \mathcal{N}(\rvx_t \mid \sqrt{1 - \nu_t} \rvx_0, \nu_t\mathbf{I}),
            \quad \text{~where~~}
            \nu_t = 1 - e^{ - \int_0^t \beta_{t'} dt' }.
            \label{label:009}
        \end{equation}
        This model is referred to as the variance-preserving (VP) model by~\citet{song2020score}.
        This model has good properties such as
        the scale of data $\|\rvx_t\|_2$ is almost homogeneous,
        which is advantageous in neural models.
        Although \citet{song2020score} also considered variance-exploding (VE) models,
        we only consider VP-type models in this paper.

    \paragraph{Training Objective:} \label{label:010}
        In diffusion-based generative models,
        one estimates the \textit{score function}
        $\nabla_{\rvx_t} \log p(\rvx_t, t)= \nabla_{\rvx_t} \log \mathbb{E}_{p(\rvx_0,0)}[p (\rvx_t, t \mid \rvx_0,0)]$
        by a neural network $\scorefn(\rvx_t, t)$.
        This sort of learning has been referred to as the
        score matching~\citep{hyvarinen2005estimation,vincent2011connection}.
        However, the exact evaluation of this training target is clearly intractable because of the expectation
        $\mathbb{E}_{p(\rvx_0,0)}[\cdot]$, so it has been common to consider a
        Variational Bayesian surrogate loss;
        \citet{ho2021classifier} showed that the following loss function approximates the negative ELBO,
        \begin{align}
        \textstyle
        \mathcal{L} & \coloneqq
        \E    [
                \lVert
                    - \sqrt{\nu_t}\nabla_{\rvx_t} \log p(\rvx_t, t \mid \rvx_0, 0)
                    - \scorefn(\rvx_t, t)
                \rVert^2_2
            ]
        =
        \textstyle
        \E[
            \lVert \frac{\rvx_t - \sqrt{1 - \nu_t} \rvx_0}{\sqrt{\nu_t}} - \scorefn(\rvx_t, t)\rVert_2^2
        ]
        \\
        &=
        \E    [
                \lVert
                    \rvw - \scorefn
                    (
                    {\textstyle \sqrt{\vphantom{1}1 - \nu_t} \rvx_0 + \sqrt{\nu_t}\rvw},
                    t
                    )
                \rVert^2_2
            ],
            \label{label:011}
        \end{align}
        where the expectation in \eqref{label:011} is taken w.r.t.\
        $\rvx_0     \sim \mathcal{D}$,
        $\rvw \sim \gN(\bm{0}, \mathbf{I})$,
        and $t \sim \text{Uniform}([0, T]).$
        Some variants of the score matching objectives are also studied.
        For example, \citet{chen2020wavegrad} reported that the $L_1$ loss
        gave better results than the $L_2$ loss in speech synthesis.
        Also, \citet{kingma2021variational} argued that the weighted loss with SNR-based weights improves the performance.

        It should be noted that the above loss function will actually be very close
        to the ideal score matching loss function in practice,
        where the probability is not conditioned on $\rvx_0$,
        i.e.,\
        \begin{equation}
            \mathcal{L}_\text{ideal}
            =\E    [
            \lVert
                - \sqrt{\nu_t}\nabla_{\rvx_t} \log p(\rvx_t, t)
                - \scorefn(\rvx_t, t)
            \rVert^2_2
            ].
        \end{equation}
        This is because
        there almost always exists a point $\rvx_0$ on the data manifold
        such that
        $\nabla_{\rvx_t} \log p(\rvx_t, t) \approx \nabla_{\rvx_t} \log p(\rvx_t, t \mid \rvx_0, 0)$
        holds with very high accuracy
        in very high-dim cases,
        because of the well-known ``\mbox{log-sum-exp} $\approx$ max'' law.
        For more details, see \secref{label:020} and \secref{label:039}.

    \paragraph{Sampling Schemes for R-SDE and PF-ODE:}\label{label:012}
        Thus obtained $\scorefn (\rvx_t, t)$ is expected to finely approximate
        $-\sqrt{\nu_t}\nabla_{\rvx_t} \log p(\rvx_t, t)$,
        and we may use it in~\eqref{label:006}.
        One of the simplest numerical schemes for solving SDEs is the Euler-Maruyama method~\citep[Theorem.~1]{maruyama1955continuous}
        as follows, and many diffusion generative models are actually using it.
        \begin{equation}
            \text{Euler-Maruyama: } \quad
            \rvx_{t - h} \gets
            \rvx_t - h \fst (\rvx_t, t) +  \sqrt{h} g(t) \rvw, \quad \text{where~~}
            \rvw \sim \mathcal{N}(\bm{0}, \mathbf{I})
            \label{label:013}
        \end{equation}
        where $h > 0$ is the step size.
        The error of the Euler-Maruyama method is the order of $O(\sqrt{h})$ in general,
        though it is actually $O(h)$ in our case; this is because $\nabla_{\rvx_t} g(t) = 0$.
        As a better solver for the R-SDE, the Predictor-Corrector (PC)-based sampler was proposed in~\citep{song2020score}.
        The PC sampler outperformed the Predictor-only strategy,
        but it requires many NFEs in the correction process, so we will exclude it in our discussion.
        Another R-SDE solver is the one proposed by \citet{jolicoeur2021gotta},
        whose NFE per refinement step is 2.

        On the other hand, there are also deterministic samplers for PF-ODE~\eqrefs{label:006}{label:008} as follows,
        \begin{align}
            \text{Euler:} \quad& \rvx_{t - h} \gets \rvx_t - h \fdet (\rvx_t, t)
            \label{label:014} \\
            \text{Runge-Kutta:} \quad&
                \textstyle
                \rvx_{t -  h} \gets \rvx_t - h \sum_{i=1}^m b_i \rvk_i,
                \text{~~where~~} \rvk_i = \fdet (\rvx_t - h \sum_{j=1}^{i-1}a_{ij} \rvk_j, t - h c_i)
        \end{align}
        where $\{a_{ij}\}, \{b_i\}, \{c_i\}$ are coefficients of the Runge-Kutta (RK) method (see \secref{label:071}).
        The error of the Euler method is $o(h)$,
        and that of the RK method is $o(h^p), p \le m$ in general~\citep[\S~16]{press2007numerical}.
        Another deterministic sampler is DDIM~\citep[Eq.~(13)]{song2020denoising},
        and is also understood as a PF-ODE solver~\citep{salimans2022progressive}.
        Its NFE per step is only 1, and is capable of efficiently generate samples.
        \begin{equation}
            \text{DDIM:} \quad \rvx_{t - h} \gets
                \sqrt{\frac{1 - \nu_{t - h}}{1 - \nu_t}} \rvx_{t}
                + \frac{\sqrt{(1 - \nu_t)\nu_{t-h}} - \sqrt{(1 - \nu_{t-h})\nu_t}  }{\sqrt{1 - \nu_t}} \scorefn(\rvx_t, t).
            \label{label:015}
        \end{equation}

        Other techniques that aimed to make sampling faster include as follows.
        \citet{song2020improved} proposed some techniques to accelerate the sampling.
        \citet{watson2021learning} proposed a DP-based optimization method to tune noise schedules for faster sampling.
        \citet{salimans2022progressive} proposed distilling
        the pretrained teacher model to a student model that can predict teacher's several steps in a single step,
        which is efficient during the sampling but the distillation-based training is costly.

\section{Proposed Method: Quasi-Taylor Samplers}\label{label:016}
    \subsection{Motivation: Higher-order Straightforward Solvers for R-SDE and PF-ODE}
        As mentioned above, DDIM already exists as an efficient sampler of diffusion generative models,
        but it is not necessarily derived directly from PF-ODEs, and its relationship to PF-ODEs
        was revealed through a little argumentation~\citep{song2020denoising,salimans2022progressive}.
        Since PF-ODE gives a basis for the diffusion generative models,
        developing a sampler that directly solves it through intuitive and straightforward argumentation
        would be beneficial in various contexts in the future development of this field.
        Of course, the RK method mentioned directly solves the PF-ODE,
        but it requires several NFEs in a single refinement step,
        and it seems to be behind in competitiveness despite its high convergence rate,
        since the state-of-the-art samplers e.g.~\citep{salimans2022progressive}
        are already generating data in single digit NFEs.

        From these motivations, we propose a simple sampler based on the Taylor expansion,
        a very basic technique that is familiar to many researchers and practitioners.
        In general, Taylor methods are not very popular as numerical schemes because they
        require
        higher-order derivatives, which are not always tractable.
        Nevertheless,
        in diffusion models, the derivatives are expected to have good structure,
        and are effectively evaluated.
        This section describes the details of the idea, and derives
        solvers for both PF-ODE and R-SDE.
        We summarize the entire procedures as pseudocodes in \secref{label:074}.

    \subsection{Taylor Scheme for ODE and It\^o-Taylor Scheme for SDE}
        For simplicity, we consider the 1-dim case in this section.
        It can be easily generalized to multidimensional cases simply by parallelizing each dimension,
        since the Jacobian matrix is diagonal assuming that each dimension is independent of each other.

        \paragraph{Taylor Scheme for Deterministic Systems}
        Given a ODE $\dot{x}_t = a(x_t, t)$,
        we can consider the Taylor expansion of it.
        Using a differential operator
        $\ldet \coloneqq \left(\diff{t} + a(t, x_t) \diff{x_t} \right)$,
        we can write the Taylor expansion of the path $x_t$ as follows.
        If we ignore $o(h^p)$ terms of the series, we may obtain a numerical scheme of order $p$.
        See also \secref{label:058}.
        \begin{equation}
            x_{t + h} = x_t + ha(x_t, t) + \frac{h^2}{2!}\ldet a(x_t,t) + \frac{h^3}{3!}\ldet^2 a(x_t,t) + \cdots.
        \end{equation}

        \paragraph{It\^o-Taylor Scheme for Stochastic Systems}
        In stochastic systems, the Taylor expansion requires modifications.
        If $x_t$ obeys a stochastic system
        $dx_t = a(x_t, t)dt + b(x_t, t) dB_t$,
        then the path is written in a stochastic version of Taylor-like series,
        which is often called the It{\^o}-Taylor expansion,
        a.k.a.\ Wagner-Platen expansion~\citep{wagner1982taylor};\citep[\S~2.3.B]{kloeden1994numerical};\citep[\S~8.2]{sarkka2019applied}.
        The It\^o-Taylor expansion is based on the following differential operators
        $\lst, \gst$, which are based on It{\^o}'s formula~\citep{ito1944}.
        \begin{equation}
            \lst \coloneqq \diff{t} +
                 a(x, t) \diff{x}  +
                 \frac{1}{2} b(x, t)^2 \diff[2]{x} ,\quad
            \gst \coloneqq b(x, t) \diff{x}
        \label{label:017}
        \end{equation}
        In \citep{kloeden1992stochastic}, a number of higher order numerical schemes for SDEs
        based on the It{\^o}-Taylor expansion are presented.
        One of the simplest of them is as follows. See also \secref{label:061}.
        \begin{thm}[{\citet[\S~14.2]{kloeden1992stochastic}: An It{\^o}-Taylor scheme of weak order $\beta = 2$}]\label{label:018}
            Let $x_t$ obeys the above SDE,
            and let the differential operators $\lst, \gst$ be given by \eqref{label:017}.
            Then, the following numerical scheme weakly converges with the order of $\beta=2$.
            \begin{equation}
                x_{t + h} \gets x_t + ha + \tilde{w}_tb
                + \frac{\tilde{w}_t^2 - h}{2} \gst b + \frac{h^2}{2}\lst a
                + (\tilde{w}_th - \tilde{z}_t)\lst b + \tilde{z}_t \gst a
                \label{label:019}
            \end{equation}
            where $\tilde{w}_t = \sqrt{h}{w}_t, \tilde{z}_t = h\sqrt{h} {z}_t$
            are correlated Gaussian random variables,
            and $w_t, z_t$ are given by
            $w_t = u_1$ and $z_t = \frac{1}{2}u_1 + \frac{1}{2\sqrt{3}} u_2$,
            where $u_1, u_2 \sim \gN(0, 1)$ \textup{(}i.i.d.\textup{)}.
            The notations $a, \lst a$, etc.\ are the abbreviations for $a(x_t, t), (\lst a)(x_t, t)$, etc.
        \end{thm}
        This scheme has the order $\beta = 2$ of weak convergence (see \secref{label:070}).
        In addition, in a special case where $\gst^2 b \equiv 0$, the
        strong $\gamma = 1.5$ convergence is also guaranteed~\citep[\S~10.4]{kloeden1992stochastic}.

    \subsection{Single Point Approximation of the Score Function}\label{label:020}
        Before proceeding, let us introduce the single point approximation of score function
        that $\nabla_{\rvx_t}\log p(\rvx_t, t)$ almost certainly has a some point $\rvx_0$ on the data manifold
        such that the following approximation holds,
        \begin{equation}
            \nabla_{\rvx_t}\log p(\rvx_t, t) = \nabla_{\rvx_t}\log \int p(\rvx_t, t \mid \rvx_0,0)p(\rvx_0,0)d\rvx_0
            \approx \nabla_{\rvx_t}\log p(\rvx_t, t \mid \rvx_0, 0).
            \label{label:021}
        \end{equation}
        To date, this approximation has often been understood as ``a tractable surrogate.''
        However, the error between the integral and the single point approximation is actually very small
        in practical scenarios.
        More specifically, the following facts can be shown
        under some assumptions.
        \begin{enumerate}
            \item The relative $L_2$ distance between $\nabla_{\rvx_t}\log p(\rvx_t, t)$
            and $\nabla_{\rvx_t}\log p(\rvx_t, t \mid \rvx_0, 0)$ is bounded above by $\sqrt{{(1 - \nu_t)}/{\nu_t}}$
            for any point $\rvx_0$ on the ``data manifold'' in practical scenarios.
            \item When the noise level is low $\nu_t\approx 0$, and the data space is sufficiently high-dimensional,
            the distant points far from $\rvx_t$ do not contribute to the integral.
            If the data manifold, which is sufficiently flat locally, is locally a $k$-dim subspace of the entire $d$-dim data space where $1 \ll k \ll d$,
            then the relative $L_2$ distance is bounded above by $2\sqrt{k/d}$.
        \end{enumerate}
        Of course, the single point approximation is not always valid.
        In fact, the approximation tends to break down when the noise level $\nu_t$ is around 0.9
        (SNR $=(1-\nu_t)/\nu_t$ is around $0.1$).
        In this region, the single point approximation
        can deviates from the true gradient by about 20\%
        in some cases.
        Conversely, however,
        it would be also said that
        the error is as small as this level even in the worst empirical cases.
        For more details on this approximation, see \secref{label:039}.

    \subsection{Taylor and It\^o-Taylor Schemes with Ideal Derivatives}\label{label:022}
        Now let us adopt the above Taylor schemes to our problem setting where the base SDE is \eqref{label:006},
        and $\fst$ and $\fdet$ are given by \eqrefs{label:007}{label:008}.
        Under these conditions, the operators are written as follows;
        note that the time evolves backward in time in our case,
        the temporal derivative should be $-\diff{t}$,
        \begin{gather}
            \ldet =
                - \diff{t}
                - \left(\fdet(\rvx_t, t) \cdot \nabla_{\rvx_t}\right), \quad
            \lst =
                - \diff{t}
                - \left(\fst(\rvx_t, t) \cdot \nabla_{\rvx_t}\right)
                + \frac{\beta_t}{2} \Delta_{\rvx_t},
            \quad
            \gst = \sqrt{\beta_t}\left(\bm{1} \cdot \nabla_{\rvx_t}\right), \nonumber\\
            \text{where~~}
            \fdet(\rvx_t, t) = -\frac{\beta_t}{2}\rvx_t + \frac{\beta_t}{2\sqrt{\nu_t}} \scorefn(\rvx_t, t),\quad
            \fst(\rvx_t, t)  = -\frac{\beta_t}{2}\rvx_t + \frac{\beta_t}{\sqrt{\nu_t}} \scorefn(\rvx_t, t).
        \end{gather}
        It is not easy in general to evaluate expressions involving such many derivatives.
        Indeed, for example, $\ldet(-\fdet)$ has the derivatives of the learned score function,
        viz.\ $\diff{t} \scorefn(\rvx_t, t)$ and $(\bullet \cdot \nabla_{\rvx_t})\scorefn(\rvx_t, t)$,
        which are intractable to evaluate exactly.
        Fortunately, however,
        by using the trick which the authors call the ``\textit{ideal derivative substitution}",
        we may write all of the derivatives above explicitly.
        Since the score function has a single point approximation \eqref{label:021} in many cases,
        we may assume that the derivatives should ideally hold following equalities.
        For derivation, see \secref{label:050}.
        \begin{conj}[Ideal Derivatives]
            Let $\rva$ be an arbitrary vector.
            Following approximations hold in many cases. We call them the ``ideal derivatives''.
            \begin{equation}
                \left(\rva \cdot \nabla_{\rvx_t}\right) \scorefn(\rvx_t, t) =
                    \frac{1}{\sqrt{\nu_t}} \rva, \quad
                - \diff{t} \scorefn(\rvx_t, t) =
                    -\frac{\beta_t}{2\sqrt{\nu_t}}
                    \left(\rvx_t - \frac{\scorefn(\rvx_t, t)}{\sqrt{\nu_t}} \right). \label{label:023}
            \end{equation}
        \end{conj}
        Using the above ideal derivatives as well as the relation $\dot{\nu}(t) = (1 - \nu_t)\beta_t$,
        we can compute symbolic expressions
        for $\ldet(-\fdet)$, $\lst(-\fst)$, $\lst(g)$, $\gst(-\fst)$ and $\gst(g)$
        by routine calculations, which can be easily automated by computer algebra systems
        such as SymPy~\citep{10.7717/peerj-cs.103} and Mathematica.
        These expressions consist of only $\rvx_t, \scorefn(\rvx_t, t), \nu_t, \beta_t$ and derivatives of $\beta_t$.
        Using them, we can derive Taylor schemes
        for both deterministic and stochastic systems as follows.

        \begin{alg}[Quasi-Taylor Sampler with Ideal Derivatives for PF-ODE]\label{label:024}
            Starting from a Gaussian noise $\rvx_T \sim \gN(\bm{0}, \mathbf{I})$,
            iterate the following refinement steps
            until $\rvx_0$ is obtained.
            \begin{gather}
                \rvx_{t - h} = \rho^\flat_{t,h} \rvx_t + \mu^\flat_{t,h} \scorefn(\rvx_t, t) / \sqrt{\nu_t},
                \text{where} \label{label:025}
            \\
            \displaystyle
                \rho^\flat_{t,h}
                    = 1 + \frac{\beta_th}{2}
                    + \frac{h^2}{4}
                    \left(
                        \frac{\beta_t^2}{2}
                        - \dot\beta_t
                    \right)
                    + \frac{h^3}{4}
                    \left(
                        \frac{\beta_t^3}{12}
                        - \frac{\beta_t\dot\beta_t}{2}
                        + \frac{\ddot\beta_t}{3}
                    \right) + \cdots,
                    \label{label:026}
            \\
            \displaystyle
                \mu^\flat_{t,h}
                    = -\frac{\beta_th}{2}
                    +
                    \frac{h^2}{4}
                    \left(
                        \dot{\beta}(t) - \frac{\beta^2(t)}{2\nu_t}
                    \right)
                    +
                    \frac{h^3}{4}
                    \left(
                        \frac{\beta_t^3(-\nu_t^2+3\nu_t - 3)}{12\nu_t^2}
                        +\frac{\beta_t\dot\beta_t}{2\nu_t}
                        -\frac{\ddot\beta_t}{3}
                    \right)+ \cdots.
            \label{label:027}
            \end{gather}
        \end{alg}
        Using terms up to $O(h^2)$,
        the sampler will have 2nd-order convergence (henceforth referred to as \textit{Taylor 2nd}),
        and using terms up to $O(h^3)$,
        the sampler will 3rd-order convergent (similarly, \textit{Taylor 3rd}).
        If we use up to the $O(h)$ terms, the algorithm is same as the Euler method.
        \begin{alg}[Quasi-It{\^o}-Taylor Sampler with Ideal Derivatives for R-SDE]\label{label:028}
            Starting from a Gaussian noise $\rvx_T \sim \gN(\bm{0}, \mathbf{I})$,
            iterate the following refinement steps
            until $\rvx_0$ is obtained.
            \begin{gather}
                \rvx_{t - h} = \rho^\sharp_{t,h} \rvx_t +
                    \mu^\sharp_{t,h} \scorefn(\rvx_t, t) /\sqrt{\nu_t}
                    + \rvn^\sharp_{t,h},
                    \text{where}
                    \label{label:029}
            \\
                \displaystyle
                \rho^\sharp_{t,h}
                    = 1 + \frac{\beta_t}{2}h +
                    \frac{h^2}{4}
                    \left(
                        \frac{\beta_t^2}{2} - \dot{\beta}(t)
                    \right),
                \qquad
                \mu^\sharp_{t,h}
                    = -\beta_th
                    + \frac{\dot{\beta}(t)h^2}{2},
                \label{label:030}
            \\
            \displaystyle
                \rvn^\sharp_{t,h}
                    = \sqrt{\beta_t}\sqrt{h} \rvw_t
                    + h^{3/2}\left(
                            - \frac{\dot\beta_t}{2 \sqrt{\beta_t}}(\rvw_t - \rvz_t)
                            + \frac{\beta_t^{3/2}(\nu_t - 2)}{2 \nu_t }\rvz_\tau
                    \right).
            \label{label:031}
            \end{gather}
            The Gaussian variables $\rvw_{t}$ and $\rvz_{t}$ have dimension-wise correlations,
            and each dimension is sampled similarly to Theorem~\ref{label:018}.
        \end{alg}
        At first glance, these algorithms may appear to be very complex.
        However, the computational complexity hardly increases compared to the Euler or Euler-Maruyama methods,
        because almost all of the computational cost is accounted for by the neural network $\scorefn(\rvx_t, t)$,
        and the costs for scalar values $\rho^\bullet_{t,h}, \mu^\bullet_{t,h}$
        and noise generation $\rvn^\sharp_{t,h}$
        are almost negligible.
        It should also be noted that these scalar values can be pre-computed and stored in the memory before synthesis.
        Thus the computational complexity of these methods are practically equal to Euler, Euler-Maruyama, and DDIM methods.

    \subsection{Specific Schedules of $\beta_t, \nu_t$ and $h$}
        \paragraph{Lipschitz-continuity-aware Noise Schedule Functions:}
        Let us specify the noise schedule.
        A simplest option would be the linear one, i.e.,
        $
            \beta_t = \beta_0 + 2 \beta_1 t,
            \nu_t = 1 - e^{- \beta_0t - \beta_1t^2},
        $
        and another would be the cosine schedule proposed in~\citep{nichol2021improved},
        $
            \nu_t = \sin^2 (\pi t /2),
            \beta_t = \min(\textit{threshold}, \pi\tan(\pi t / 2)).
        $
        However, we propose the following parametric noise schedule,
        \begin{equation}
            \beta_t = \dot\lambda(t) \tanh(\lambda(t)/{2}),
            \quad
            \nu_t = \tanh^2(\lambda(t)/{2}).
            \label{label:032}
        \end{equation}
        The parameter function $\lambda(t)$ has some options,
        but we chose the softplus,
        $\lambda(t) = \log(1 + A e^{kt})$ where $0 < A \ll 1$ and $0 < k$,
        because under the ideal-derivative approximation,
        it satisfies the
        Lipschitz continuity condition required by the Picard-Lindel\"of (Cauchy-Lipschitz) theorem,
        which guarantees the existence of the unique solution of the ODE;
        though it does not rigorously satisfy the requirement that $\nu_0=0$ (see \eqref{label:009}).

        These scheduling functions have some advantages.
        Firstly, all the coefficients do not diverge.
        In particular, the term $\beta_t/\sqrt{\nu_t}$ that appears in $\mu_\bullet(t, h) / \sqrt{\nu_t}$
        equals to $\dot\lambda(t)$ which does not explode near $t \sim 0$,
        contrary to the case of linear schedule where the term diverges,
        i.e.\ $\lim_{t \to 0} \beta_t/\sqrt{\nu_t} = \infty$.
        Moreover, unlike the cosine schedule where $\beta_t$
        is modified in an ad~hoc manner,
        our $\beta_t$ has an advantage  of being naturally bounded and differentiable
        (the differentiability of $\beta_t$ is particularly essential for Taylor schemes)
        while our $\nu_t$ has a similar curve to that of the empirically validated cosine schedule.
        Another merit is that its parameters $A,k$ are easy to tune,
        since they are explicitly written in a closed form as follows,
        if initial and terminal noise levels $\nu_0 \approx 0, \nu_T \approx 1$ are specified,
        \begin{equation}
        \displaystyle
            A = \frac{2\sqrt{\nu_0}}{1 - \sqrt{\nu_0}}, \quad
            k = \frac{1}{T}\left( \log \frac{2\sqrt{\nu_T}}{1 - \sqrt{\nu_T}} - \log A\right).
        \label{label:033}
        \end{equation}

        \paragraph{Schedule of step-size $h$:}
        The simplest step-size schedule is the equally spaced one, i.e., $h = T/N$,
        where $T$ is the number of refinement steps.
        Alternatively, another simple choice would be the use of exponentially decreasing $h_i$-s,
        i.e., $h_i = r^{i-1} h_1, (1 \le i \le N)$,
        where $h_i$ is the step-size of $i$-th refinement step.
        In this paper, $h_1$ and $r$ were determined so that
        $\sum_{i = 1}^N h_i = T$ and $r^N=0.1$.
        In our pilot study,
        the exponentially decreasing schedule was found to give faster convergence.

\begin{table*}[!t]
    \begingroup
        \centering
        \begin{subfigure}[t]{0.24\textwidth}
            \centering
            \includegraphics[width=0.95\linewidth]{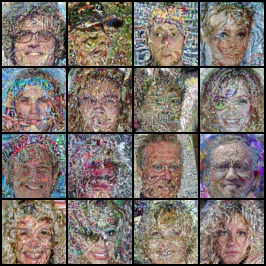}
            \caption{Euler}
        \end{subfigure}
        \begin{subfigure}[t]{0.24\textwidth}
            \centering
            \includegraphics[width=0.95\linewidth]{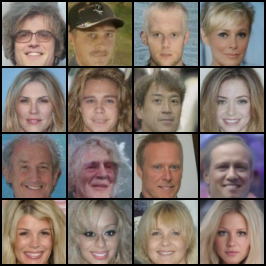}
            \caption{DDIM}
        \end{subfigure}
        \begin{subfigure}[t]{0.24\textwidth}
            \centering
            \includegraphics[width=0.95\linewidth]{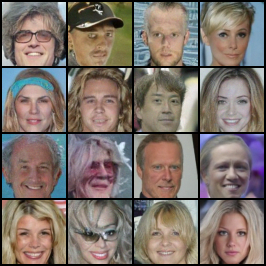}
            \caption{Heun (NFE=2)}
        \end{subfigure}
        \begin{subfigure}[t]{0.24\textwidth}
            \centering
            \includegraphics[width=0.95\linewidth]{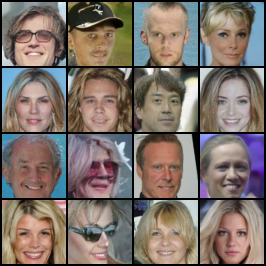}
            \caption{RK4 (NFE=4)}
        \end{subfigure}
        \begin{subfigure}[t]{0.24\textwidth}
            \centering
            \includegraphics[width=0.95\linewidth]{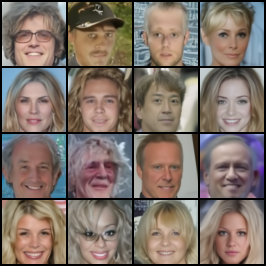}
            \caption{\textbf{Taylor (2nd)}}
        \end{subfigure}
        \begin{subfigure}[t]{0.24\textwidth}
            \centering
            \includegraphics[width=0.95\linewidth]{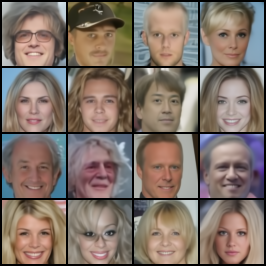}
            \caption{\textbf{Taylor (3rd)}}
        \end{subfigure}
        \begin{subfigure}[t]{0.24\textwidth}
            \centering
            \includegraphics[width=0.95\linewidth]{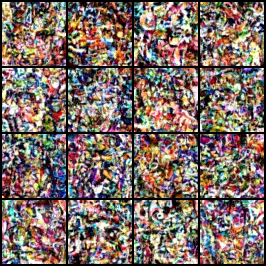}
            \caption{Euler-Maruyama}
        \end{subfigure}
        \begin{subfigure}[t]{0.24\textwidth}
            \centering
            \includegraphics[width=0.95\linewidth]{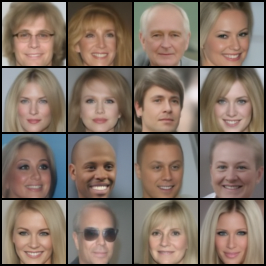}
            \caption{\textbf{It\^o-Taylor}}
        \end{subfigure}
        \captionof{figure}{
            Comparison of the synthesis results of CelebA ($64\times 64$) data.
            The noise schedule for synthesis was the condition (ii).
            The number of refinement steps is $N=12$.
        }
        \label{label:034}
    \endgroup
    \vspace*{0.5\floatsep}
    \begingroup
        \centering
        \begin{subfigure}[t]{0.49\textwidth}
            \centering
            \includegraphics[width=0.95\linewidth]{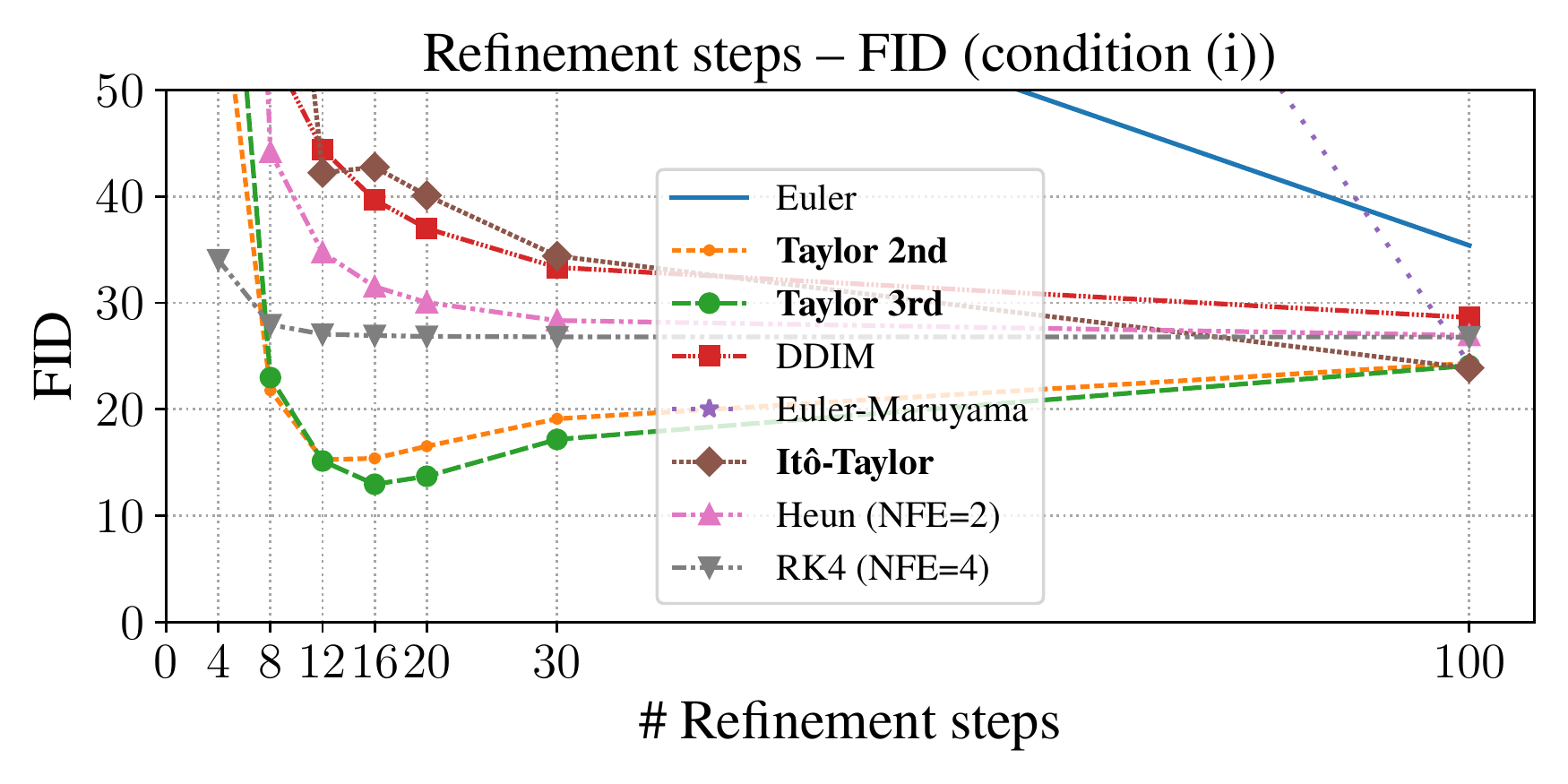}
        \end{subfigure}
        \begin{subfigure}[t]{0.49\textwidth}
            \centering
            \includegraphics[width=0.95\linewidth]{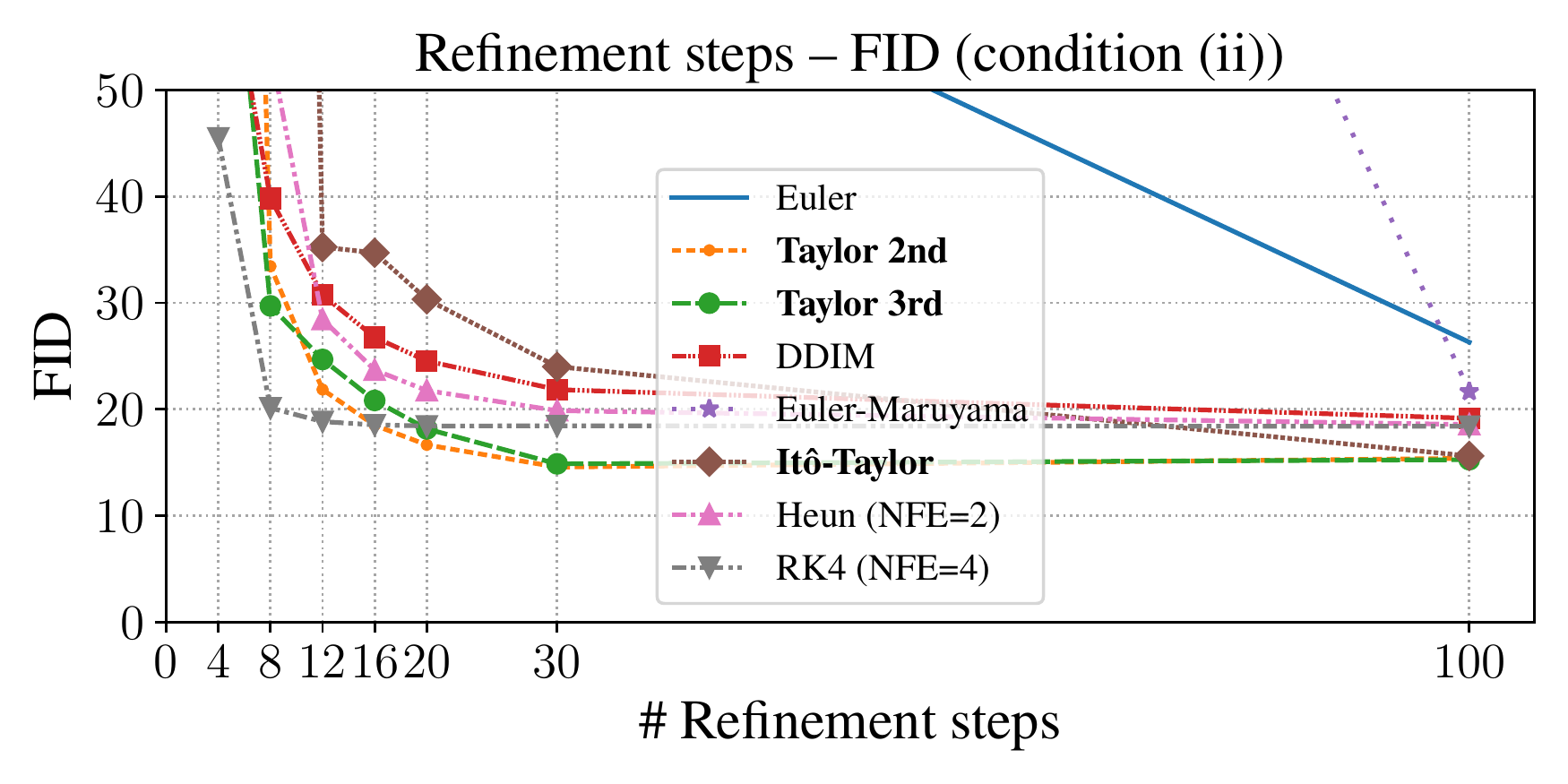}
        \end{subfigure}
        \captionof{figure}{
            FID scores ($\downarrow$) for generated CelebA images (64 $\times$ 64).
        }
        \label{label:035}
    \endgroup
\end{table*}

\section{Image Synthesis Experiment}\label{label:036}
    \paragraph{General Configuration:}
    In this section, we conduct experiments to verify the effectiveness of the methods developed in this paper.
    Specifically, we compare the performance of the
    Euler scheme \eqref{label:014},
    \textit{Taylor 2nd} (proposed; Alg.~\ref{label:024}), \textit{Taylor 3rd} (proposed; Alg.~\ref{label:024}),
    DDIM~\citep{song2020denoising}, and the Runge Kutta methods (Heun and Classical RK4 \secref{label:071};
    these are less efficient than others because of NFEs per step) for PF-ODE,
    as well as the Euler-Maruyama scheme \eqref{label:013}
    and \textit{It\^o-Taylor} (proposed; Alg.~\ref{label:028}) for R-SDE.

    \myparagraph{Data}
        In our experiment, we used the
        CIFAR-10 ($32 \times 32$)~\citep{Krizhevsky09learningmultiple} and
        CelebA ($64 \times 64$)~\citep{LiuLWT15} datasets.
        Separate networks were trained for each dataset.

    \myparagraph{Network and Loss}
        The network structure of $\scorefn(\rvx_t, t)$ and
        the experiment code were based on the implementation provided by \citet{song2020score}.
        The network structure was ``NCSN++'' implemented in their official PyTorch code.
        The network consisted of 4 levels of resolution,
        with the feature dimension of each level being $128 \to 128 \to 256 \to 256 \to 256$.
        Each level consisted of BigGAN-type ResBlocks,
        and the number of ResBlocks in each level was 8 (CIFAR-10) and 4 (CelebA).
        When feeding the time parameter $t$ to the network,
        we used the noise embedding $\sqrt{1 - \nu_t}$
        instead of the default $t$ embedding, following e.g.~\citep{chen2020wavegrad}.
        This can mitigate the noise schedule mismatches between training and synthesis,
        and will be useful to search synthesis noise schedules.
        The loss function we used was
        the SNR-weighted $L_2$ loss~\cite{kingma2021variational}.

    \myparagraph{Training}
        The optimizer was Adam~\citep{kingma2014adam}.
        The machine used for training was an in-house Linux server dedicated
        to medium-scale machine learning training with four GPUs (NVIDIA Tesla V100).
        The batch size was 256.
        The number of training steps was 100k,
        and the training took about a day for each dataset.
        The integration duration was $T=1$,
        and the noising schedule was determined by the
        initial and terminal values $(\nu_0, \nu_T) = (5 \times 10^{-4}, 0.995)$.

    \myparagraph{Synthesis and Evaluation}
        During the synthesis, the noise schedule was
        determined by the following two initial and terminal values;
        condition (i): $(\nu_0, \nu_T) = (5 \times 10^{-4}, 0.995)$,
        and condition (ii): $(\nu_0, \nu_T) = (1 \times 10^{-4}, 0.99)$.
        As a quality assessment metric,
        we used the Fr\'echet Inception Distance (FID)~\citep{heusel2017gans,Seitzer2020FID}.
        50,000 images were randomly generated for each condition to compute the FID scores.

\begin{table*}[!t]
    \begingroup
        \centering
        \begin{subfigure}[t]{0.24\textwidth}
            \centering
            \includegraphics[width=0.95\linewidth]{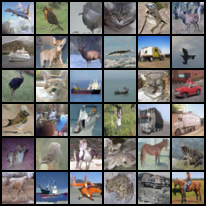}
            \caption{DDIM}
        \end{subfigure}
        \begin{subfigure}[t]{0.24\textwidth}
            \centering
            \includegraphics[width=0.95\linewidth]{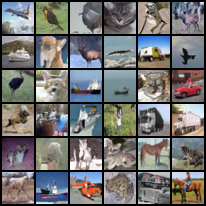}
            \caption{RK4 (NFE=4)}
        \end{subfigure}
        \begin{subfigure}[t]{0.24\textwidth}
            \centering
            \includegraphics[width=0.95\linewidth]{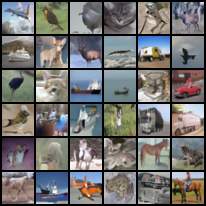}
            \caption{\textbf{Taylor (3rd)}}
        \end{subfigure}
        \begin{subfigure}[t]{0.24\textwidth}
            \centering
            \includegraphics[width=0.95\linewidth]{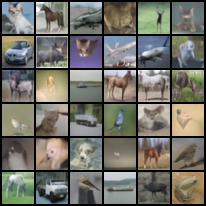}
            \caption{\textbf{It\^o-Taylor}}
        \end{subfigure}
        \captionof{figure}{
            Comparison of the synthesis results of CIFAR-10 ($32\times 32$) data.
            The noise schedule for synthesis was the condition (ii).
            The number of refinement steps is $N=30$.
        }
        \label{label:037}
    \endgroup
    \vspace*{0.5\floatsep}
    \begingroup
        \centering
        \begin{subfigure}[t]{0.49\textwidth}
            \centering
            \includegraphics[width=0.95\linewidth]{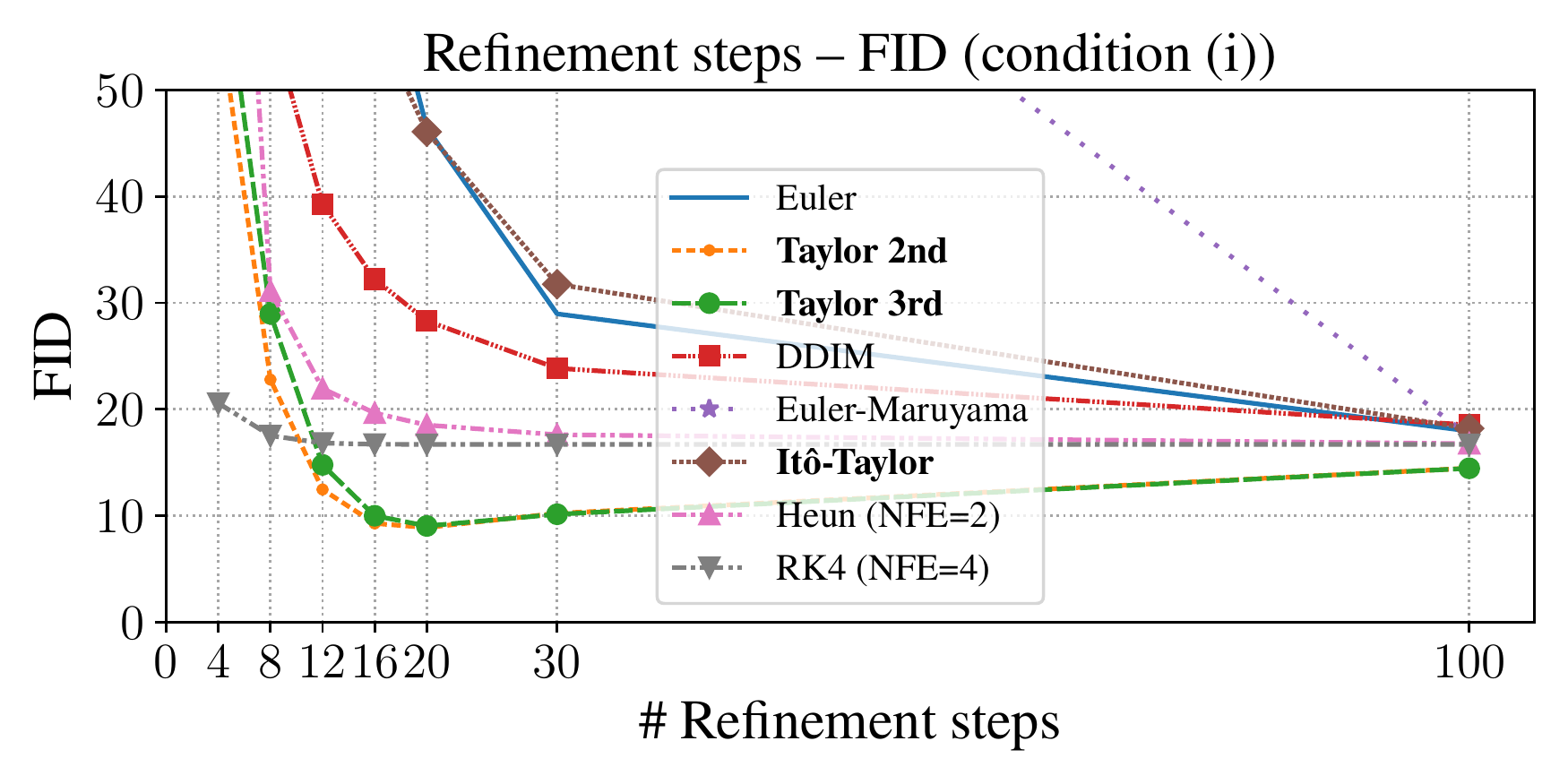}
        \end{subfigure}
        \begin{subfigure}[t]{0.49\textwidth}
            \centering
            \includegraphics[width=0.95\linewidth]{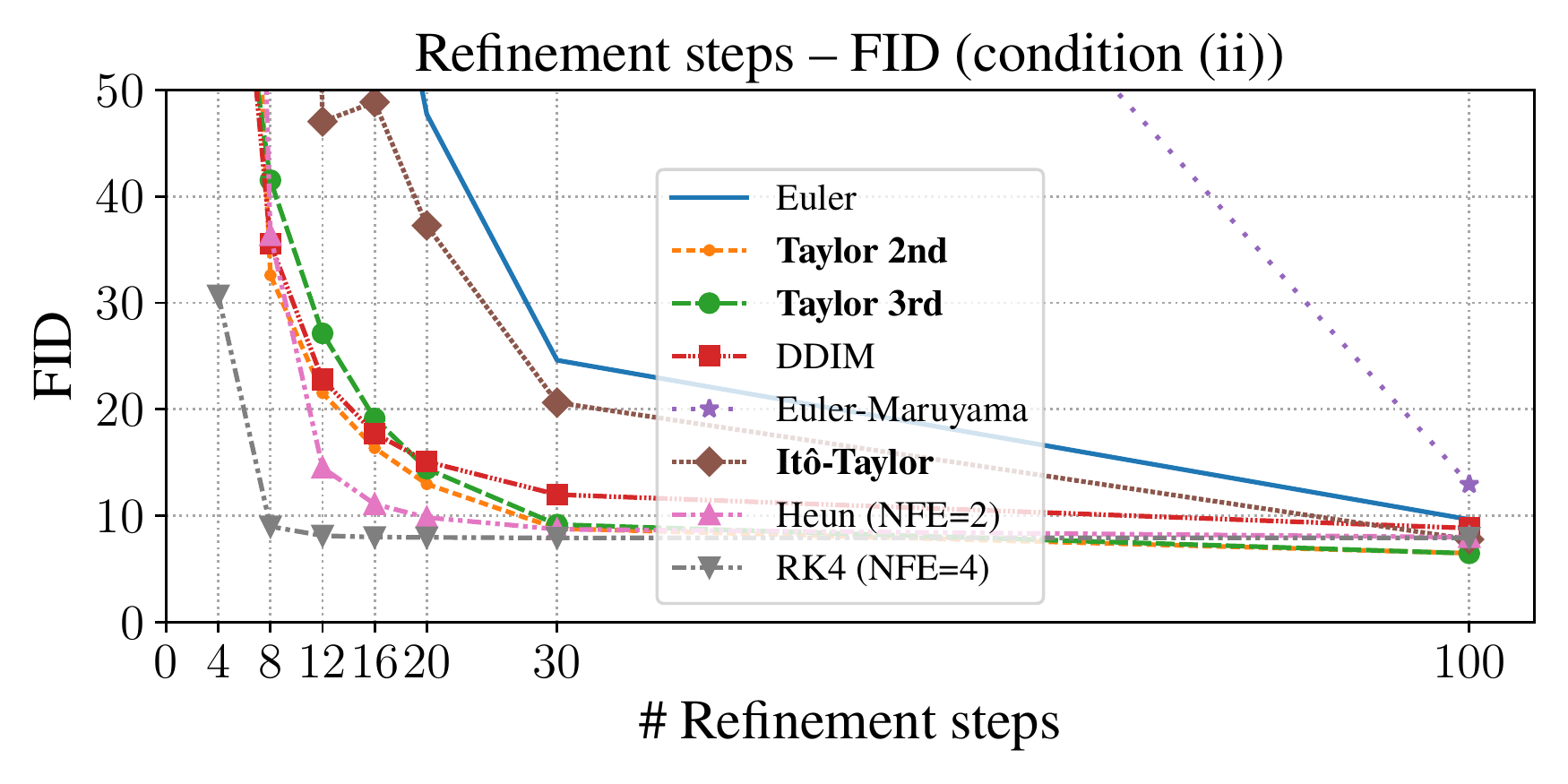}
        \end{subfigure}
        \captionof{figure}{
            FID scores ($\downarrow$) for generated CIFAR-10 images (32 $\times$ 32).
        }
        \label{label:038}
    \endgroup
\end{table*}

    \paragraph{Result and Discussion:}
    \figref{label:034} and \figref{label:037} show random samples for each sampler.
    More examples are available in \secref{label:076}.
    \figref{label:035} and \figref{label:038} reports the FID scores.
    The deterministic samplers considered in this paper generated plausible images much faster than the
    vanilla Euler-Maruyama sampler,
    and the proposed stochastic sampler, i.e.\ the It\^o-Taylor, also showed sufficiently competitive results,
    in terms of both FID scores and visual impression.
    Based on these results,
    we may conclude that the use of the Euler-Maruyama sampler is not recommended,
    but other higher-order methods should be used in situations where the synthesis efficiency matters.
    In particular, when the synthesis noise schedule was (i),
    the proposed Quasi-Taylor methods tend to give good results around $N=16, 20$,
    and the FID deteriorates from there.
    The cause is yet unclear at this point, and should be investigated in the future.
    Nevertheless, regardless of the cause, this fact may be useful in applications.

    It can be observed that all of the PF-ODE solvers produce images with generally similar trends,
    although there is some variation in quality.
    This would be the consequence of the fact that all the methods are
    solving the same ODE although the ideas and the accuracy are different.
    The R-SDE solvers
    produced images that tend to be slightly different from these deterministic samplers
    because they are solving different problems.
    However, some similarities can be seen in the overall direction (see \figref{label:077}).

    What would be particularly important here would be the
    fact that our Quasi-Taylor methods show qualitatively similar results as
    derivative-free Runge-Kutta methods (Heun and Classical RK4).
    This implicitly suggests that
    the approximated \textit{ideal derivatives} of score function we are using
    are actually very close to the \textit{true derivatives},
    since the $p$-th order Runge-Kutta method actually approximates up to $p$-th order terms
    of the \textit{true} Taylor series based on the true derivatives of the learned score function; see \secref{label:071}.
    This implicitly supports the validity of our hypotheses
    of {single point approximation}, {ideal derivative approximation},
    and assumptions to support them.

    Another important issue is the relation to the DDIM.
    As we will discuss in \secref{label:054},
    DDIM is identical to our Quasi-Taylor methods at least up to 3rd order terms,
    which suggests that the ideas behind DDIM and our ideal derivative are logically closely related.
    However, despite this fact, DDIM is considerably less effective than the Quasi-Taylor methods in some cases.
    It could be hypothesized that this difference may be due to the less-reliable higher order derivatives that are automatically
    incorporated into DDIM, which may have a negative effect.
    Further studies will be needed on this point.

\section{Concluding Remarks}
    This paper proposed a Taylor-expansion approach for diffusion generative models,
    particularly the Probability Flow ODE (PF-ODE) and the reverse-time SDE (R-SDE) solvers.
    The assumptions to derive our sampler were minimalistic,
    and the derivation process was straightforward.
    We just substituted the derivatives in the Taylor series by ideal ones
    which are approximately specified by the model structure.

    The obtained Quasi-Taylor and Quasi-It\^o-Taylor samplers performed better than or on par with
    DDIM and Runge-Kutta methods.
    This fact implicitly supports the validity of our approximations.
    Conversely, if we could find some examples where the Quasi-Taylor methods, DDIM and RK methods gave
    decisively different results,
    we might be able to gain a deeper understanding of the structure of data manifold and
    the fundamentals of diffusion models by investigating the causes of discrepancy.

    \paragraph{Reproducibility Statement}
    Pseudocodes of the proposed methods are available in \secref{label:074},
    and the derivation of the proposed method is described in \secref{label:050}, \secref{label:055}.
    The experiment is based on open source code with minimal modifications to match the proposed method,
    and all the data used in this paper are publicly available.
    Experimental conditions are elaborated in \secref{label:036}.

    \paragraph{Ethics Statement}
    As a final note,
    negative aspects of generative models are generally pointed out,
    such as the risk of reproducing bias and discrimination in training data and the risk of
    being misused for deep fakes.
    Since this method only provides a solution to existing generative models,
    it does not take special measures against these problems.
    Maximum ethical care should be taken in the practical application of this method.

\bibliography{tachibana}

\begin{thebibliography}{62}
\providecommand{\natexlab}[1]{#1}
\providecommand{\url}[1]{\texttt{#1}}
\expandafter\ifx\csname urlstyle\endcsname\relax
  \providecommand{\doi}[1]{doi: #1}\else
  \providecommand{\doi}{doi: \begingroup \urlstyle{rm}\Url}\fi

\bibitem[Anderson(1982)]{anderson1982reverse}
Brian~DO Anderson.
\newblock Reverse-time diffusion equation models.
\newblock \emph{Stochastic Processes and their Applications}, 12\penalty0
  (3):\penalty0 313--326, 1982.

\bibitem[Austin et~al.(2021)Austin, Johnson, Ho, Tarlow, and van~den
  Berg]{austin2021structured}
Jacob Austin, Daniel Johnson, Jonathan Ho, Daniel Tarlow, and Rianne van~den
  Berg.
\newblock Structured denoising diffusion models in discrete state-spaces.
\newblock \emph{Advances in Neural Information Processing Systems}, 34, 2021.

\bibitem[Chen et~al.(2020)Chen, Zhang, Zen, Weiss, Norouzi, and
  Chan]{chen2020wavegrad}
Nanxin Chen, Yu~Zhang, Heiga Zen, Ron~J Weiss, Mohammad Norouzi, and William
  Chan.
\newblock {WaveGrad}: Estimating gradients for waveform generation.
\newblock In \emph{International Conference on Learning Representations
  \textup{(}ICLR\textup{)}}, 2020.

\bibitem[Chen et~al.(2021)Chen, Zhang, Zen, Weiss, Norouzi, Dehak, and
  Chan]{chen2021wavegrad}
Nanxin Chen, Yu~Zhang, Heiga Zen, Ron~J Weiss, Mohammad Norouzi, Najim Dehak,
  and William Chan.
\newblock {WaveGrad} 2: Iterative refinement for text-to-speech synthesis.
\newblock In \emph{Proceedings of INTERSPEECH}, pp.\  3765--3769. ISCA, 2021.

\bibitem[Chen et~al.(2018)Chen, Rubanova, Bettencourt, and
  Duvenaud]{chen2018neural}
Ricky~TQ Chen, Yulia Rubanova, Jesse Bettencourt, and David Duvenaud.
\newblock Neural ordinary differential equations.
\newblock \emph{arXiv preprint arXiv:1806.07366}, 2018.

\bibitem[Choi et~al.(2021)Choi, Kim, Jeong, Gwon, and Yoon]{choi2021ilvr}
Jooyoung Choi, Sungwon Kim, Yonghyun Jeong, Youngjune Gwon, and Sungroh Yoon.
\newblock {ILVR}: Conditioning method for denoising diffusion probabilistic
  models.
\newblock \emph{arXiv preprint arXiv:2108.02938}, 2021.

\bibitem[Cox \& Miller(1965)Cox and Miller]{cox2017theory}
David~Roxbee Cox and Hilton~David Miller.
\newblock \emph{The theory of stochastic processes}.
\newblock Routledge, 1965.

\bibitem[Dhariwal \& Nichol(2021)Dhariwal and Nichol]{dhariwal2021diffusion}
Prafulla Dhariwal and Alexander~Quinn Nichol.
\newblock Diffusion models beat {GAN}s on image synthesis.
\newblock In \emph{Advances in Neural Information Processing Systems}, 2021.

\bibitem[Dockhorn et~al.(2021)Dockhorn, Vahdat, and Kreis]{dockhorn2021score}
Tim Dockhorn, Arash Vahdat, and Karsten Kreis.
\newblock Score-based generative modeling with critically-damped {Langevin}
  diffusion.
\newblock \emph{arXiv preprint arXiv:2112.07068}, 2021.

\bibitem[Goodfellow et~al.(2014)Goodfellow, Pouget-Abadie, Mirza, Xu,
  Warde-Farley, Ozair, Courville, and Bengio]{goodfellow2014generative}
Ian Goodfellow, Jean Pouget-Abadie, Mehdi Mirza, Bing Xu, David Warde-Farley,
  Sherjil Ozair, Aaron Courville, and Yoshua Bengio.
\newblock Generative adversarial nets.
\newblock \emph{Advances in Neural Information Processing Systems}, 27, 2014.

\bibitem[Heusel et~al.(2017)Heusel, Ramsauer, Unterthiner, Nessler, and
  Hochreiter]{heusel2017gans}
Martin Heusel, Hubert Ramsauer, Thomas Unterthiner, Bernhard Nessler, and Sepp
  Hochreiter.
\newblock {GANs} trained by a two time-scale update rule converge to a local
  {Nash} equilibrium.
\newblock \emph{Advances in Neural Information Processing Systems}, 30, 2017.

\bibitem[Ho \& Salimans(2021)Ho and Salimans]{ho2021classifier}
Jonathan Ho and Tim Salimans.
\newblock Classifier-free diffusion guidance.
\newblock In \emph{NeurIPS 2021 Workshop on Deep Generative Models and
  Downstream Applications}, 2021.

\bibitem[Ho et~al.(2020)Ho, Jain, and Abbeel]{ho2020denoising}
Jonathan Ho, Ajay Jain, and Pieter Abbeel.
\newblock Denoising diffusion probabilistic models.
\newblock \emph{Advances in Neural Information Processing Systems}, 2020.

\bibitem[Ho et~al.(2021)Ho, Saharia, Chan, Fleet, Norouzi, and
  Salimans]{ho2021cascaded}
Jonathan Ho, Chitwan Saharia, William Chan, David~J Fleet, Mohammad Norouzi,
  and Tim Salimans.
\newblock Cascaded diffusion models for high fidelity image generation.
\newblock \emph{arXiv preprint arXiv:2106.15282}, 2021.

\bibitem[Ho et~al.(2022)Ho, Salimans, Gritsenko, Chan, Norouzi, and
  Fleet]{ho2022video}
Jonathan Ho, Tim Salimans, Alexey Gritsenko, William Chan, Mohammad Norouzi,
  and David~J Fleet.
\newblock Video diffusion models.
\newblock \emph{arXiv preprint arXiv:2204.03458}, 2022.

\bibitem[Hoogeboom et~al.(2021)Hoogeboom, Nielsen, Jaini, Forr{\'e}, and
  Welling]{hoogeboom2021argmax}
Emiel Hoogeboom, Didrik Nielsen, Priyank Jaini, Patrick Forr{\'e}, and Max
  Welling.
\newblock Argmax flows and multinomial diffusion: Towards non-autoregressive
  language models.
\newblock \emph{Advances in Neural Information Processing Systems}, 2021.

\bibitem[Hyv{\"a}rinen \& Dayan(2005)Hyv{\"a}rinen and
  Dayan]{hyvarinen2005estimation}
Aapo Hyv{\"a}rinen and Peter Dayan.
\newblock Estimation of non-normalized statistical models by score matching.
\newblock \emph{Journal of Machine Learning Research}, 6\penalty0 (4), 2005.

\bibitem[It{\^o}(1944)]{ito1944}
Kiyosi It{\^o}.
\newblock Stochastic integral.
\newblock \emph{Proceedings of the Imperial Academy}, 20\penalty0 (8):\penalty0
  519--524, 1944.
\newblock \doi{10.3792/pia/1195572786}.

\bibitem[Jolicoeur-Martineau et~al.(2021)Jolicoeur-Martineau, Li,
  Pich{\'e}-Taillefer, Kachman, and Mitliagkas]{jolicoeur2021gotta}
Alexia Jolicoeur-Martineau, Ke~Li, R{\'e}mi Pich{\'e}-Taillefer, Tal Kachman,
  and Ioannis Mitliagkas.
\newblock Gotta go fast when generating data with score-based models.
\newblock \emph{arXiv preprint arXiv:2105.14080}, 2021.

\bibitem[Kameoka et~al.(2020)Kameoka, Kaneko, Tanaka, Hojo, and
  Seki]{kameoka2020voicegrad}
Hirokazu Kameoka, Takuhiro Kaneko, Kou Tanaka, Nobukatsu Hojo, and Shogo Seki.
\newblock {VoiceGrad}: Non-parallel any-to-many voice conversion with annealed
  {Langevin} dynamics.
\newblock \emph{arXiv preprint arXiv:2010.02977}, 2020.

\bibitem[Karlin \& Taylor(1981)Karlin and Taylor]{karlin1981second}
Samuel Karlin and Howard~E Taylor.
\newblock \emph{A second course in stochastic processes}.
\newblock Academic Press, 1981.

\bibitem[Kingma \& Ba(2014)Kingma and Ba]{kingma2014adam}
Diederik~P Kingma and Jimmy Ba.
\newblock Adam: A method for stochastic optimization.
\newblock \emph{arXiv preprint arXiv:1412.6980}, 2014.

\bibitem[Kingma \& Welling(2019)Kingma and Welling]{kingma2019introduction}
Diederik~P Kingma and Max Welling.
\newblock An introduction to variational autoencoders.
\newblock \emph{arXiv preprint arXiv:1906.02691}, 2019.

\bibitem[Kingma et~al.(2021)Kingma, Salimans, Poole, and
  Ho]{kingma2021variational}
Diederik~P Kingma, Tim Salimans, Ben Poole, and Jonathan Ho.
\newblock Variational diffusion models.
\newblock \emph{arXiv preprint arXiv:2107.00630}, 2021.

\bibitem[Kloeden \& Platen(1992)Kloeden and Platen]{kloeden1992stochastic}
Peter~E Kloeden and Eckhard Platen.
\newblock Numerical solution of stochastic differential equations.
\newblock Stochastic Modelling and Applied Probability, Applications of
  Mathematics. Springer, 1992.
\newblock \doi{10.1007/978-3-662-12616-5}.
\newblock Corrected 3rd Printing.

\bibitem[Kloeden et~al.(1994)Kloeden, Platen, and Schurz]{kloeden1994numerical}
Peter~E Kloeden, Eckhard Platen, and Henri Schurz.
\newblock \emph{Numerical solution of SDE through computer experiments}.
\newblock Universitext. Springer, 1994.
\newblock \doi{10.1007/978-3-642-57913-4}.
\newblock Corrected 3rd Printing.

\bibitem[Kong et~al.(2021)Kong, Ping, Huang, Zhao, and
  Catanzaro]{kong2020diffwave}
Zhifeng Kong, Wei Ping, Jiaji Huang, Kexin Zhao, and Bryan Catanzaro.
\newblock Diffwave: A versatile diffusion model for audio synthesis.
\newblock In \emph{International Conference on Learning Representations
  \textup{(}ICLR\textup{)}}, 2021.

\bibitem[Krizhevsky(2009)]{Krizhevsky09learningmultiple}
Alex Krizhevsky.
\newblock Learning multiple layers of features from tiny images.
\newblock Technical report, 2009.
\newblock Dataset license: MIT.

\bibitem[Laurent \& Massart(2000)Laurent and Massart]{laurent2000adaptive}
Beatrice Laurent and Pascal Massart.
\newblock Adaptive estimation of a quadratic functional by model selection.
\newblock \emph{Annals of Statistics}, pp.\  1302--1338, 2000.

\bibitem[LeCun et~al.(2010)LeCun, Cortes, and Burges]{lecun2010mnist}
Yann LeCun, Corinna Cortes, and CJ~Burges.
\newblock {MNIST} handwritten digit database, 2010.
\newblock \url{http://yann.lecun.com/exdb/mnist}, Dataset license: CC-SA 3.0.

\bibitem[Liu et~al.(2015)Liu, Luo, Wang, and Tang]{LiuLWT15}
Ziwei Liu, Ping Luo, Xiaogang Wang, and Xiaoou Tang.
\newblock Deep learning face attributes in the wild.
\newblock In \emph{Proceedings of the IEEE international conference on computer
  vision \textup{(}ICCV\textup{)}}, pp.\  3730--3738, 2015.
\newblock Dataset license: available for non-commercial research purpose only.

\bibitem[Maruyama(1955)]{maruyama1955continuous}
Gisir{\=o} Maruyama.
\newblock Continuous {Markov} processes and stochastic equations.
\newblock \emph{Rendiconti del Circolo Matematico di Palermo}, 4\penalty0
  (1):\penalty0 48--90, 1955.

\bibitem[Meurer et~al.(2017)Meurer, Smith, Paprocki, \v{C}ert\'{i}k, Kirpichev,
  Rocklin, Kumar, Ivanov, Moore, Singh, Rathnayake, Vig, Granger, Muller,
  Bonazzi, Gupta, Vats, Johansson, Pedregosa, Curry, Terrel, Rou\v{c}ka, Saboo,
  Fernando, Kulal, Cimrman, and Scopatz]{10.7717/peerj-cs.103}
Aaron Meurer, Christopher~P. Smith, Mateusz Paprocki, Ond\v{r}ej
  \v{C}ert\'{i}k, Sergey~B. Kirpichev, Matthew Rocklin, AMiT Kumar, Sergiu
  Ivanov, Jason~K. Moore, Sartaj Singh, Thilina Rathnayake, Sean Vig, Brian~E.
  Granger, Richard~P. Muller, Francesco Bonazzi, Harsh Gupta, Shivam Vats,
  Fredrik Johansson, Fabian Pedregosa, Matthew~J. Curry, Andy~R. Terrel,
  \v{S}t\v{e}p\'{a}n Rou\v{c}ka, Ashutosh Saboo, Isuru Fernando, Sumith Kulal,
  Robert Cimrman, and Anthony Scopatz.
\newblock {SymPy}: symbolic computing in {Python}.
\newblock \emph{PeerJ Computer Science}, 3:\penalty0 e103, January 2017.
\newblock ISSN 2376-5992.
\newblock \doi{10.7717/peerj-cs.103}.
\newblock URL \url{https://doi.org/10.7717/peerj-cs.103}.

\bibitem[Mittal et~al.(2021)Mittal, Engel, Hawthorne, and
  Simon]{mittal2021symbolic}
Gautam Mittal, Jesse Engel, Curtis Hawthorne, and Ian Simon.
\newblock Symbolic music generation with diffusion models.
\newblock In \emph{Proceedings of the 22nd International Society for Music
  Information Retrieval Conference \textup{(}ISMIR\textup{)}}, 2021.

\bibitem[Nichol \& Dhariwal(2021)Nichol and Dhariwal]{nichol2021improved}
Alex Nichol and Prafulla Dhariwal.
\newblock Improved denoising diffusion probabilistic models.
\newblock In \emph{International Conference on Machine Learning
  \textup{(}ICML\textup{)}}. PMLR, 2021.

\bibitem[Nichol et~al.(2021)Nichol, Dhariwal, Ramesh, Shyam, Mishkin, McGrew,
  Sutskever, and Chen]{nichol2021glide}
Alex Nichol, Prafulla Dhariwal, Aditya Ramesh, Pranav Shyam, Pamela Mishkin,
  Bob McGrew, Ilya Sutskever, and Mark Chen.
\newblock Glide: Towards photorealistic image generation and editing with
  text-guided diffusion models.
\newblock \emph{arXiv preprint arXiv:2112.10741}, 2021.

\bibitem[{\O}ksendal(2013)]{oksendal2013stochastic}
Bernt {\O}ksendal.
\newblock \emph{Stochastic differential equations: an introduction with
  applications}.
\newblock Springer, 2013.

\bibitem[Platen \& Wagner(1982)Platen and Wagner]{wagner1982taylor}
Eckhard Platen and Wolfgang Wagner.
\newblock On a {Taylor} formula for a class of {Ito} processes.
\newblock \emph{Probability and Mathematical Statistics}, 3:\penalty0 37--51,
  1982.

\bibitem[Popov et~al.(2021)Popov, Vovk, Gogoryan, Sadekova, and
  Kudinov]{popov2021grad}
Vadim Popov, Ivan Vovk, Vladimir Gogoryan, Tasnima Sadekova, and Mikhail
  Kudinov.
\newblock {Grad-TTS}: A diffusion probabilistic model for text-to-speech.
\newblock In \emph{International Conference on Machine Learning
  \textup{(}ICML\textup{)}}, pp.\  8599--8608. PMLR, 2021.

\bibitem[Press et~al.(2007)Press, Teukolsky, Vetterling, and
  Flannery]{press2007numerical}
William~H Press, Saul~A Teukolsky, William~T Vetterling, and Brian~P Flannery.
\newblock \emph{Numerical recipes 3rd edition: The art of scientific
  computing}.
\newblock Cambridge university press, 2007.

\bibitem[Ramesh et~al.(2022)Ramesh, Dhariwal, Nichol, Chu, and
  Chen]{ramesh2022hierarchical}
Aditya Ramesh, Prafulla Dhariwal, Alex Nichol, Casey Chu, and Mark Chen.
\newblock Hierarchical text-conditional image generation with {CLIP} latents.
\newblock \emph{arXiv preprint arXiv:2204.06125}, 2022.

\bibitem[Rezende \& Mohamed(2015)Rezende and Mohamed]{rezende2015variational}
Danilo Rezende and Shakir Mohamed.
\newblock Variational inference with normalizing flows.
\newblock In \emph{International conference on machine learning}, pp.\
  1530--1538. PMLR, 2015.

\bibitem[Roberts \& Tweedie(1996)Roberts and Tweedie]{roberts1996exponential}
Gareth~O Roberts and Richard~L Tweedie.
\newblock Exponential convergence of {Langevin} distributions and their
  discrete approximations.
\newblock \emph{Bernoulli}, pp.\  341--363, 1996.

\bibitem[Rossky et~al.(1978)Rossky, Doll, and Friedman]{rossky1978brownian}
Peter~J Rossky, Jimmie~D Doll, and Harold~L Friedman.
\newblock Brownian dynamics as smart {Monte} {Carlo} simulation.
\newblock \emph{The Journal of Chemical Physics}, 69\penalty0 (10):\penalty0
  4628--4633, 1978.

\bibitem[Saharia et~al.(2021)Saharia, Ho, Chan, Salimans, Fleet, and
  Norouzi]{saharia2021image}
Chitwan Saharia, Jonathan Ho, William Chan, Tim Salimans, David~J Fleet, and
  Mohammad Norouzi.
\newblock Image super-resolution via iterative refinement.
\newblock \emph{arXiv preprint arXiv:2104.07636}, 2021.

\bibitem[Salimans \& Ho(2022)Salimans and Ho]{salimans2022progressive}
Tim Salimans and Jonathan Ho.
\newblock Progressive distillation for fast sampling of diffusion models.
\newblock \emph{arXiv preprint arXiv:2202.00512}, 2022.

\bibitem[S{\"a}rkk{\"a} \& Solin(2019)S{\"a}rkk{\"a} and
  Solin]{sarkka2019applied}
Simo S{\"a}rkk{\"a} and Arno Solin.
\newblock \emph{Applied stochastic differential equations}, volume~10 of
  \emph{Institute of Mathematical Statistics Textbooks}.
\newblock Cambridge University Press, 2019.

\bibitem[Sasaki et~al.(2021)Sasaki, Willcocks, and Breckon]{sasaki2021unit}
Hiroshi Sasaki, Chris~G Willcocks, and Toby~P Breckon.
\newblock {UNIT-DDPM}: Unpaired image translation with denoising diffusion
  probabilistic models.
\newblock \emph{arXiv preprint arXiv:2104.05358}, 2021.

\bibitem[Seitzer(2020)]{Seitzer2020FID}
Maximilian Seitzer.
\newblock {pytorch-fid: FID Score for PyTorch}.
\newblock \url{https://github.com/mseitzer/pytorch-fid} (Apache-2.0 license),
  August 2020.
\newblock Version 0.1.1.

\bibitem[Shreve(2004)]{shreve2004stochastic}
Steven~E Shreve.
\newblock \emph{Stochastic calculus for finance {II}: Continuous-time models},
  volume~11.
\newblock Springer, 2004.

\bibitem[Sohl-Dickstein et~al.(2015)Sohl-Dickstein, Weiss, Maheswaranathan, and
  Ganguli]{sohl2015deep}
Jascha Sohl-Dickstein, Eric Weiss, Niru Maheswaranathan, and Surya Ganguli.
\newblock Deep unsupervised learning using nonequilibrium thermodynamics.
\newblock In \emph{International Conference on Machine Learning
  \textup{(}ICML\textup{)}}, pp.\  2256--2265. PMLR, 2015.

\bibitem[Song et~al.(2020{\natexlab{a}})Song, Meng, and
  Ermon]{song2020denoising}
Jiaming Song, Chenlin Meng, and Stefano Ermon.
\newblock Denoising diffusion implicit models.
\newblock In \emph{International Conference on Learning Representations
  \textup{(}ICLR\textup{)}}, 2020{\natexlab{a}}.

\bibitem[Song \& Ermon(2019)Song and Ermon]{song2019generative}
Yang Song and Stefano Ermon.
\newblock Generative modeling by estimating gradients of the data distribution.
\newblock \emph{Advances in Neural Information Processing Systems}, 2019.

\bibitem[Song \& Ermon(2020)Song and Ermon]{song2020improved}
Yang Song and Stefano Ermon.
\newblock Improved techniques for training score-based generative models.
\newblock \emph{Advances in Neural Information Processing Systems}, 2020.

\bibitem[Song et~al.(2020{\natexlab{b}})Song, Sohl-Dickstein, Kingma, Kumar,
  Ermon, and Poole]{song2020score}
Yang Song, Jascha Sohl-Dickstein, Diederik~P Kingma, Abhishek Kumar, Stefano
  Ermon, and Ben Poole.
\newblock Score-based generative modeling through stochastic differential
  equations.
\newblock In \emph{International Conference on Learning Representations
  \textup{(}ICLR\textup{)}}, 2020{\natexlab{b}}.
\newblock \url{https://github.com/yang-song/score\_sde\_pytorch} (Apache-2.0
  license).

\bibitem[Stratonovich(1965)]{stratonovich1965conditional}
Ruslan~Leont’evich Stratonovich.
\newblock Conditional {Markov} processes.
\newblock In \emph{Non-linear transformations of stochastic processes}, pp.\
  427--453. Elsevier, 1965.

\bibitem[Vahdat et~al.(2021)Vahdat, Kreis, and Kautz]{vahdat2021score}
Arash Vahdat, Karsten Kreis, and Jan Kautz.
\newblock Score-based generative modeling in latent space.
\newblock \emph{Advances in Neural Information Processing Systems}, 2021.

\bibitem[van~den Oord et~al.(2016{\natexlab{a}})van~den Oord, Dieleman, Zen,
  Simonyan, Vinyals, Graves, Kalchbrenner, Senior, and
  Kavukcuoglu]{oord2016wavenet}
Aaron van~den Oord, Sander Dieleman, Heiga Zen, Karen Simonyan, Oriol Vinyals,
  Alex Graves, Nal Kalchbrenner, Andrew Senior, and Koray Kavukcuoglu.
\newblock {WaveNet}: A generative model for raw audio.
\newblock \emph{arXiv preprint arXiv:1609.03499}, 2016{\natexlab{a}}.

\bibitem[van~den Oord et~al.(2016{\natexlab{b}})van~den Oord, Kalchbrenner, and
  Kavukcuoglu]{van2016pixel}
Aaron van~den Oord, Nal Kalchbrenner, and Koray Kavukcuoglu.
\newblock Pixel recurrent neural networks.
\newblock In \emph{International conference on machine learning}, pp.\
  1747--1756. PMLR, 2016{\natexlab{b}}.

\bibitem[Vincent(2011)]{vincent2011connection}
Pascal Vincent.
\newblock A connection between score matching and denoising autoencoders.
\newblock \emph{Neural computation}, 23\penalty0 (7):\penalty0 1661--1674,
  2011.

\bibitem[Watson et~al.(2021)Watson, Ho, Norouzi, and Chan]{watson2021learning}
Daniel Watson, Jonathan Ho, Mohammad Norouzi, and William Chan.
\newblock Learning to efficiently sample from diffusion probabilistic models.
\newblock \emph{arXiv preprint arXiv:2106.03802}, 2021.

\bibitem[Xu et~al.(2022)Xu, Yu, Song, Shi, Ermon, and Tang]{xu2022geodiff}
Minkai Xu, Lantao Yu, Yang Song, Chence Shi, Stefano Ermon, and Jian Tang.
\newblock Geodiff: A geometric diffusion model for molecular conformation
  generation.
\newblock \emph{arXiv preprint arXiv:2203.02923}, 2022.

\end{thebibliography}
\bibliographystyle{iclr2023_conference}

\vfill
\newpage

\appendix
\section{On the Approximation
$\nabla_{\rvx_t} \log p(\rvx_t, t) \approx
\nabla_{\rvx_t} \log p(\rvx_t, t \mid \rvx_0, 0)$}\label{label:039}

    Although $\nabla_{\rvx_t} \log p(\rvx_t, t)$ and $\nabla_{\rvx_t} \log p(\rvx_t, t \mid \rvx_0, 0)$
    are clearly distinct concepts,
    they are nearly equivalent in practical situations that the data space is high-dimensional
    and the data are distributed in a small subset (low-dimensional manifold) of the space.
    That is, in such a case, the integrated gradient  $\nabla_{\rvx_t} \log p(\rvx_t, t)$, given by
    \begingroup
    \allowdisplaybreaks
    \begin{align}
        \nabla_{\rvx_t} \log p(\rvx_t, t)
        &= \nabla_{\rvx_t} \log \int p(\rvx_t, t \mid \rvx_0 , 0) p(\rvx_0, 0) d\rvx_0 \nonumber \\
        &= \frac{\nabla_{\rvx_t} \int p(\rvx_t, t \mid \rvx_0 , 0) p(\rvx_0, 0) d\rvx_0}
            {\int p(\rvx_t, t \mid \rvx_0 , 0) p(\rvx_0, 0) d\rvx_0} \nonumber \\
        &
        = \frac{\mathbb{E}_{p(\rvx_0)} \left[
                \nabla_{\rvx_t} p(\rvx_t, t \mid \rvx_0 , 0)
                \right]}
            {\mathbb{E}_{p(\rvx_0)}\left[
                p(\rvx_t, t \mid \rvx_0 , 0)
                \right]} \nonumber \\
        &= \frac{\mathbb{E}_{p(\rvx_0)} \left[
                -\frac{\rvx_t - \sqrt{1 - \nu_t}\rvx_0}{\nu_t}
                p(\rvx_t, t \mid \rvx_0 , 0)
                \right]}
            {\mathbb{E}_{p(\rvx_0)}\left[
                p(\rvx_t, t \mid \rvx_0 , 0)
                \right]} \nonumber \\
        &= \mathbb{E}_{p(\rvx_0)} \Bigg[
                \underbrace{
                    -\frac{\rvx_t - \sqrt{1 - \nu_t}\rvx_0}{\nu_t}
                }_{=\nabla_{\rvx_t}\log p(\rvx_t, t \mid \rvx_0, 0)}
                \underbrace{
                    \frac{
                        e^{-\|\rvx_t - \sqrt{1 - \nu_t}\rvx_0\|^2/2\nu_t}
                    }
                    {\mathbb{E}_{p(\rvx_0)}[
                        e^{-\|\rvx_t - \sqrt{1 - \nu_t}\rvx_0\|^2/2\nu_t}
                        ]}
                }_{\eqqcolon q(\rvx_0 \mid \rvx_t)}
                \Bigg],\label{label:040}
    \end{align}
    \endgroup
    almost always has a single point approximation
    which is written as $\nabla_{\rvx_t} \log p(\rvx_t, t \mid \rvx_0^{(i)}, 0)$,
    where $\rvx_0^{(i)}$ is a certain point on the data manifold.
    There are two phases depending on the noise scale $\bar\alpha_t = 1 - \nu_t$,
    and both phases have different reasons for the validity of the approximation.

\subsection{Phase (1): Anyone can be a representative $(\bar{\alpha}_t = 1 - \nu_t \sim 0)$}
\begin{figure}[!t]
    \centering
    \includegraphics[width=0.7\linewidth]{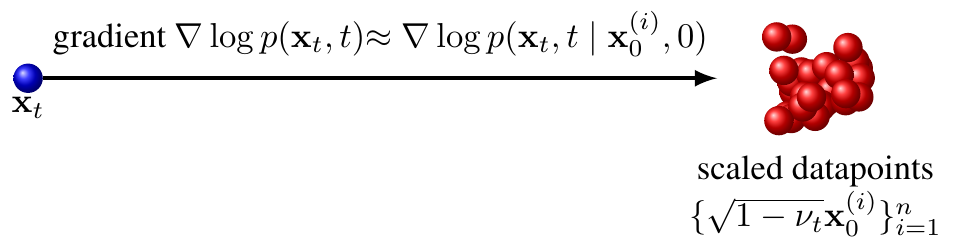}
    \caption{
        Intuitive reasons why a single point approximation is valid.
        The case where $1 - \nu_t \approx 0$}
    \label{label:041}
\end{figure}

    If $\rvx_t$ is far from any of $\{\sqrt{1 - \nu_t} \rvx_0^{(i)} \mid \rvx_0^{(i)} \sim p(\rvx_0)\}$,
    the gradients from each scaled data points are almost the same.
    Therefore, the integrated one viz.\ $\nabla \log p(\rvx_t, t)$ can be approximated by
    any of $\nabla \log p(\rvx_t, t \mid \rvx_0^{(i)}, 0)$ for any $\rvx_0^{(i)} \sim p(\rvx_0)$.
    \figref{label:041} intuitively illustrates the reason.

    More quantitatively,
    noting that the weight function $q(\cdot)$ satisfies $\mathbb{E}_{p(\rvx_0)}[q(\rvx_0 \mid \rvx_t)] = 1$,
    the $L_2$ distance between $\nabla \log p(\rvx_t, t)$ and
    $\nabla \log p(\rvx_t, t \mid \rvx_0^{(i)}, 0)$ is
    bounded above as follows,
    \begingroup
    \allowdisplaybreaks
    \begin{align}
        &\|\nabla \log p(\rvx_t, t) - \nabla \log p(\rvx_t, t \mid \rvx_0^{(i)}, 0)\|_2^2
        \nonumber\\
        &=\left\|
            \mathbb{E}_{p(\rvx_0)}\left[\frac{\rvx_t - \sqrt{1 - \nu_t}\rvx_0}{\nu_t} q(\rvx_0 \mid \rvx_t)\right]
            - \frac{\rvx_t - \sqrt{1 - \nu_t}\rvx_0^{(i)}}{\nu_t}
        \right\|_2^2 \nonumber \\
        &=
        \left\|
            \mathbb{E}_{p(\rvx_0)}\left[
                \left(
                \frac{\rvx_t - \sqrt{1 - \nu_t}\rvx_0}{\nu_t}
                - \frac{\rvx_t - \sqrt{1 - \nu_t}\rvx_0^{(i)}}{\nu_t}
                \right) q(\rvx_0 \mid \rvx_t)
            \right]
        \right\|_2^2 \nonumber \\
        &=
        \frac{1 - \nu_t}{\nu_t^2}
        \left\|
            \mathbb{E}_{p(\rvx_0)}\left[
                 (\rvx_0 - \rvx_0^{(i)})
                q(\rvx_0 \mid \rvx_t)
            \right]
        \right\|_2^2 \nonumber \\
        & \le
        \frac{1 - \nu_t}{\nu_t^2}
            \mathbb{E}_{p(\rvx_0)}\left[
                \|\rvx_0 - \rvx_0^{(i)}\|_2^2
                q(\rvx_0 \mid \rvx_t)
            \right]
        \quad(\because \text{Jensen's inequality}) \nonumber \\
        & \le
        \frac{1 - \nu_t}{\nu_t^2}
        \left(
            \max_{\rvx_0}
                \|\rvx_0 - {\rvx}_0^{(i)}\|_2^2
            \mathbb{E}_{p(\rvx_0)}\left[
                q(\rvx_0 \mid \rvx_t)
            \right]
        \right)  \nonumber \\
        & =
        \frac{1 - \nu_t}{\nu_t^2}
        \left(
            \max_{\rvx_0}
                \|\rvx_0 - {\rvx}_0^{(i)}\|_2^2
        \right) \nonumber \\
        &\le
            \frac{4(1 - \nu_t)R_{\mathcal{M}}^2}{\nu_t^2},
    \end{align}
    \endgroup
    where $R_{\mathcal{M}}$ is the radius of the smallest ball that covers the data manifold $\mathcal{M}$.
    The radius $R_{\mathcal{M}}$ will be a finite constant in most practical scenarios we are interested in.
    For example, if the data space is $d$-dim box, i.e., $\mathcal{M} \subset [0,1]^d$, then
    $R_{\mathcal{M}}$ is bounded above by $R_{\mathcal{M}} \le \sqrt{d}/2$.

    Noting that $\|\nabla \log p(\rvx_t, t \mid \rvx_0^{(i)}, 0) \|_2 \approx \sqrt{d/\nu_t}$
    (this is actually a random variable that follows the $\chi$-distribution\footnote{
        Given Gaussian random variables $X_i \sim \mathcal{N}(0, 1)$,
        then the squared sum of them $\sum_{i=1}^d X_i^2$ follows the $\chi^2$-distribution with $d$ degrees of freedom.
        It is well known that $\chi^2$ distribution
        converges to $\mathcal{N}(d, 2d)$ as $d$ increases.
        Therefore, If $d$ is sufficiently large, $\|\rvx\|_2 \coloneqq (\sum_{i=1}^d X_i^2)^{1/2}
        \approx \sqrt{d \pm \sqrt{2d}}
        \approx \sqrt{d} \pm \sqrt{1/2}$.
        Thus the $L_2$ norm of $d$-dim Gaussian variable is approximated by $\sqrt{d}$,
        and the error is as small as the scale of $\sqrt{1/2}$.
        It implies that Gaussian distribution looks like a sphere in high-dimensional space,
        contrary to our low-dimensional intuition.
        Indeed, the following inequality holds~\citep[Lem.~1]{laurent2000adaptive}:
        $p((1 - 2\sqrt{y/d})^{1/2} \le \|\rvx\|_2/\sqrt{d} \le (1 + 2\sqrt{y/d} + 2y/d)^{1/2}) \ge 1 - 2e^{-y}$.
    }),
    the relative $L_2$ error is evaluated as follows,
    \begin{equation}
    \text{(relative $L_2$ error)} =
        \frac{\|\nabla \log p(\rvx_t, t) - \nabla \log p(\rvx_t, t \mid \rvx_0^{(i)}, 0) \|_2}
        {\|\nabla \log p(\rvx_t, t \mid \rvx_0^{(i)}, 0) \|_2}
        \lessapprox \sqrt{\frac{1 - \nu_t}{\nu_t}}. \label{label:042}
    \end{equation}
    Similarly, the cosine similarity is evaluated as
    \begin{equation}
        \text{cossim}(\nabla \log p(\rvx_t, t), \nabla \log p(\rvx_t, t \mid \rvx_0^{(i)}, 0))
        \gtrapprox 1 - \frac{1}{2}\left(1 + \sqrt{\frac{1 - \nu_t}{\nu_t}}\right) \frac{1 - \nu_t}{\nu_t}.
        \label{label:043}
    \end{equation}
    These bounds suggest that the relative error between the true gradient and the single point approximation
    approaches to $0$, and the cosine similarity goes to $1$, as \mbox{$\nu_t \to 1$} \mbox{$(\bar\alpha_t \to 0)$}.
    That is, the single point approximation is largely valid when the noise level $\nu_t$ is high.

    \paragraph{Proofs for some facts used in this section}$\quad$

    \myparagraph{Jensen's inequality} A sketchy proof of Jensen's inequality for $L_2$-norm is as follows:
    \begin{multline}
    \left\|\sum_n w_n \rvx^{(n)} \right\|_2^2
    = \sum_i {\left[ \sum_n w_n \rvx^{(n)}\right]_i}^2
    = \sum_i \left( \sum_n w_n x_i^{(n)} \right)^2 \\
    \le \sum_i \sum_n w_n (x_i^{(n)})^2
    = \sum_n w_n \sum_i (x_i^{(n)})^2
    = \sum_n w_n \|\rvx^{(n)}\|_2^2.
    \end{multline}

    \myparagraph{Cosine similarity} Given the relative distance between $\rvx$ and $\rvy$
    as $\|\rvx - \rvy\|_2/\|\rvx\|_2 = r$ (for simplicity, let $r \ll 1$)
    then the cosine similarity between the vectors are evaluated as follows,
    \begin{align}
        \text{cossim}(\rvx, \rvy)
        &= \frac{\langle \rvx, \rvy \rangle}{\|\rvx\|_2\|\rvy\|_2}
        = \frac{\|\rvx\|_2}{\|\rvy\|_2}  \cdot \frac{\langle \rvx, \rvy \rangle}{\|\rvx\|_2^2}
        = \frac{1}{2}\frac{\|\rvx\|_2}{\|\rvy\|_2} \cdot \frac{\|\rvx\|_2^2+ \|\rvy\|_2^2 -\|\rvx - \rvy\|_2^2}{\|\rvx\|_2^2} \nonumber \\
        &=\frac{1}{2} \frac{\|\rvx\|_2}{\|\rvy\|_2} \cdot \left(1 + \frac{\|\rvy\|_2^2}{\|\rvx\|_2^2} -r^2 \right)
        = \frac{1}{2} \left(\frac{\|\rvx\|_2}{\|\rvy\|_2} + \frac{\|\rvy\|_2}{\|\rvx\|_2} - \frac{\|\rvx\|_2}{\|\rvy\|_2} r^2 \right).
    \end{align}
    Noting that the triangle inequality implies that the $L_2$ norm of $\rvy$ is bounded as follows,
    \begin{equation}
        (1 - r) \|\rvx\|_2 \le \|\rvy\|_2 \le (1 + r) \|\rvx\|_2,
    \end{equation}
    we can put $\|\rvy\|_2/\|\rvx\|_2 = 1 + \delta, (0 < |\delta| < r \ll 1)$,
    and can evaluate the cosine similarity as follows,
    \begin{align}
        \text{cossim}(\rvx, \rvy)
        &= \frac{1}{2} \left(\frac{1}{1 + \delta} + (1 + \delta) - \frac{1}{1 + \delta} r^2 \right) \nonumber \\
        &\approx \frac{1}{2}\left((1 - \delta) + (1 + \delta) - (1 - \delta)r^2 \right)
        > 1 - \frac{1}{2}(1 + r)r^2.
    \end{align}

\begin{figure}[!t]
    \centering
    \begin{subfigure}[t]{0.49\linewidth}
        \centering
        \includegraphics[width=0.95\linewidth]{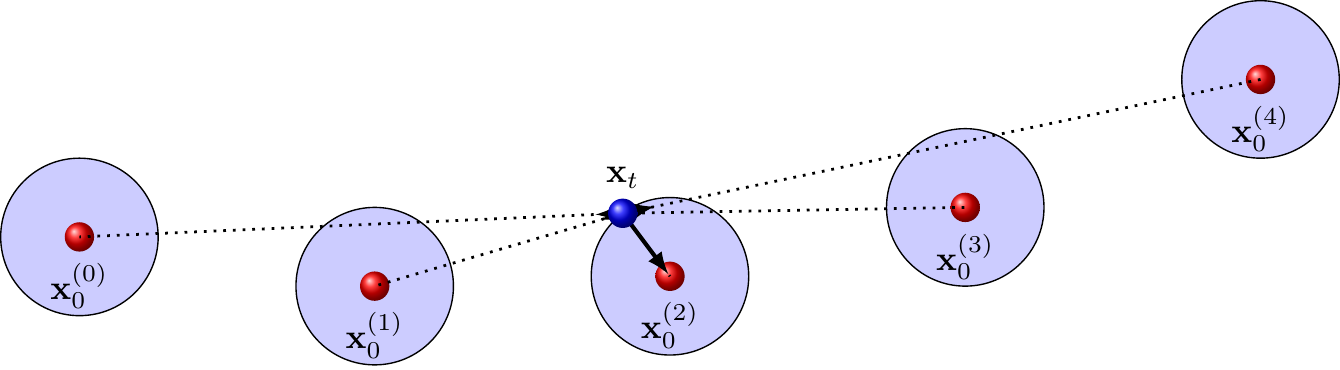}
        \caption{Discrete case}\label{label:044}
    \end{subfigure}
    \begin{subfigure}[t]{0.49\linewidth}
        \centering
        \includegraphics[width=0.95\linewidth]{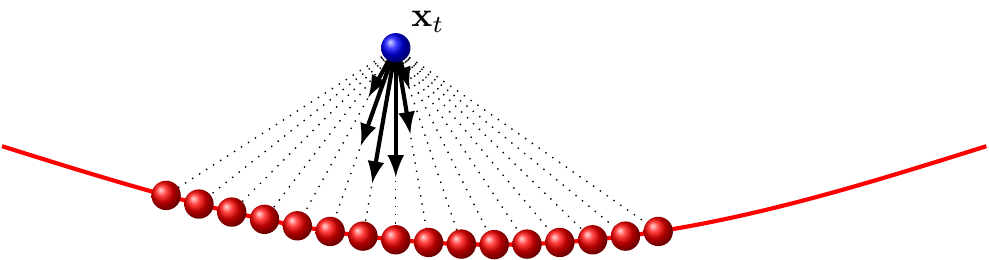}
        \caption{Continuous case}\label{label:045}
    \end{subfigure}
    \caption{
    Intuitive reasons why a single point approximation is valid.
    The case where $1 - \nu_t \approx 1$}
    \label{label:046}
\end{figure}
\subsection{Phase (2): Winner takes all $(\bar{\alpha}_t = 1 - \nu_t \gg 0)$}
    The above bounds suggest that the single point approximation
    is valid only when the noise level $\nu_t$ is high ($\nu_t \approx 1$).
    However, we can also show that the approximation is also valid
    because of another reason when the noise level $\nu_t$ is low ($\nu_t \approx 0$).

    If $p(\rvx_0)$ is a discrete distribution, and if $\nu_t \approx 0$,
    the weight factor $q (\rvx_0^{(i)} \mid \rvx_t)$ in \eqref{label:040} is almost certainly a ``one-hot vector''
    such that
    \begin{equation}
        q (\rvx_0^{(i)} \mid \rvx_t) \approx 1,\quad \text{and}\quad q (\rvx_0^{(j)} \mid \rvx_t) \approx 0~(j \ne i).
    \end{equation}
    That is, the entropy of $q (\rvx_0 \mid \rvx_t)$ is nearly zero (see \figref{label:049}).
    It is almost certain (prob $\approx$ 1),
    since the converse is quite rare;
    i.e., there exists another data point $\rvx_0^{(j)}$ near $\rvx_t \sim p(\rvx_t, t \mid \rvx_0^{(j)}, 0)$
    in such a situation\footnote{
        Let us evaluate the probability more qualitatively using a toy model.
        Let us denote a $k$-dim ball of radius $r$ by $B^k(r)$,
        and let $D_i$ be the $L_2$ distance between $\rvx_t$ and $\sqrt{1 - \nu_t}\rvx_0^{(i)}$,
        i.e.,
        $
        D_i \coloneqq \|\rvx_t - \sqrt{1 - \nu_t}\rvx_0^{(i)}\|_2 \approx \sqrt{\nu_td}.
        $
        The question is whether there exists another data point $\rvx_0^{(j)} (j \ne i)$
        in the discrete data distribution $\mathcal{D} = \{\rvx_0^{(i)}\}_{i}$
        such that $D_j < D_i$.
        In other words, whether there exists another data point in $B^d(\sqrt{\nu_td})$
        centered at $\sqrt{1 - \nu_t}\rvx_0^{(i)}$.
        If it is the case, $q(\rvx_0^{(j)}\mid \rvx_t)$
        has significantly large value compared to $q(\rvx_0^{(i)} \mid \rvx_t)$,
        and it will have strong effects on the result of integration \eqref{label:040}.
        Now let us show that it is unlikely in a high-dimensional data space.
        As it is difficult evaluating the probability for general cases,
        let us consider a toy model
        that the all the $n$ points are uniformly distributed in a $k$-dim ball $(k \ll d)$.
        That is, we assume that the data manifold is $\mathcal{M} = B^{k}(\sqrt{k})$
        which satisfies $\mathcal{D} \subset \mathcal{M} \subset \mathbb{R}^d$.
        Then, the probability we are interested in is roughly evaluated as
        \begin{equation}
            p(\exists j \ne i, D_j < D_i)
            \approx
            1 - \left(
                1 -
                \frac{
                    \mathrm{vol}[B^k(\sqrt{ \nu_t k})]
                }{
                    \mathrm{vol}[\sqrt{1 - \nu_t}\mathcal{M}]
                }
            \right)^{n}=
            1 - \left(
                1 -
                \frac{
                    \mathrm{vol}[B^k(\sqrt{ \nu_t k})]
                }{
                    \mathrm{vol}[B^k(\sqrt{(1 - \nu_t)k})]
                }
            \right)^{n}
            . \nonumber
        \end{equation}
        Remembering that the volume of $k$-dim ball is given by
        $\mathrm{vol}[B^k(r)] = \frac{\pi^{k/2}}{\Gamma(k/2 + 1)} r^k,$
        the probability is approximated as follows,
        \begin{equation}
            \text{r.h.s}= 1 - \left(1 - \left(\frac{\nu_t k}{(1 - \nu_t) k}\right)^{k/2}\right)^n
            \approx n\left(\frac{\nu_t}{1 - \nu_t}\right)^{k/2},
            \nonumber\quad\text{if~~} \nu_t \approx 0, k\gg 1.
        \end{equation}
        Thus, when $\nu_t$ is sufficiently small and the data manifold is sufficiently high-dimensional,
        it is almost unlikely that there exists such a point $\rvx_0^{(j)}~(j \ne i)$ unless $n$ is very large.
    }.
    \figref{label:044} intuitively shows the reason.

    If $p(\rvx_0)$ is a continuous distribution,
    very small area (almost a single point) of the data manifold contributes to the
    integration \eqref{label:040}.
    \figref{label:045} intuitively shows the situation.
    The weight function will become nearly the delta function,
    and the integration will be almost the same as the single point approximation,
    \begin{equation}
        q (\rvx_0 \mid \rvx_t) \approx \delta(\rvx_0 - \rvx_0^*).
    \end{equation}
    where $\rvx_0^*$ is a certain point close to the perpendicular projection of $\rvx_t$ on the data manifold.
    These are related to the well-known formulae
    $\varepsilon\log\sum\exp(\cdot/\varepsilon)$ and $\varepsilon\log\int\exp(\cdot/\varepsilon)$ goes to $\max(\cdot)$ when $\varepsilon \to 0$.
    Naturally, this is more likely to be true for high-dimensional space and is
    expected to break down for low-dimensional toy data.

    Now let us elaborate the above discussion.
    As the scale we are interested in now is very small,
    the data manifold is approximated as a flat $k$-dim space.
    In addition, as $q(\cdot \mid \rvx_t)$ is a Gaussian function,
    it decays rapidly with distance from the center (perpendicular projection of $\rvx_t \in \mathbb{R}^d$ to the $k$-dim subspace).
    For this reason, only a small ball around the perpendicular foot of $\rvx_t$ contributes to the integration.
    As the majority of $k$-dim Gaussian variables of variance $\sigma^2 \mathbf{I}$ are
    distributed on a sphere of radius $\sqrt{k}\sigma$,
    it would be sufficient to consider a ball of radius $(\sqrt{k}+2)\sqrt{\nu_t}$.

    Now the integration \eqref{label:040} is approximated as follows,
    \begin{multline}
        \text{\eqref{label:040}} \approx
        \int_{\mathcal{B}}
        \left[
        \nabla_{\rvx_t}\log p(\rvx_t, t \mid \rvx_0, 0) q(\rvx_0 \mid \rvx_t) p(\rvx_0)
        \right]
        d\rvx_0,\\
        \text{where~}
        \mathcal{B} = \{\rvx_0 \mid \|\sqrt{1 - \nu_t} \rvx_0 - \rvx_t\|_2 < (\sqrt{k}+2)\sqrt{\nu_t} \}.
    \end{multline}
    As $\mathcal{B}$ is small,
    the gradient vector $\nabla_{\rvx_t}\log p(\rvx_t, t \mid \rvx_0, 0)$ is almost constant in this region.
    Quantitatively, let $\rvx_0^{(1)}, \rvx_0^{(2)} \in \mathcal{B}$,
    then the distance between these points is evidently bounded above by
    $\|\rvx_0^{(1)} - \rvx_0^{(2)}\|_2 \le 2(\sqrt{k}+2)\sqrt{\nu_t}$,
    which implies
    \begingroup
    \allowdisplaybreaks
    \begin{align}
        &\left\|
        \nabla \log p(\rvx_t, t \mid \rvx_0^{(1)}, 0)
        -
        \nabla \log p(\rvx_t, t \mid \rvx_0^{(2)}, 0)
        \right\|_2 \nonumber \\
        =&
        \left\|
        \frac{\rvx_t - \sqrt{1-\nu_t}\rvx_0^{(1)}}{\nu_t}
        -
        \frac{\rvx_t - \sqrt{1-\nu_t}\rvx_0^{(2)}}{\nu_t}
        \right\|_2 \nonumber \\
        =&\left\|\frac{
                \sqrt{1-\nu} (\rvx_0^{(1)} - \rvx_0^{(2)} )
            }{\nu_t}\right\|_2
        \le \frac{\sqrt{1 - \nu_t}}{\nu_t} \cdot 2(\sqrt{k}+2)\sqrt{\nu_t} \nonumber \\
        =& 2(\sqrt{k}+2)\sqrt{\frac{1 -\nu_t}{\nu_t}}.
    \end{align}
    \endgroup
    Noting that $\|\nabla \log p(\rvx_t, t \mid \rvx_0, 0)\|_2 \approx \sqrt{d/\nu_t}$,
    the relative $L_2$ error between two gradients is bounded above by
    \begin{equation}
        \frac{
            \|
        \nabla \log p(\rvx_t, t \mid \rvx_0^{(1)}, 0)
        -
        \nabla \log p(\rvx_t, t \mid \rvx_0^{(2)}, 0)
            \|_2
        }{
            \|\nabla \log p(\rvx_t, t \mid \rvx_0^{(2)}, 0)\|_2
        }
        \lessapprox 2\frac{(\sqrt{k}+2)\sqrt{1 - \nu_t}}{\sqrt{d}} \lessapprox 2\sqrt{\frac{k}{d}}.
    \end{equation}
    By integrating both sides \footnote{
        We used the following relation:
        $\|\mathbb{E}_\rvx[f(\rvx)] - \rvy\|_2 =
        \|\mathbb{E}_\rvx[f(\rvx) - \rvy]\|_2 \le
        \mathbb{E}_\rvx[\|f(\rvx) - \rvy\|_2]$.
    }
    w.r.t.\ $\rvx_0^{(1)}$, i.e.
    $\int
    (\cdot)
    q(\rvx_0^{(1)} \mid \rvx_t) p(\rvx_0^{(1)}) d \rvx_0^{(1)}
    $,
    we obtain
    the following inequality for any point $\rvx_0^*\in \mathcal{B}$,
    \begin{equation}
        \frac{
            \|
        \nabla \log p(\rvx_t, t)
        -
        \nabla \log p(\rvx_t, t \mid \rvx_0^*, 0)
            \|_2
        }{
            \|\nabla \log p(\rvx_t, t \mid \rvx_0^*, 0)\|_2
        }
        \lessapprox 2\sqrt{\frac{k}{d}}.
    \end{equation}
    Now, let us recall that a noised data $\rvx_t \sim p(\rvx_t, t)$
    is generated in the following procedure.
    \begin{enumerate}
        \item Draw a data $\rvx_0^{(i)}$ from $p(\rvx_0, 0)$.
        \item Draw a Gaussian $\rvx_t \sim p(\rvx_t, t \mid \rvx_0^{(i)}, 0) = \mathcal{N}(\sqrt{1 - \nu_t}\rvx_0^{(i)}, \nu_t\mathbf{I}_d)$.
    \end{enumerate}
    Here, the data point $\rvx_0^{(i)} \in \mathcal{M}$ is in the ball $\mathcal{B}$,
    because the distance between $\sqrt{1 - \nu_t}\rvx_0^{(i)}$ and the perpendicular foot of $\rvx_t$ is about $\sqrt{\nu_t k}$,
    which is clearly smaller than $(\sqrt{k} + 2)\sqrt{\nu_t}$.
    Thus we can write the following inequality.
    \begin{equation}
        \frac{
            \|
        \nabla \log p(\rvx_t, t)
        -
        \nabla \log p(\rvx_t, t \mid \rvx_0^{(i)}, 0)
            \|_2
        }{
            \|\nabla \log p(\rvx_t, t \mid \rvx_0^{(i)}, 0)\|_2
        }
        \lessapprox 2\sqrt{\frac{k}{d}},
    \end{equation}
    and we may conclude that the following approximation is largely valid if $\nu_t \approx 0, k \ll d$,
    \begin{equation}
        \nabla \log p(\rvx_t, t) \approx
        \nabla \log p(\rvx_t, t \mid \rvx_0^{(i)}, 0).
    \end{equation}

\vfill
\newpage
    \subsection{Plots using Real Data}
    Let us empirically validate that single point approximation using real data,
    MNIST~\citep{lecun2010mnist} 60,000 samples and CIFAR-10~\citep{Krizhevsky09learningmultiple} 50,000 samples.
    In specific, we evaluated following three metrics.
    \begin{itemize}
        \item Relative $L_2$ error between $\nabla \log p(\rvx_t, t)$ and $\nabla \log p(\rvx_t, t \mid \rvx_0, 0)$,
        \item Cosine similarity between $\nabla \log p(\rvx_t, t)$ and $\nabla \log p(\rvx_t, t \mid \rvx_0, 0)$,
        \item Entropy of $q(\rvx_0 \mid \rvx_t)$.
    \end{itemize}
    \figref{label:047} shows the relative $L_2$ distance,
    for both datasets.
    \figref{label:048}  similarly
    show the distribution (random 3000 trials) of the cosine similarity,
    and \figref{label:049} shows the entropy.
    Dashed curves indicate the
    bounds evaluated in \eqref{label:042} and \eqref{label:043}.

    These figures show that the range of intermediate region between Phase (1) and Phase (2) will
    not have impact in practical situations
    since we do not evaluate the neural network $\scorefn(\cdot, \cdot)$ in this range so many times
    (i.e., $\bar\alpha_t \sim 10^{-3}$ to $10^{-1} \Leftrightarrow \nu_t \sim 0.999$ to $0.9$).
    Moreover, the approximation accuracy is still very high even in this region.
    Furthermore, although MNIST and CIFAR-10 are quite ``low-dimensional'' for real-world images,
    approximations are established with such high accuracy.
    Therefore, it is expected to be established with higher accuracy for more realistic images.

    \subsection{Summary of This Section}
    Thus, we can assume that the single point approximation almost always holds practically.
    \begin{equation}
    -\frac{\scorefn(\rvx_t, t)}{\sqrt{\nu_t}}
        \stackrel{\text{model}}{\approx} \nabla_{\rvx_t} \log p(\rvx_t, t)
        \stackrel{\text{almost equal}}{\approx} \nabla_{\rvx_t} \log p(\rvx_t, t \mid \rvx_0^{(i)}, 0)
        = - \frac{\rvx_t - \sqrt{1 - \nu_t} \rvx_0^{(i)}}{\nu_t}.
        \nonumber
    \end{equation}
    Therefore, we may also expect that the similar approximation will be valid for their derivatives.
    Of course, such an expectation is mathematically incorrect;
    for example, let $g(x) = f(x) + \varepsilon\sin \omega x$, then
    the difference $g(x) - f(x) = \varepsilon \sin\omega x$ goes to zero as $\varepsilon \to 0$,
    but the difference of derivatives $g'(x) - f'(x) = \varepsilon \omega \cos\omega x$ does not if $\omega \to \infty$ faster than $1/\varepsilon$.
    In other words, the single point approximation does not necessarily imply the ideal derivative approximation,
    and if it is to be mathematically rigorous,
    it must be supported by other nontrivial knowledge on the data manifold.
    Conversely, if the approximation holds for the derivative,
    it would provide indirect evidence that such nontrivial logic holds
    that the derivative of error term is small in the region of interest,
    and would provide a better understanding of the fundamental properties of diffusion generative models.
    This nontrivial leap is the most important ``conjecture''
    made in this paper and should be examined more closely in the future.

    \vfill
    \newpage

    \begin{minipage}{\textwidth}
        \begin{figure}[H]
            \centering
            \begin{subfigure}[t]{0.49\linewidth}
                \centering
                \includegraphics[width=0.95\linewidth]{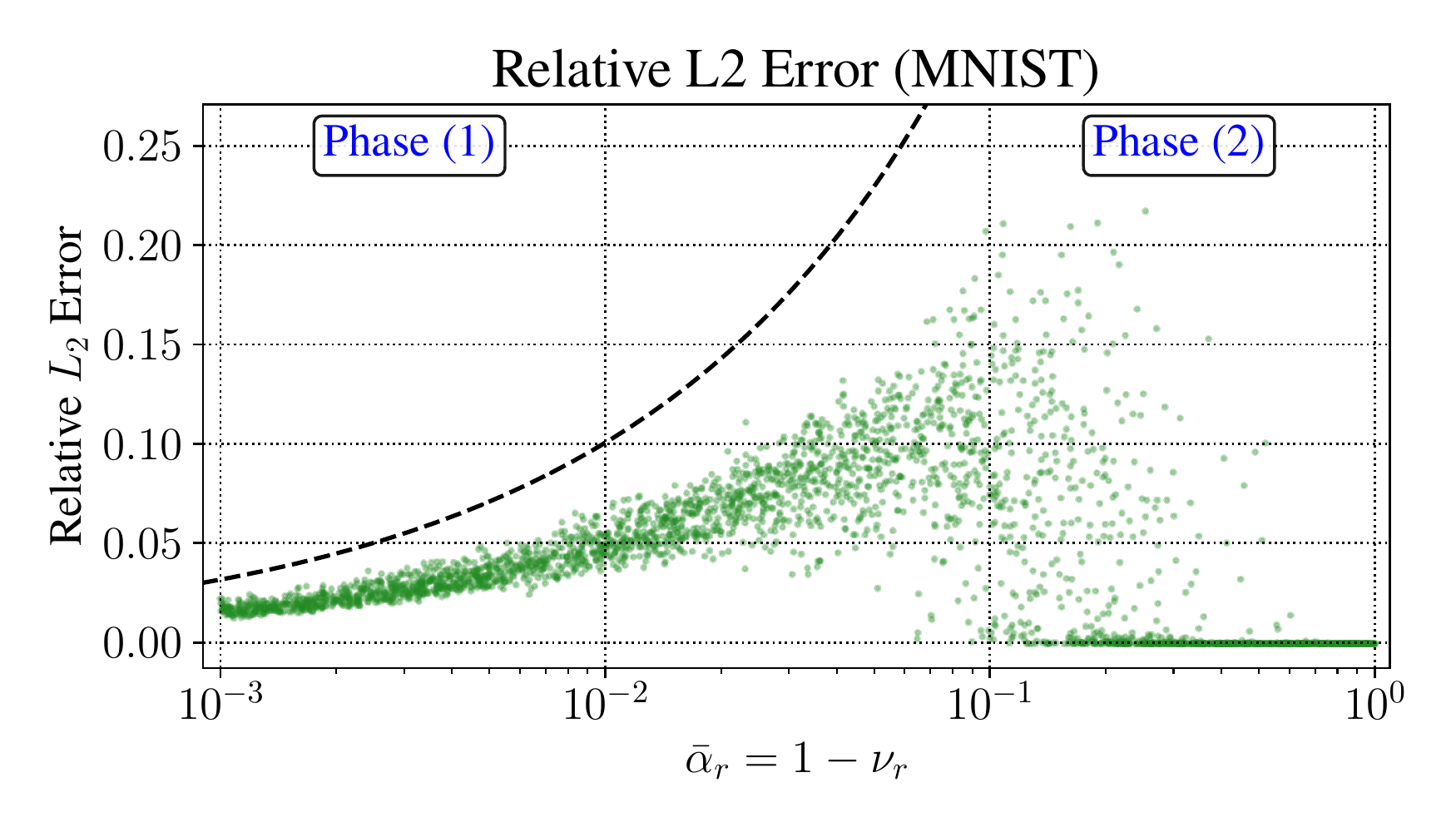}
                \caption{MNIST training data. 784-dim ($=28 \times 28$)}
            \end{subfigure}
            \begin{subfigure}[t]{0.49\linewidth}
                \centering
                \includegraphics[width=0.95\linewidth]{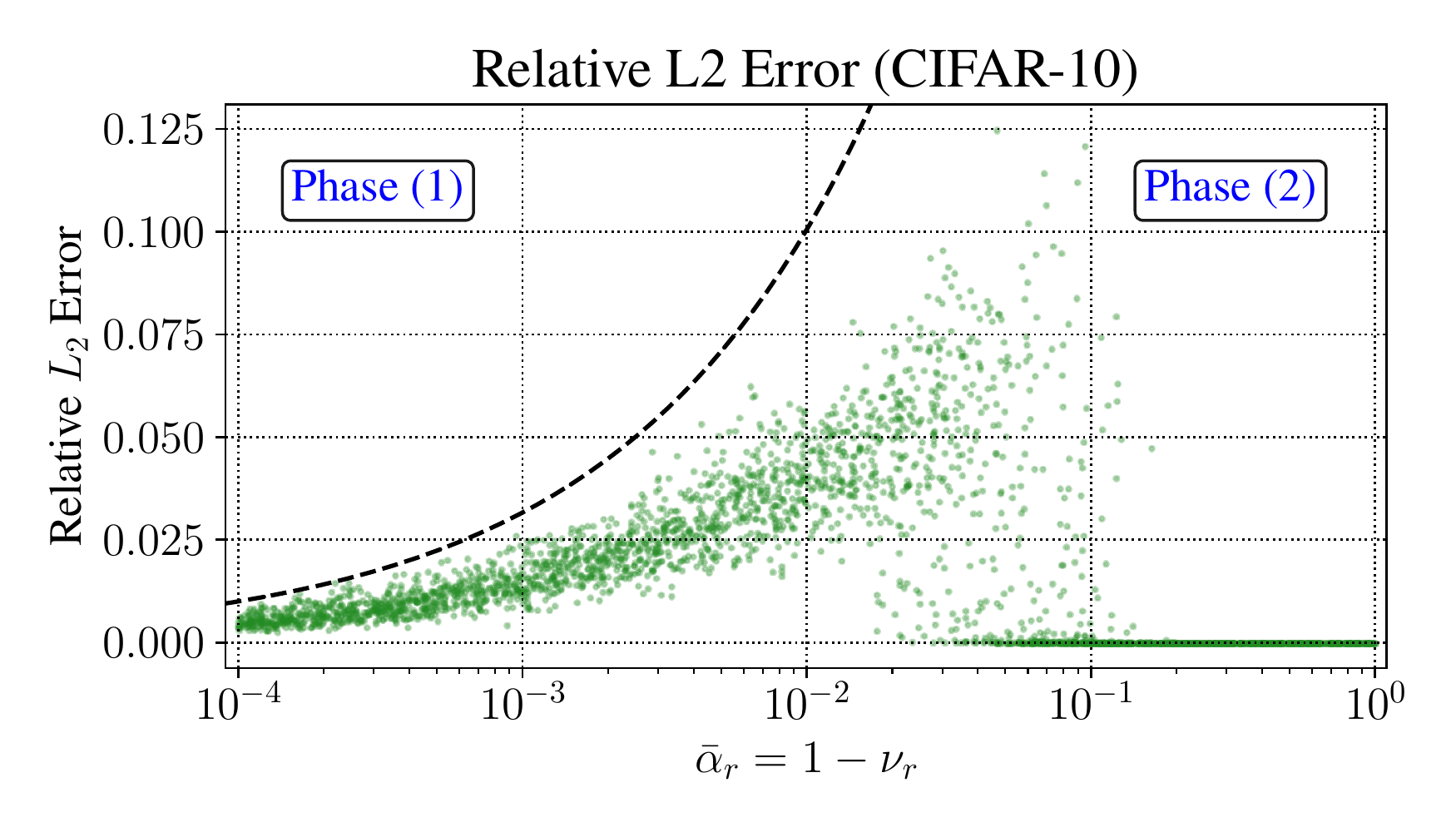}
                \caption{CIFAR10 data. 3072-dim ($=32 \times 32 \times 3$)}
            \end{subfigure}
            \caption{
                Relative $L_2$ distance between the true gradient and the single point approximation depending on the noise level $\nu_t$.
                Dashed curves indicate the upper bounds evaluated in \eqref{label:042}.
            }\label{label:047}%
            \vspace{1cm}
            \centering
            \begin{subfigure}[t]{0.49\linewidth}
                \centering
                \includegraphics[width=0.95\linewidth]{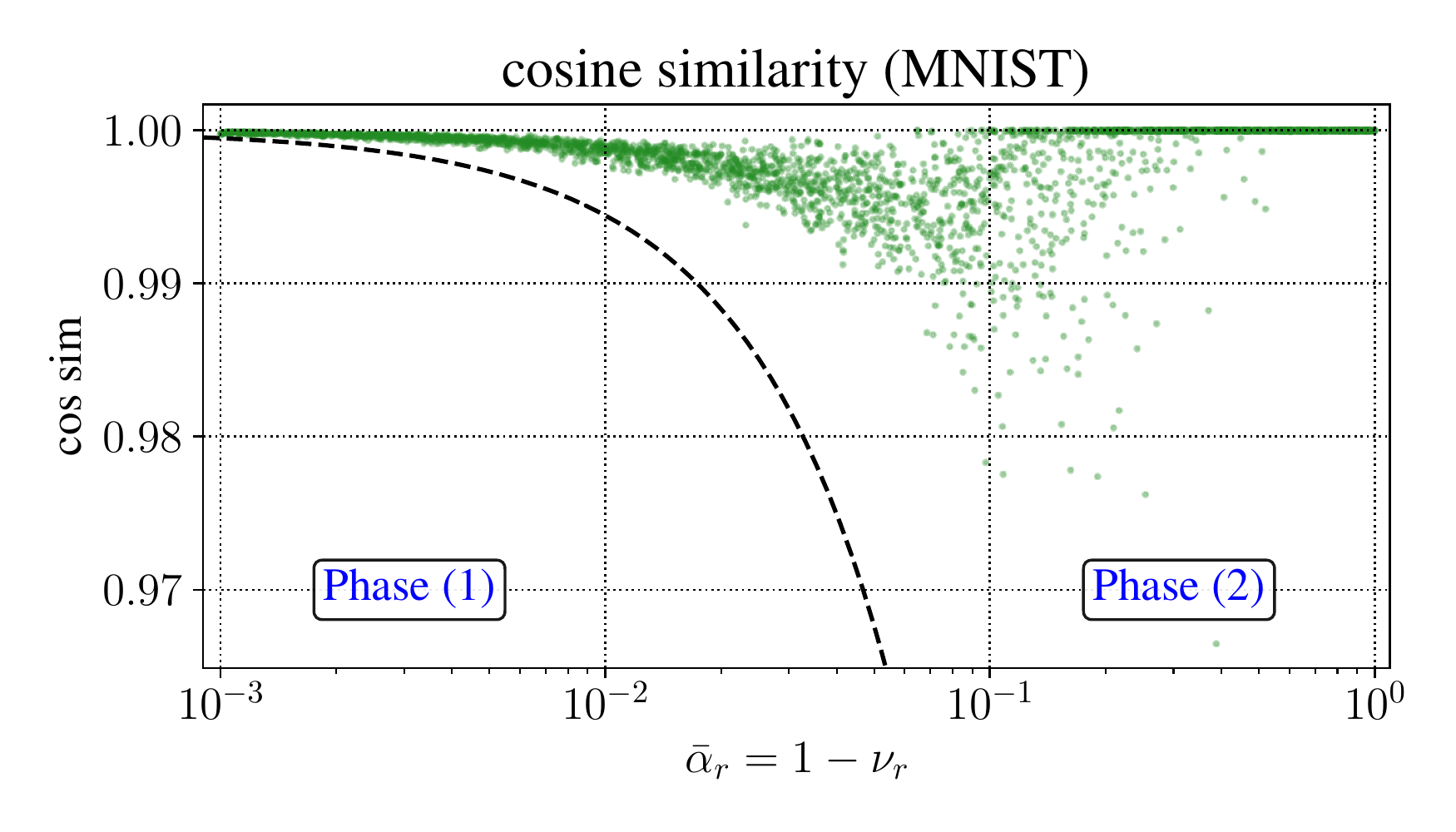}
                \caption{MNIST training data. 784-dim ($=28 \times 28$)}
            \end{subfigure}
            \begin{subfigure}[t]{0.49\linewidth}
                \centering
                \includegraphics[width=0.95\linewidth]{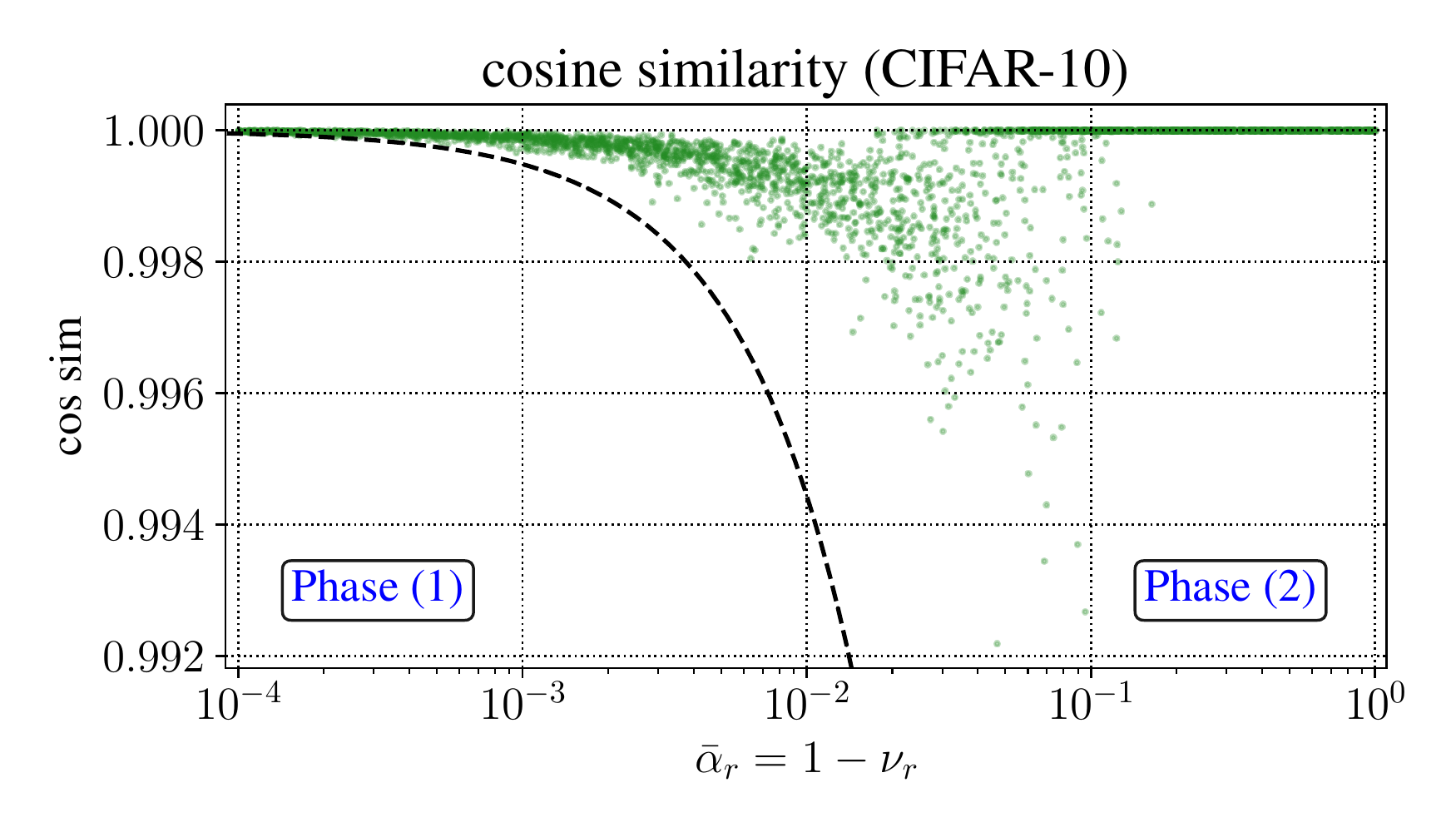}
                \caption{CIFAR10 data. 3072-dim ($=32 \times 32 \times 3$)}
            \end{subfigure}
            \caption{
                Cosine similarity between the true gradient and the single point approximation depending on the noise level $\nu_t$.
                Dashed curves indicate the lower bounds evaluated in \eqref{label:043}.
            }\label{label:048}%
            \vspace{1cm}
            \centering
            \begin{subfigure}[t]{0.49\linewidth}
                \centering
                \includegraphics[width=0.95\linewidth]{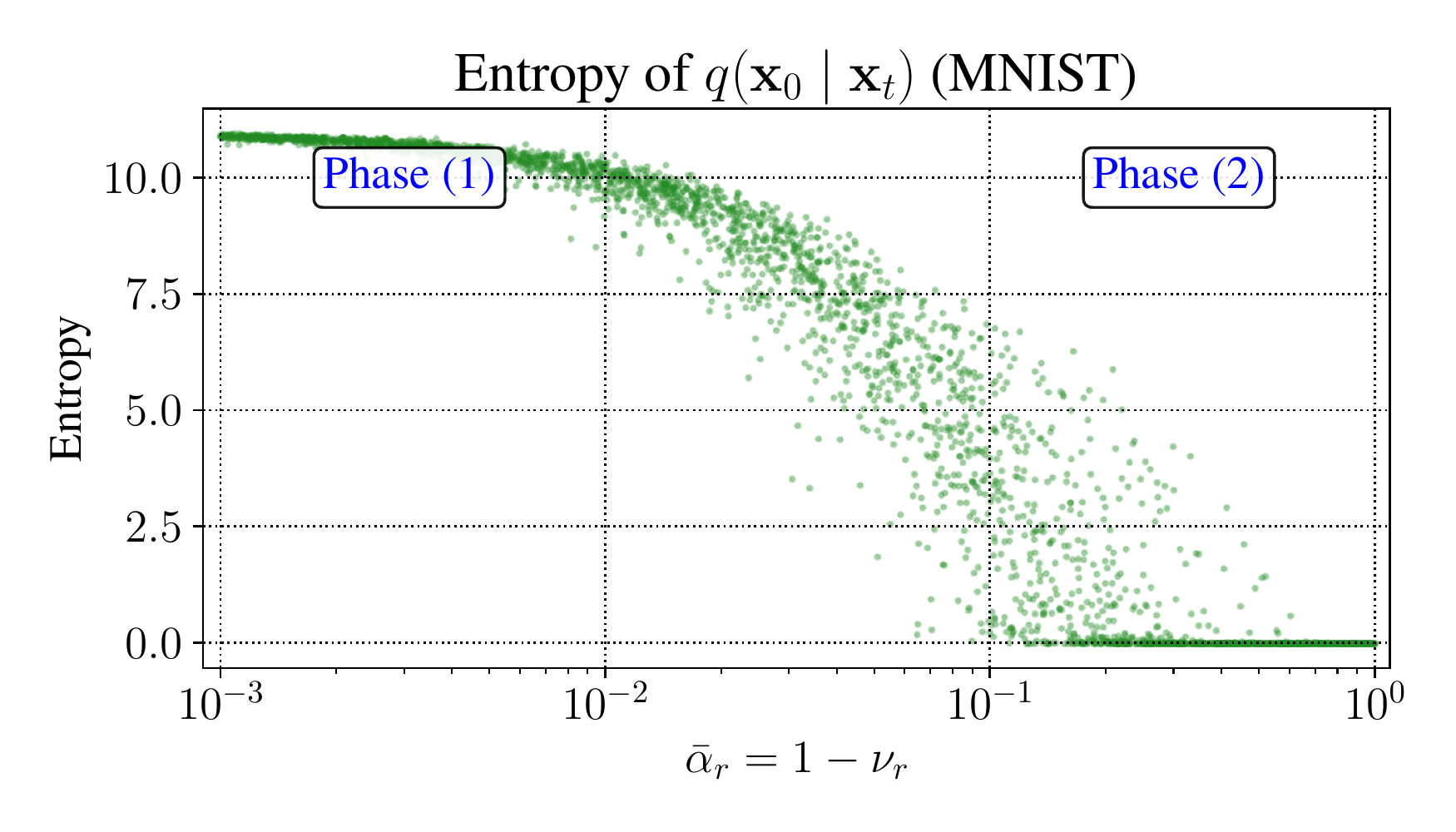}
                \caption{MNIST training data. 784-dim ($=28 \times 28$)}
            \end{subfigure}
            \begin{subfigure}[t]{0.49\linewidth}
                \centering
                \includegraphics[width=0.95\linewidth]{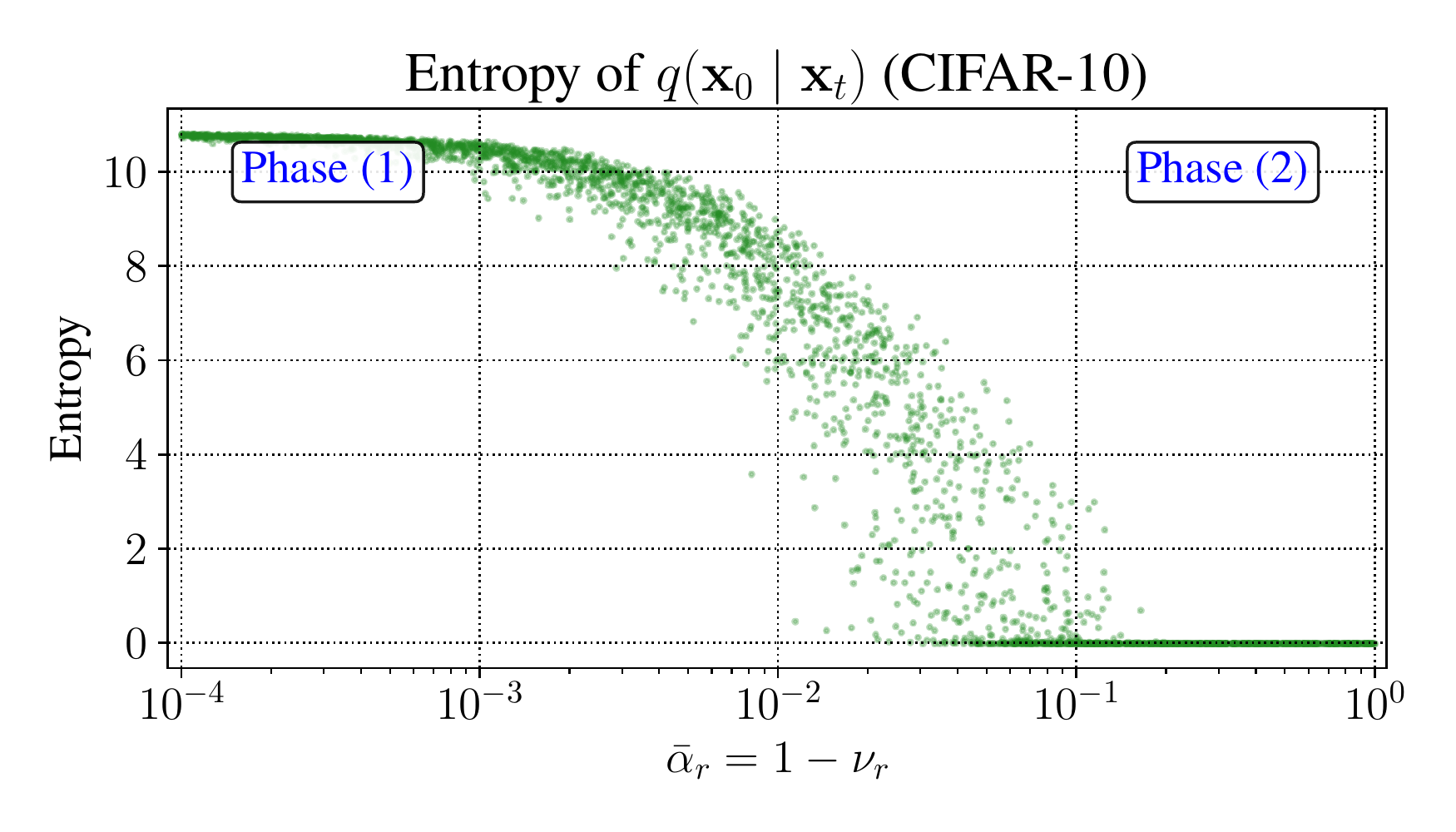}
                \caption{CIFAR10 data. 3072-dim ($=32 \times 32 \times 3$)}
            \end{subfigure}
            \caption{
                Entropy of $q(\rvx_0\mid\rvx_t)$ depending on the noise level $\nu_t$.
            }\label{label:049}%
        \end{figure}
        \end{minipage}

\vfill
\newpage
\section{Derivation of the ``Ideal Derivatives''}\label{label:050}

    Because of the discussion in \secref{label:039},
    the true score function
    $\nabla_{\rvx_t} \log p(\rvx_t, t)$
    is finely approximated by a single point approximation
    $\nabla_{\rvx_t} \log p(\rvx_t, t \mid \rvx_0, 0)$.
    Now we may also assume that the
    derivatives of both will also be close.

    In this paper, we are interested in the Taylor expansion
    of the following form (see also \secref{label:058}),
    \begin{equation}
        \bm{\psi}(\rvx_h, h) = \bm{\psi}(\rvx_0, 0)
        + \sum_{k = 1}^{\infty} \frac{h^k}{k!} \left(\diff{t} +
        \rva(\rvx_t, t)\cdot\nabla_{\rvx_t}\right)^k \bm{\psi}(\rvx_t, t) \bigg|_{t=0}.
    \end{equation}
    If the function $\bm{\psi}(\rvx_t, t)$ is separable in each dimension (i.e., $\diff{x_i} \psi_j = 0$ for $i \ne j$),
    the following relation holds,
    \begin{equation}
        \left( \rva(\rvx_t, t)\cdot\nabla_{\rvx_t}\right) \bm{\psi}(\rvx_t, t) = \rva(\rvx_t, t) \odot \nabla_{\rvx_t} \odot \bm{\psi}(\rvx_t, t),
    \end{equation}
    where $\odot$ is the element-wise product or operation.
    If $\rva(\rvx_t, t)$ is also separable in each dimension\footnote{
        In general, $(\rva \cdot \nabla)^2
            = (\sum_i a_i \diff{i})^2
            = (\sum_i a_i \diff{i})(\sum_j a_j \diff{j})
            = \sum_i a_i \sum_j (\diff{i}a_j  + a_j \diff{i}\diff{j})
        $.
        If $\rva$ is separable in each dimension, the $\diff{i}a_j (i \ne j)$ terms vanish, and
        $(\rva \cdot \nabla)^2 = \sum_i (a_i\diff{i}a_i+ \sum_j a_i a_j \diff{i}\diff{j})$.
        If the function $\bm{\psi}(\rvx_t, t)$ is separable in each dimension, then
        $(\rva \cdot \nabla)^2\psi_k
            = \sum_i (a_i\diff{i}a_i+ \sum_j a_i a_j \diff{i}\diff{j}) \psi_k
            = (a_k\diff{k}a_k+ a_k^2\diff[2]{k}) \psi_k
        $.
        Thus we can formally write $(\rva \cdot \nabla)^2 \bm{\psi}
            = (\rva\odot\nabla\odot\rva + \rva\odot\rva\odot\nabla\odot\nabla)\odot \bm{\psi}
            = \rva\odot (\nabla\odot \rva + \rva \odot \nabla\odot\nabla) \odot \bm{\psi}
            = \rva\odot \nabla \odot (\rva \odot \nabla) \odot \bm{\psi}
            = (\rva\odot \nabla \odot)^2 \bm{\psi} = (\rva \odot \diff{\rvx})^{2} \bm{\psi}
        $.
        (Note that the operator $(\rva \cdot \nabla)$ is scalar while $(\rva \odot \diff{\rvx})$ is $d$-dim vector.)
        We can similarly show
        $(\rva \cdot \nabla)^k \bm{\psi}= (\rva \odot \diff{\rvx})^{k}\bm{\psi}$ for $k \ge 3$.
    }
    the Taylor series is formally rewritten as follows,
    \begin{equation}
        \bm{\psi}(\rvx_t, t) = \bm{\psi}(\rvx_0, 0)
        + \sum_{k = 1}^{\infty} \frac{t^k}{k!} \left(\bm{1} \diff{t} + \rva(\rvx_t, t) \odot \diff{\rvx_t} \right)^k \bm{\psi}(\rvx_t, t) \bigg|_{t=0}
    \end{equation}
    where $\diff{\rvx_t} \coloneqq \nabla_{\rvx_t} \odot$ is the element-wise derivative operator.
    This is formally the same as the 1-dim Taylor series.
    Therefore, it is sufficient to consider the 1-dim Taylor series first,
    and parallelize each dimension later.
    Thus the derivatives we actually need are the following two.
    \begin{equation}
        \diff{\rvx_t} \scorefn(\rvx_t, t) = \nabla_{\rvx_t}\odot \scorefn(\rvx_t, t),\quad
        \diff{t}\scorefn(\rvx_t, t) =  (\bm{1} \diff{t}) \odot \scorefn(\rvx_t, t).
    \end{equation}

\subsection{Spatial Derivative $\diff{\rvx_t} \scorefn(\rvx_t, t) \coloneqq \nabla_{\rvx_t}\odot \scorefn(\rvx_t, t)$}
    Let us first compute the spatial derivative $(\rva \cdot \nabla_{\rvx_t}) \scorefn(\rvx_t, t)$.
    \begingroup
    \allowdisplaybreaks
    \begin{align}
        (\rva  \cdot \nabla_{\rvx_t}) \scorefn(\rvx_t,t)
        &=
        \left(\sum_i a_i \diff{{x_t}^i} \right)\scorefn(\rvx_t,t)
        \approx
        \left(\sum_i a_i \diff{{x_t}^i} \right) \frac{\rvx_t - \sqrt{1 - \nu_t} \rvx_0}{\sqrt{\nu_t}}
        \nonumber \\
        &=
        \frac{1}{\sqrt{\nu_t}}
        \begin{bmatrix}
            \left(\sum_i a_i \diff{{x_t}^i} \right) (\rvx_t - \sqrt{1 - \nu_t} \rvx_0)^1 \\
            \vdots \\
            \left(\sum_i a_i \diff{{x_t}^i} \right) (\rvx_t - \sqrt{1 - \nu_t} \rvx_0)^d
        \end{bmatrix}
        \nonumber \\
        &
        =
        \frac{1}{\sqrt{\nu_t}}
        \begin{bmatrix}
            \left(\sum_i a_i \diff{{x_t}^i} \right) ({x_t}^1 - \sqrt{1 - \nu_t} {x_0}^1) \\
            \vdots \\
            \left(\sum_i a_i \diff{{x_t}^i} \right) ({x_t}^d - \sqrt{1 - \nu_t} {x_0}^d)
        \end{bmatrix}
        \nonumber \\
        &
        =
        \frac{1}{\sqrt{\nu_t}}
        \begin{bmatrix}
            \left(a_1 \diff{{x_t}^1} \right) ({x_t}^1 - \sqrt{1 - \nu_t} {x_0}^1) \\
            \vdots \\
            \left(a_d \diff{{x_t}^d} \right) ({x_t}^d - \sqrt{1 - \nu_t} {x_0}^d)
        \end{bmatrix}
        \nonumber \\
        &
        =
        \frac{1}{\sqrt{\nu_t}}
        \begin{bmatrix}
            a_1 \\
            \vdots \\
            a_d
        \end{bmatrix}
        = \frac{1}{\sqrt{\nu_t}}\rva
        = \rva \odot \frac{1}{\sqrt{\nu_t}}\bm{1}.
    \end{align}
    \endgroup
    (Here, we used the notation ${x_t}^i$ to denotes the $i$-th component of a vector $\rvx_t$.)
    Thus we have formally obtained the relation
    \begin{equation}
        \rva \odot (\diff{\rvx_t}\scorefn(\rvx_t, t))
    =
        \rva \odot \frac{1}{\sqrt{\nu_t}}\bm{1},
    \end{equation}
    and therefore, we may formally write
    \begin{equation}
        \diff{\rvx_t}\scorefn(\rvx_t, t)
    =
        \frac{1}{\sqrt{\nu_t}}\bm{1}.
    \end{equation}
\subsection{Time Derivative $-\diff{t} \scorefn(\rvx_t, t)$}
    Next, let us compute $-\diff{t} \scorefn(\rvx_t, t)$.
    During the computation, $\rvx_0$ is replaced by the relation
    \begin{equation}
        \rvx_0 = \frac{1}{\sqrt{1 - \nu_t}} \left(\rvx_t - \sqrt{\nu_t}\scorefn(\rvx_t, t)\right).
    \end{equation}
    We also use the following relations between $\nu_t, \beta_t$, which is immediately obtained from the definition of $\nu_t$,
    \begin{equation}
        \dot{\nu}_t = (1 - \nu_t)\beta (t).
        \label{label:051}
    \end{equation}

    Using the above information,
    we may compute $-\diff{t} \scorefn(\rvx_t, t)$ as follows.
    \begingroup
    \allowdisplaybreaks
    \begin{align}
        -\diff{t} \scorefn(\rvx_t, t)
        &
        \approx
            -\diff{t} \frac{\rvx_t - \sqrt{1 - \nu_t}\rvx_0}{\sqrt{\nu_t}} \nonumber \\
        &= -\frac{1}{\sqrt{\nu_t}}
            \left( \frac{1}{2}\dot{\nu}_t(1 - \nu_t)^{-1/2}\rvx_0 \right)
        - (\rvx_t - \sqrt{1 - \nu_t}\rvx_0)
            \left( -\frac{1}{2}\dot{\nu}_t\nu_t^{-3/2}\right) \nonumber \\
        &= -\frac{\dot\nu_t}{2 \nu_t ^{3/2}}
            \left(
                \frac{\nu_t}{\sqrt{1 - \nu_t}}\rvx_0  - (\rvx_t - \sqrt{1 - \nu_t}\rvx_0)
            \right) \nonumber \\
        &= -\frac{\dot\nu_t}{2 \nu_t ^{3/2}}
            \left(
                - \rvx_t + \frac{1}{\sqrt{1 - \nu_t}} \rvx_0
            \right) \nonumber \\
        &= -\frac{\dot\nu_t}{2 \nu_t ^{3/2}}
            \left(
                - \rvx_t + \frac{1}{\sqrt{1 - \nu_t}} \frac{1}{\sqrt{1 - \nu_t}} \left(\rvx_t - \sqrt{\nu_t}\scorefn(\rvx_t, t)\right)
            \right) \nonumber \\
        &= -\frac{\dot\nu_t}{2 \nu_t ^{3/2}}
            \left(
                \left(- 1 + \frac{1}{1 - \nu_t} \right)\rvx_t
                - \frac{\sqrt{\nu_t}}{ 1 - \nu_t }
                \scorefn(\rvx_t, t)
            \right) \nonumber \\
        &= -\frac{1}{2 \nu_t ^{3/2}} \frac{ \dot{\nu}_t }{ 1 - \nu_t }
            \left(
                \nu_t \rvx_t - \sqrt{\nu_t} \scorefn(\rvx_t, t)
            \right) \nonumber \\
        &= - \frac{1}{2\nu_t^{3/2}} \beta (t)
            \left(
                \nu_t \rvx_t - \sqrt{\nu_t}\scorefn(\rvx_t, t)
            \right) \nonumber \\
        &= - \frac{\beta (t)}{2\sqrt{\nu_t}}
            \left(
                \rvx_t - \frac{\scorefn(\rvx_t, t)}{\sqrt{\nu_t}}
            \right).
    \end{align}
    \endgroup

    The ``derivatives" have some good points.
    For example, the partial derivatives commute,
    \begin{equation}
        \diff{\rvx_t}\diff{t} \scorefn(\rvx_t, t)
        =
        \diff{t}\diff{\rvx_t} \scorefn(\rvx_t, t).
    \end{equation}

\vfill
\newpage
\subsection{The Derivatives $\ldet (-\fdet)$, $\lst (-\fst)$, $\lst (g)$, $\gst (-\fst)$, $\gst (g)$}
    The computation of the derivative $\ldet (-\fdet)$, $\lst (-\fst)$, $\lst (g)$, $\gst (-\fst)$, $\gst (g)$
    does not require any particular nontrivial process.
    All we have to do is rewrite a term every time we encounter a derivative of $\scorefn(x_t, t)$ or $\nu_t$,
    and the rest is at the level of elementary exercises in introductory calculus.
    To execute this symbolic computation,
    the use of computer algebra systems will be a good option.
    It should be noted, however, that
    some implementation tricks to process such custom derivatives
    are required (in other words, the term-rewriting system should be customized).

    The results are shown below.
    Although these expressions appear complex at first glance, the code generation system can automatically generate code for such expressions.
    \begin{align}
        \ldet(-\fdet) (x_t, t)
            &= \left(\frac{\beta_t^2}{4} - \frac{\dot\beta_t}{2}\right) x_t
                + \left(\frac{\dot\beta_t}{2\sqrt{\nu_t}} - \frac{\beta_t^2}{4\nu_t^{3/2}} \right)\scorefn(x_t, t)
                \label{label:052}\\
        \lst (-\fst) (x_t, t)
            &= \left(\frac{\beta_t^2}{4} - \frac{\dot\beta_t}{2}\right) x_t
                + \frac{\dot\beta_t}{\sqrt{\nu_t}} \scorefn(x_t, t)\\
        \gst (-\fst) (x_t, t)
            &= \left(\frac{1}{2} - \frac{1}{\nu_t}\right)\beta_t^{3/2} \\
        \lst g (t)
            &= - \frac{\dot{\beta_t}}{2 \sqrt{\beta_t}}\\
        \gst g (t)
            &= 0.
    \end{align}
    We may also compute higher order derivatives. For example,
    \begin{align}
        \ldet\ldet(-\fdet) (x_t, t)
            &= \left(\frac{\beta_t^3}{8} - \frac{3\beta_t\dot\beta_t}{4} + \frac{\ddot\beta_t}{2}\right) x_t
            \nonumber \\
            &\phantom{=}
                + \left(
                    \frac{\beta_t^3(-\nu_t^2+3\nu_t - 3)}{8\nu_t^{5/2}}
                    + \frac{3\beta_t\dot\beta_t}{4\nu_t^{3/2}}
                    - \frac{\ddot\beta_t}{2\sqrt{\nu_t}}
                \right) \scorefn(x_t, t) \label{label:053}\\
        \lst\lst (-\fst) (x_t, t)
            &= \left(\frac{\beta_t^3}{8} - \frac{3\beta_t\dot\beta_t}{4} + \frac{\ddot\beta_t}{2}\right) x_t
                - \frac{\beta_t^3 + 4\ddot\beta_t}{4\sqrt{\nu_t}} \scorefn(x_t, t)\\
        \lst\gst (-\fst) (x_t, t)
            &= \frac{\sqrt{\beta_t}}{\nu_t^2}
            \left(\frac{\nu_t(2\beta_t^2 + 3\dot\beta_t)}{2} - \beta_t^2 - \frac{3\nu_t^2\dot\beta_t}{4}\right) \\
        \gst\lst (-\fst) (x_t, t)
            &= \sqrt{\beta_t}
            \left(\frac{\beta_t^2}{4} - \frac{\dot\beta_t}{2} + \frac{\dot\beta_t}{{\nu_t}}\right) \\
        \gst\gst (-\fst) (x_t, t)
            &= 0 \\
        \lst\lst g (t)
            &= \frac{2\beta_t\ddot\beta_t - \dot\beta_t^2}{4 \beta_t^{3/2}}\\
        \lst\gst g (t)
            &= 0 \\
        \gst\lst g (t)
            &= 0 \\
        \gst\gst g (t)
            &= 0.
    \end{align}
    As we can see, no factors other than
    integers, $x_t$, $\scorefn(x_t, t)$, $\nu_t$, $\beta_t$ and derivatives of $\beta_t$ appear.
    This is also true for higher order derivatives, which can be easily shown.

\vfill
\newpage
\subsection{SymPy Code Snippet for Automatic Symbolic Computation of Derivatives}

The following code snippet is a minimalistic example of SymPy code to
compute the above derivatives using the customized derivative method.
We used SymPy~1.11 to test the following code snippet.
{
\scriptsize
\begin{lstlisting}[frame=single,mathescape=true,escapechar=\%]
from sympy import Function, symbols, sqrt, simplify

x, t = symbols('x t') # ${\color{greencomments}{x, t}}$
B = Function('beta') # ${\color{greencomments}{\beta_t}}$

# define customized derivatives of $\color{greencomments}{\nu_t}$
class nu(Function):
    def fdiff(self, argindex=1):
        t, = self.args
        return (1 - nu(t)) * B(t) # ${\color{greencomments}{\dot\nu_t = (1 - \nu_t)\beta_t}}$

# define customized derivatives of $\color{greencomments}{\scorefn(x, t)}$
class S_theta(Function):
    def fdiff(self, argindex=1):
        x, t = self.args
        if argindex == 1: # ${\color{greencomments}{\partial/\partial x}}$
            d =  1 / sqrt(nu(t))
        elif argindex == 2: # ${\color{greencomments}{\partial/\partial t}}$
            d = (x - S_theta(x, t)/sqrt(nu(t))) * B(t) / (2 * sqrt(nu(t)))
        return d

# define $\color{greencomments}{\fdet}$
class f_flat(Function):
    @classmethod
    def eval(cls, x, t):
        return - B(t) * x / 2 + S_theta(x, t) * B(t) / (2 * sqrt(nu(t)))

# define differential operator $\color{greencomments}{L_\flat}$
class L_flat(Function):
    @classmethod
    def eval(cls, fxt):
        return -fxt.diff(t) - f_flat(x, t) * fxt.diff(x)

# show each derivative
print(f_flat(x, t))
print(simplify(L_flat(f_flat(x,t)))) # $\color{greencomments}{L_\flat \fdet (x_t,t); \text{~see \eqref{label:052}}}$
print(simplify(L_flat(L_flat(f_flat(x,t))))) # $\color{greencomments}{L_\flat L_\flat \fdet (x_t, t); \text{~see \eqref{label:053}}}$,

# we can similarly define $\color{greencomments}{\fst, L_\sharp, G_\sharp}$ and compute other derivatives.
\end{lstlisting}
}
The result will look like
\begin{align}
\text{[Out 1] }&
- \frac{x \beta{\left(t \right)}}{2} + \frac{S_{\theta}{\left(x,t \right)} \beta{\left(t \right)}}{2 \sqrt{\nu{\left(t \right)}}} \nonumber \\
\text{[Out 2] }&
- \frac{x \beta^{2}{\left(t \right)}}{4} + \frac{x \frac{d}{d t} \beta{\left(t \right)}}{2} + \frac{S_{\theta}{\left(x,t \right)} \beta^{2}{\left(t \right)}}{4 \nu^{\frac{3}{2}}{\left(t \right)}} - \frac{S_{\theta}{\left(x,t \right)} \frac{d}{d t} \beta{\left(t \right)}}{2 \sqrt{\nu{\left(t \right)}}} \nonumber \\
\text{[Out 3] }&
- \frac{x \beta^{3}{\left(t \right)}}{8} + \frac{3 x \beta{\left(t \right)} \frac{d}{d t} \beta{\left(t \right)}}{4} - \frac{x \frac{d^{2}}{d t^{2}} \beta{\left(t \right)}}{2}
+ \frac{S_{\theta}{\left(x,t \right)} \beta^{3}{\left(t \right)}}{8 \sqrt{\nu{\left(t \right)}}} \nonumber \\
&
- \frac{3 S_{\theta}{\left(x,t \right)} \beta^{3}{\left(t \right)}}{8 \nu^{\frac{3}{2}}{\left(t \right)}} + \frac{3 S_{\theta}{\left(x,t \right)} \beta^{3}{\left(t \right)}}{8 \nu^{\frac{5}{2}}{\left(t \right)}} - \frac{3 S_{\theta}{\left(x,t \right)} \beta{\left(t \right)} \frac{d}{d t} \beta{\left(t \right)}}{4 \nu^{\frac{3}{2}}{\left(t \right)}} + \frac{S_{\theta}{\left(x,t \right)} \frac{d^{2}}{d t^{2}} \beta{\left(t \right)}}{2 \sqrt{\nu{\left(t \right)}}}
\nonumber
\end{align}
and so on.
Some additional coding techniques can further improve the readability of these expressions,
but there will be no need to go any deeper into such subsidiary issues here.

Thus obtained symbolic expressions can be automatically converted into executable code
in practical programming languages including Python and C++ using a code generator,
though the authors hand-coded the obtained expressions in Python for the experiments in this paper.

\vfill
\newpage

\subsection{DDIM is an Ideal-Derivative-Based Sampler}\label{label:054}
    Using SymPy, we can easily compute the Taylor expansion of a given function.
    For example, the following code
{
\scriptsize
\begin{lstlisting}[frame=single,mathescape=true,escapechar=\%]
sympy.series(B(t+h), h, 0, 4)
\end{lstlisting}
}
    yields the result like
    \[
        \beta{\left(t \right)} + h \left. \frac{d}{d \xi_{1}} \beta{\left(\xi_{1} \right)} \right|_{\substack{ \xi_{1}=t }} + \frac{h^{2} \left. \frac{d^{2}}{d \xi_{1}^{2}} \beta{\left(\xi_{1} \right)} \right|_{\substack{ \xi_{1}=t }}}{2} + \frac{h^{3} \left. \frac{d^{3}}{d \xi_{1}^{3}} \beta{\left(\xi_{1} \right)} \right|_{\substack{ \xi_{1}=t }}}{6} + O\left(h^{4}\right).
    \]
    Similarly, using the relation  $\dot\nu_t = (1 - \nu_t)\beta_t$, we can easily compute
    the Taylor expansion of $\nu_{t-h}$ as follows.
{
\scriptsize
\begin{lstlisting}[frame=single,mathescape=true,escapechar=\%]
sympy.series(nu(t-h), h, 0, 3)
\end{lstlisting}
}
    \[
        \nu_{t-h} = \nu{\left(t \right)} + h \left(\beta{\left(t \right)} \nu{\left(t \right)} - \beta{\left(t \right)}\right) + h^{2} \left(\frac{\beta^{2}{\left(t \right)} \nu{\left(t \right)}}{2} - \frac{\beta^{2}{\left(t \right)}}{2} - \frac{\nu{\left(t \right)} \left. \frac{d}{d \xi_{1}} \beta{\left(\xi_{1} \right)} \right|_{\substack{ \xi_{1}=t }}}{2} + \frac{\left. \frac{d}{d \xi_{1}} \beta{\left(\xi_{1} \right)} \right|_{\substack{ \xi_{1}=t }}}{2}\right) + O\left(h^{3}\right)
    \]

    Using this functionality of SymPy, we can easily compute the Taylor expansion of the DDIM~\citep{song2020denoising}.
    Let us recall that the DDIM algorithm is given by \eqref{label:015},
    \begin{equation}
        \text{DDIM:} \quad \rvx_{t - h} \gets
            \underbrace{
                \sqrt{\frac{1 - \nu_{t - h}}{1 - \nu_t}}
            }_{
                \eqqcolon {\rho^\text{DDIM}_{t,h}}
            } \rvx_{t}
            + \underbrace{
                \frac{\sqrt{(1 - \nu_t)\nu_{t-h}} - \sqrt{(1 - \nu_{t-h})\nu_t}  }{\sqrt{1 - \nu_t}}
            }_{
                \eqqcolon {\mu^\text{DDIM}_{t,h}}
            } \scorefn(\rvx_t, t).
        \nonumber
    \end{equation}
    Then using SymPy, the Taylor expansion of $\rho^\text{DDIM}_{t, h}$ and $\mu^\text{DDIM}_{t, h}$ are computed as follows,
    \begin{align}
        \rho^\text{DDIM}_{t, h} &= 1 + \frac{\beta_t}{2}h
        - \frac{h^2}{4}\left(\frac{\beta_t^2}{2} - \dot\beta_t\right)
        + \frac{h^3}{4}\left(\frac{\beta_t^3}{12} - \frac{\beta_t\dot\beta_t}{2} + \frac{\ddot\beta_t}{3}\right)
        + o(h^3),\\
        \sqrt{\nu_t}\mu^\text{DDIM}_{t, h} &= -\frac{\beta_t}{2}h
        + \frac{h^2}{4}\left(\dot\beta_t - \frac{\beta_t^2}{2\nu_t}\right)
        + \frac{h^3}{4}\left(-\frac{\beta_t^3}{12}+\frac{\beta_t^3}{4\nu_t} - \frac{\beta_t^3}{4\nu_t^2} + \frac{\beta_t\dot\beta_t}{2\nu_t} - \frac{\ddot\beta_t}{3}\right)
        + o(h^3).
    \end{align}
    Although it has been known that DDIM corresponds to the Euler method
    up to 1st order terms~\citep{song2020denoising,salimans2022progressive},
    this expansion gives better understanding of higher order terms.
    That is, these are exactly equivalent to our deterministic Quasi-Taylor sampler
    \eqref{label:026} and \eqref{label:027} up to 3rd-order terms.
    This fact may suggest that the assumptions behind the DDIM derivation will be
    logically equivalent to our assumptions of ideal derivatives.

    The advantage of the proposed Quasi-Taylor method is that we can
    decide the hyperparameter at which order the Taylor expansion is truncated.
    On the other hand, DDIM automatically incorporates terms of much higher order, leaving no room for order tuning.

\vfill
\newpage
\section{On the Noise Schedule} \label{label:055}
\subsection{Derivation}
The noise scheduling function $\beta_t, \nu_t, \lambda(t)$, defined by
\begin{equation}
    \beta_t = \dot\lambda(t)\tanh\frac{\lambda(t)}{2}, \quad
    \nu_t = \tanh^2\frac{\lambda(t)}{2}, \quad
    \lambda(t) = \log(1 + Ae^{kt}),
\end{equation}
may appear to be more complex than necessary,
but these were naturally derived in the process of searching for a suitable form to
control the factor $\beta_t/\sqrt{\nu_t}$ to scale the output of neural network $\scorefn(x_t, t)$.
In this section, let us derive these functions.

Since this scaling factor $\beta_t/\sqrt{\nu_t}$ plays a fairly important role
in the behavior of the algorithm,
there is a natural motivation to design this factor itself directly.
Let us write it as $\xi(t) = \beta_t/\sqrt{\nu_t}$.
As $\nu_t = 1 - e^{-\int^t_0 \beta(t') dt'}$ by definition,
we immediately have $\beta_t = \dot\nu_t/(1 - \nu_t)$,
and we can rewrite the factor as follows,
\begin{equation}
    \xi(t) = \frac{\dot\nu_t}{(1 - \nu_t) \sqrt{\nu_t}}.
\end{equation}
If this is regarded as an ODE of $\nu_t$, the solution is given by
\begin{equation}
    \nu_t = \tanh^2 \left(\frac{1}{2}\ \int^t \xi(t')dt' \right).
\end{equation}
Since the integral contained in $\tanh$ is cumbersome, it will henceforth be written as $\lambda(t)$.
From this, the above $\beta_t$  and $\nu_t$ are immediately obtained.

Next, let us derive $\lambda(t)$. Unlike $\beta_t$ and $\nu_t$,
this is not determined by logical necessity, leaving room to design the functional form.
As the Picard-Lindel\"of theorem states,
for an ODE $\dot{x} = a(x, t)$ to have a unique solution
the coefficient $a(x,t)$ should be continuous w.r.t.\ $t$ and Lipschitz continuous w.r.t.\ $x$,
and otherwise, ODEs often behave less favorably\footnote{
    For example, the ODE $\dot{x} = x^2, x(0)=1$ has the solution $x = 1/(1-t)$ when $t < 1$,
    and it blows up at $t = 1$. It is usually impossible to consider what happens after $t > 1$ in ordinary contexts.
    Another well-known example is $\dot{x} = \sqrt{x}, x(0) = 0$. It has a solution $x = t^2/4$,
    but $x \equiv 0$ is also a solution.
    It actually has infinitely many solutions
    $x = 0$ (if $t \le t_0$),  $x = (t - t_0)^2/4$ (if $t > t_0$),  where $t_0 \ge 0$ is an arbitrary constant.
    In both cases, the coefficient $a(\cdot, \cdot)$ is not Lipschitz continuous.
    Even these seemingly simplest ODEs behave very complexly unless the coefficients are carefully designed.
}.
In PF-ODE, the Lipschitz condition is written as follows,
\begin{equation}
\left| \diff{x_t} \left(\frac{\beta_t}{2} x_t - \frac{\beta_t}{2\sqrt{\nu_t}} \scorefn(x_t, t) \right)  \right| < \infty.
\end{equation}
Using the ideal derivative of $\scorefn(x_t, t)$, this condition translates as $\beta_t|1 - 1/\nu_t| < \infty$.
It further simplifies as follows,
\begin{align}
\beta_t \left| 1 - \frac{1}{\nu_t} \right|
&= \beta_t \frac{1 - \nu_t}{\nu_t} = \frac{\dot\nu_t}{\nu_t} = \frac{d}{dt} \log \nu_t \nonumber \\
&\propto \frac{d}{dt} \log \tanh \frac{\lambda(t)}{2} = \frac{\dot\lambda(t)}{\sinh(\lambda(t)/2)\cosh(\lambda(t)/2)} \nonumber \\
&\propto \frac{\dot\lambda(t)}{\sinh \lambda(t)} < \infty.
\end{align}
When the argument is large,
$\sinh$ increases exponentially,
so it is very easy to make $\dot\lambda(t)/\sinh\lambda(t)$ bounded above by a finite constant.
To the contrary, when the $\lambda(t)$ is small,
$\dot\lambda(t)/\sinh\lambda(t)$ is approximated by $\dot\lambda(t)/\lambda(t)$.
In order to make this bounded above by a finite constant,
we have chosen the softplus function as one of the simplest functions.
It meets the Lipschitz condition and the requirement that $\nu_0 \approx 0$.
However, there may be better alternatives.

\subsection{Derivation of $A$ and $k$}\label{label:056}
If $\nu_0$ and $\nu_T$ are specified, we can explicitly compute $A$ and $k$ from them by solving
\begin{equation}
\begin{cases}
    \nu_0 = \tanh^2 \frac{1}{2}\log (1 + A) \\
    \nu_T = \tanh^2 \frac{1}{2}\log (1 + Ae^{kT}).
\end{cases}
\end{equation}
This simplifies using the following relation,
\begin{align}
&&
\nu_t &= \tanh^2 \frac{1}{2}\log(1 + Ae^{kt}) \nonumber \\
&\Leftrightarrow&
\mathrm{arctanh} \sqrt{\nu_t} &= \frac{1}{2}\log(1 + Ae^{kt}) \nonumber \\
&\Leftrightarrow&
\frac{1}{2}\log \frac{1 + \sqrt{\nu_t}}{1 - \sqrt{\nu_t}} &= \frac{1}{2}\log (1 + Ae^{kt}) \nonumber \\
&\Leftrightarrow&
\frac{1}{2}\log \left(1 + \frac{2\sqrt{\nu_t}}{1 - \sqrt{\nu_t}}\right) &= \frac{1}{2}\log (1 + Ae^{kt}) \nonumber \\
&\Leftrightarrow&
\frac{2\sqrt{\nu_t}}{1 - \sqrt{\nu_t}} &= Ae^{kt}.
\end{align}

\subsection{Plots}
\begin{figure}[H]
    \centering
    \includegraphics[width=0.99\linewidth]{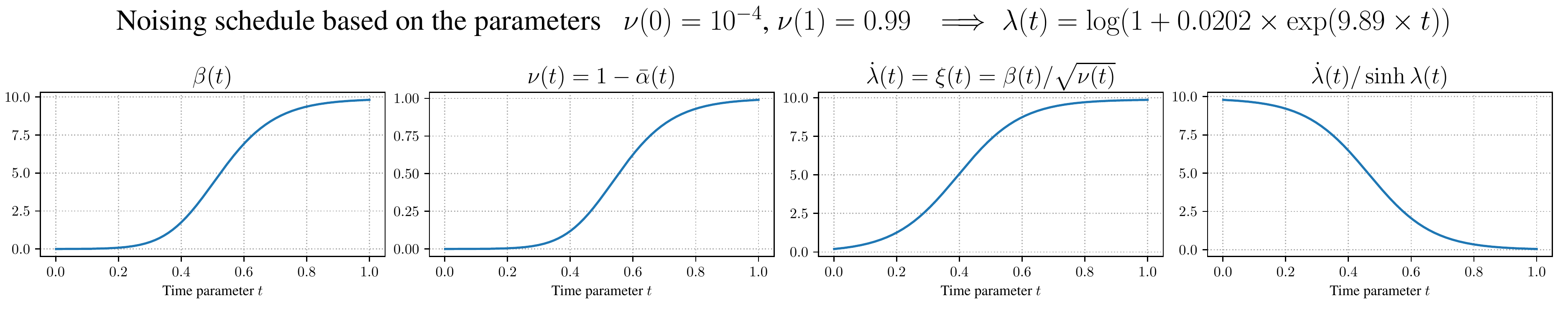}
    \vspace{3mm}\\
    \includegraphics[width=0.99\linewidth]{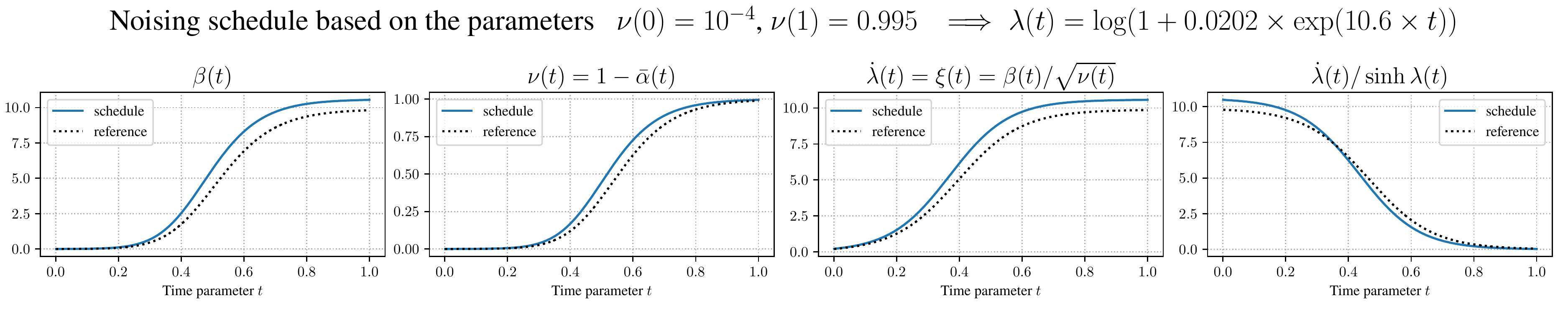}
    \vspace{3mm}\\
    \includegraphics[width=0.99\linewidth]{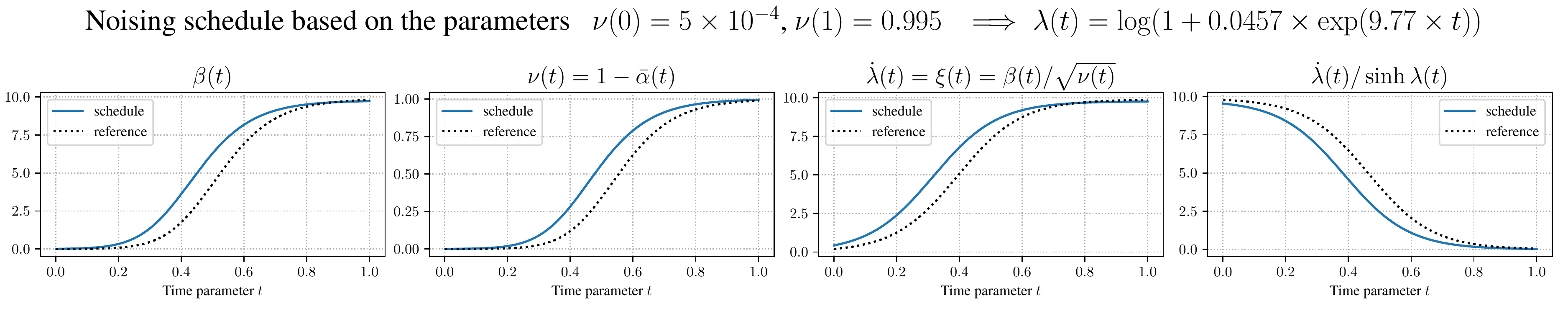}
    \vspace{3mm}\\
    \includegraphics[width=0.99\linewidth]{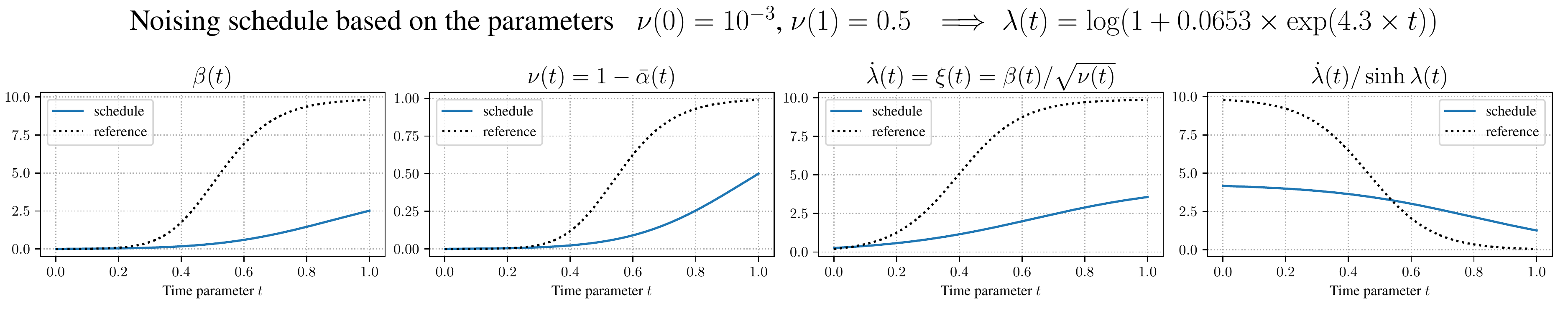}
    \caption{
        Some noise schedules.
        The dotted lines are the reference baselines,
        corresponding to the case where $\nu_0=10^{-4}, \nu(1) = 0.99$ (same as the top one).
    }\label{label:057}%
\end{figure}

Some noise schedules obtained by the above methods are shown in \figref{label:057}.

\vfill
\newpage
\section{Supplement on Fundamentals}
For convenience, let us summarize some basics behind the ideas in this paper.
The contents of this section are not particularly novel,
but
the authors expect that this section will give a better understanding of
the ideas of this paper and the continuous-time approach to diffusion generative models.

\subsection{Taylor Expansion and It\^o-Taylor Expansion}
    \subsubsection{Taylor Expansion of Deterministic Systems}\label{label:058}
        \paragraph{1-dimensional case}
        We first consider a 1-dim deterministic system $\dot{x}(t) = a(x(t), t)$.
        It is well known that this ordinary differential equation has a unique solution if $a(x, t)$ is Lipschitz
        continuous
        w.r.t.\ $x$ and continuous w.r.t.\ $t$ (Picard-Lindel\"of Theorem).
        Let $\varphi (x(t), t)$ be a differentiable function.
        Its total derivative is written as
        \begingroup
        \allowdisplaybreaks
        \begin{align}
            d\varphi &=
                \frac{\partial \varphi}{\partial t} dt +
                \frac{\partial \varphi}{\partial x} dx
                =
                \frac{\partial \varphi}{\partial t} dt +
                \frac{\partial \varphi}{\partial x}\frac{d x}{d t} dt
                =
                \left(
                    \frac{\partial \varphi}{\partial t} +
                    \frac{\partial \varphi}{\partial x} a(x, t)
                \right)dt \nonumber \\
                &=
                \underbrace{
                    \left(
                        \frac{\partial }{\partial t} +
                        a(x, t) \frac{\partial }{\partial x}
                    \right)
                    }_{\eqqcolon \ldet} \varphi dt.
            \label{label:059}
        \end{align}
        \endgroup
        By integrating both sides from $0$ to $t$, we have
        \begin{equation}
            \varphi(x(t), t) = \varphi(x(0), 0) + \int_{0}^{t} (\ldet\varphi)(x(s),s) ds.
            \label{label:060}
        \end{equation}
        We use this formula recursively to obtain the Taylor series of the above system.
        Let $\varphi(x(t), t) = x(t)$, then we have
        \begin{align}
            x(t)
                &=
                x(0) + \int_0^{t} (\ldet x)(x(s), s) ds \nonumber\\
                &=
                x(0) + \int_0^{t} a(x(s), s) ds.
        \end{align}
        Let $\varphi(x(t), t) = a(x(t), t)$, then we have
        \begin{equation}
            a(x(t), t) =
                a(x(0), 0) + \int_0^{t} (\ldet a)(x(s), s) ds.
        \end{equation}
        Using the above two equations, we have
        \begin{align}
            x(t) &= x(0) + \int_0^{t} \bigg( a(x(0), 0)
                +  \int_0^{t_1} (\ldet a)(x(t_2), t_2) dt_2 \bigg) dt_1 \nonumber \\
                &= x(0) + \int_0^t a(x(0),0 ) dt_1
                + \int_0^t \int_0^{t_1} (\ldet a)(x(t_2), t_2) dt_2 dt_1. \nonumber \\
                &= x(0) + t a(x(0),0 )
                + \int_0^t \int_0^{t_1} (\ldet a)(x(t_2), t_2) dt_2 dt_1.
        \end{align}
        We can expand again the term inside of the integral by using the following relation,
        \begin{equation}
            (\ldet a)(x(t), t) =
                (\ldet a)(x(0), 0) + \int_0^{t} (\ldet^2a)(x(s), s) ds,
        \end{equation}
        and obtain the expansion as follows,
        \begin{equation}
            x(t) = x(0) + t a(x(0),0 ) + \frac{t^2}{2}(\ldet a)(x(0), 0)
                + \int_0^t \int_0^{t_1} \int_0^{t_2}  (\ldet^2a)(x(t_3), t_3) dt_3 dt_2 dt_1.
        \end{equation}
        Applying this argument recursively, we can obtain the Taylor series of the deterministic system,
        if $a(\cdot, \cdot)$ is sufficiently smooth.
        \begin{equation}
            x(t) = x(0) + ta (x(0), 0) + \sum_{k = 2}^\infty \frac{t^k}{k!} (\ldet^{k-1} a)(x(0),0).
        \end{equation}
        Since the above discussion is valid when the integration interval is $(t,t+h)$ instead of $(0, t)$,
        it can be written as follows,
        \begin{equation}
            x(t+h) = x(t) + ha (x(t), t) + \sum_{k = 2}^\infty \frac{h^k}{k!} (\ldet^{k-1} a)(x(t),t).
        \end{equation}

        \paragraph{Multi-dimensional case}
        Let us consider the multi-dimensional ODE $\dot\rvx = \rva(\rvx, t)$.
        The total derivative of a smooth scalar function $\varphi(\rvx, t)$ is written as
        \begingroup
        \allowdisplaybreaks
        \begin{align}
            d\varphi &=
            \frac{\partial \varphi}{\partial t} dt +
            \sum_{i} \frac{\partial \varphi}{\partial x_i} dx_i
            =
            \frac{\partial \varphi}{\partial t} dt +
            \sum_i \frac{\partial \varphi}{\partial x_i}\frac{d x_i}{d t} dt
            =
            \frac{\partial \varphi}{\partial t} dt +
            \left(\sum_i a_i(\rvx, t) \frac{\partial }{\partial x_i}\right) \varphi
            dt \nonumber \\
            &=
            \left(
                \frac{\partial \varphi}{\partial t} +
                (\rva(\rvx, t) \cdot\nabla) \varphi
            \right)dt
            =
                \left(
                    \frac{\partial }{\partial t} +
                    \rva(\rvx, t) \cdot \nabla
                \right)
                \varphi dt.
        \end{align}
        \endgroup
        Let $\bm{\varphi}(\rvx, t)$ be a vector-valued smooth function,
        then we immediately have
        \begin{equation}
            d\bm{\varphi} =
            \left(
                \diff{t}
                + \rva(\rvx, t)\cdot \nabla
            \right)
            \bm{\varphi} dt.
        \end{equation}
        Using the scalar operator $L_\flat=(\diff{t} + \rva(\rvx, t)\cdot\nabla) $, we can
        obtain the following Taylor expansion similarly to the 1-dim case,
        \begin{align}
            \rvx(t) &=
                \rvx(0)
                + t(L_\flat \rvx)\bigg|_{t=0}
                + \frac{t^2}{2} (L_\flat^2 \rvx)\bigg|_{t=0} + \cdots \nonumber \\
            &=
                \rvx(0) +
                t\left(\diff{t} + \rva(\rvx, t)\cdot \nabla \right)\rvx\bigg|_{t=0}
                + \frac{t^2}{2} \left(\diff{t} + \rva(\rvx, t)\cdot \nabla \right)^2\rvx\bigg|_{t=0}  + \cdots
                \nonumber \\
            &=
                \rvx(0) +
                t \rva(\rvx(0), 0)
                + \frac{t^2}{2} \left(\diff{t} + \rva(\rvx, t)\cdot \nabla \right)\rva(\rvx, t)\bigg|_{t=0}  + \cdots
        \end{align}
        The second order term $(\rva\cdot \nabla) \rva$ is written as follows,
        \begin{equation}
            \left(\rva\cdot \nabla\right)\rva =
            \left(\sum_i a_i \partial_{x_i}\right)
            \begin{bmatrix}
                a_1(\rvx, t) \\
                \vdots\\
                a_d(\rvx, t)
            \end{bmatrix}
            =
            \begin{bmatrix}
                a_1 \diff{x_1} a_1 & \cdots & a_d \diff{x_d} a_1 \\
                \vdots & \ddots & \vdots \\
                a_1 \diff{x_1} a_d & \cdots & a_d \diff{x_d} a_d \\
            \end{bmatrix}
            \begin{bmatrix}
                1 \\ \vdots \\ 1
            \end{bmatrix}.
        \end{equation}
        In a special case where each dimension is separable, i.e.\ $\diff{x_i} a_j = 0 (i \ne j)$,
        the above $d\times d$ matrix is diagonal, and we have
        \begin{equation}
            \left(\rva\cdot \nabla\right)\rva =
            \begin{bmatrix}
                a_1 \diff{x_1} a_1 \\
                \vdots \\
                a_d \diff{x_1} a_d \\
            \end{bmatrix}
            =
            \begin{bmatrix}
                a_1  \\
                \vdots \\
                a_d \\
            \end{bmatrix}
            \odot
            \begin{bmatrix}
                \diff{x_1}  \\
                \vdots \\
                \diff{x_d} \\
            \end{bmatrix}
            \odot
            \begin{bmatrix}
                a_1  \\
                \vdots \\
                a_d \\
            \end{bmatrix}
            = \rva \odot \nabla \odot \rva.
        \end{equation}
        where $\odot$ is the element-wise product or operation.
        In this case, it is sufficient to consider each dimension separately,
        and it is formally equivalent to the 1-dim case.

    \subsubsection{It\^o-Taylor Expansion of Stochastic Systems}\label{label:061}
    For the above reasons, it is sufficient to consider the 1-dim case even in the stochastic case.
    As is well known, Taylor expansion is not valid for stochastic systems
    $x_t = a(x_t, t) dt + b(x_t, t) dB_t$.
    This is because of the relation ``$dB_t^2 \sim dt$''.
    This effect is taken into account in the celebrated It\^o's lemma,
    i.e., the stochastic version of \eqref{label:059},
    \begin{equation}
        d\varphi = \underbrace{
            \left(
                \frac{\partial}{\partial t}
                + a(x, t)\frac{\partial}{\partial x}
                + \frac{b(x, t)^2}{2}\frac{\partial^2}{\partial x^2}
            \right)}_{\eqqcolon \lst}
            \varphi dt +
            \underbrace{
            \left(
                b(x, t)\frac{\partial}{\partial x}
            \right)}_{\eqqcolon \gst}
            \varphi dB_t. \label{label:062}
    \end{equation}
    By using It{\^o}'s formula recursively, we can obtain the following higher-order expansion of a stochastic system,
    which is called  the It{\^o}-Taylor expansion.
    \begingroup
    \allowdisplaybreaks
    \begin{align}
        x_{h}
            &= x_0 + \int_{0}^{h} a(x_{t}, t) dt + \int_{0}^{h} b(x_t, t) dB_{t} \\
            &= x_0
                + \int_{0}^{h}
                    \bigg( a(x_0, 0) +
                    \int_{0}^{t} (\lst a)(x_s,s) ds + \int_{0}^{t} (\gst a)(x_s,s) dB_s \bigg) dt \nonumber \\
                &\phantom{{}= x_0{}}
                + \int_{0}^{h}
                    \bigg( b(x_0, 0) +
                    \int_{0}^{t} (\lst b)(x_s,s) ds + \int_{0}^{t} (\gst b)(x_s,s) dB_s \bigg) dB_{t}  \nonumber \\
            & = x_0 + a(x_0,0)\int_{0}^{h} dt
                    + \int_{0}^{h} \int_{0}^{t} (\lst a)(x_s,s) ds dt +\int_{0}^{h} \int_{0}^{t} (\gst a)(x_s,s) dB_s dt \nonumber\\
                &\phantom{{}= x_0{}}
                    +  b(x_0, 0) \int_{0}^{h} dB_{t}
                    + \int_{0}^{h} \int_{0}^{t} (\lst b)(x_s,s) ds dB_t + \int_{0}^{h} \int_{0}^{t} (\gst b)(x_s,s) dB_s dB_t \nonumber \\
            & = x_0 +
                 a(x_0,0)\int_{0}^{h} dt + b(x_0, 0) \int_{0}^{h} dB_{t} \nonumber \\
                &\phantom{{}= x_0{}}
                    +\int_{0}^{h} \int_{0}^{t} \bigg((\lst a)(x_0,0) + \cdots \bigg) ds dt
                    +\int_{0}^{h} \int_{0}^{t} \bigg((\gst a)(x_0,0) + \cdots \bigg) dB_s dt \nonumber \\
                &\phantom{{}= x_0{}}
                    +\int_{0}^{h} \int_{0}^{t} \bigg((\lst b)(x_0,0) + \cdots \bigg) ds dB_t
                    +\int_{0}^{h} \int_{0}^{t} \bigg((\gst b)(x_0,0) + \cdots \bigg) dB_s dB_t \nonumber \\
            & = x_0 + a(x_0,0)h + b(x_0, 0) B_h \nonumber \\
                &\phantom{{}= x_0{}}
                    + (\lst a)(x_0,0)\int_{0}^{h} \int_{0}^{t} ds dt
                    + (\gst a)(x_0,0)\int_{0}^{h} \int_{0}^{t} dB_s dt   \nonumber \\
                &\phantom{{}= x_0{}}
                    + (\lst b)(x_0,0)\int_{0}^{h} \int_{0}^{t}ds dB_{t}
                    + (\gst b)(x_0,0)\int_{0}^{h} \int_{0}^{t} dB_s dB_{t}  \nonumber \\
                &\phantom{{}= x_0{}}
                    + \textit{Remainder}, \label{label:063}
    \end{align}
    \endgroup
    where the remainder consists of triple integrals as follows,
    \begin{equation}
        \textit{Remainder} = \int_0^h\int_0^t\int_0^s (\lst^2a)(x_u,u) dudsdt + \cdots .
    \end{equation}
    We ignore these terms now.
    If we also ignore the double integrals, we obtain the Euler-Maruyama scheme.

\paragraph{Evaluation of Each Integral in \eqref{label:063}}
    Let us evaluate the double integrals.
    \begin{equation}
    \begin{cases}
        \displaystyle \int_0^h \int_0^{t} ds dt & \cdots\text{(deterministic)}\\
        \displaystyle \int_0^h \int_0^{t} dB_s dt & \cdots\text{(stochastic 1)}\\
        \displaystyle \int_0^h \int_0^{t} ds dB_{t} & \cdots\text{(stochastic 2)}\\
        \displaystyle \int_0^h \int_0^{t} dB_s dB_{t} & \cdots\text{(stochastic 3)}\\
    \end{cases} \nonumber
    \end{equation}

    \myparagraph{Deterministic}
    The deterministic one,
    $\int_{0}^{h} \int_{0}^{t}ds dt$,
    is easy to evaluate.
    \begin{equation}
            \int_{0}^{h} \int_{0}^{t}ds dt =  \int_{0}^{h}t dt =  \frac{1}{2}h^2.
    \end{equation}

    \myparagraph{Stochastic 1}
    Other integrals contain stochastic integrations.
    Let us denote the first one by $\tilde{z}$.
    \begin{equation}
        \tilde{z} \coloneqq \int_{0}^{h} \int_{0}^{t} dB_s dt
        = \int_0^h B_{t} dt.
    \end{equation}
    As $\tilde{z}$ is the limit of a sum of Gaussian variables, i.e.,
    \begin{align}
        \tilde{z}
            &= \lim_{n \to \infty} \sum_{i=0}^{n-1} \frac{h}{n} B_{hi/n}
            = \lim_{n \to \infty} \frac{h}{n}\sum_{i=0}^{n-1} \sum_{j=1}^i (B_{hj/n} - B_{h(j-1)/n)}) \nonumber \\
            &= \lim_{n \to \infty} \frac{h}{n}\sum_{i=0}^{n-1} \sum_{j=1}^i W_j, \quad W_j \sim \gN(0, \frac{h}{n}),
    \end{align}
    so $\tilde{z}$ is also a Gaussian, whose mean is $0$.
    The variance, however, requires some discussions,
    which shall be seen later.
    In addition, we shall also see that $\tilde{z}$ is correlated with $B_h$, 
    \begin{equation}
        \E [\tilde{z} \cdot  B_h ] \ne 0.
    \end{equation}

    \myparagraph{Stochastic 2}
    The second one
    has the correlation with the first one as follows.
    Here, we use the integral-by-parts formula. See e.g.~\citep[Theorem~4.1.5]{oksendal2013stochastic}.
    \begin{align}
            \int_{0}^{h} \int_{0}^{t}ds dB_{t}
        =
            \int_0^h t dB_{t}
        =
            t B_t \bigg|_0^h - \int_0^h B_{t} dt
        =
            hB_h - \tilde{z}.
    \end{align}

    \myparagraph{Stochastic 3}
    The third one is computed as follows,
    using a famous formula of It\^o integral.
    \begin{align}
            \int_{0}^{h}\int_{0}^{t} dB_s dB_{t}
        =
            \int_0^h B_{t} dB_{t}
        =
            \frac{1}{2} \left(B_h^2 - h\right).
    \end{align}
    This is derived by substituting $\varphi(x, t) = x^2$
    and $x_t = B_t$ (i.e., $dx_t = 0 \cdot dt + 1 \cdot dB_t$) into It\^o's formula \eqref{label:062}.

    \myparagraph{Substituting the Integrals to the It\^o-Taylor Expansion}
    Let us denote $\tilde{w} \coloneqq B_{h} \sim \gN(0, h)$,
    and let us shift the integral interval from $(0, h)$ to $(t,t+h)$.
    Then, we may rewrite the above second order expansion as follows,
    which we have already seen in the main text,
    \begin{align}
        x_{t + h}
            &= x_t + ha(x_t, t) + \tilde{w}b(x_t, t)   \nonumber \\
            &\phantom{{}= x_0{}}
            + (\lst a) (x_t, t) \cdot h^2/2
            + (\gst a) (x_t, t) \cdot \tilde{z} \nonumber \\
            &\phantom{{}= x_0{}}
            + (\lst b) (x_t, t) \cdot (\tilde{w}h - \tilde{z})
            + (\gst b) (x_t, t) \cdot (\tilde{w}^2 - h)/2 .
    \end{align}

\paragraph{Covariance of the Random Variables $\tilde{w}, \tilde{z}$}
    Next, let us evaluate the the variance of $\tilde{z}$,
    and the correlation between the Gaussian variables $\tilde{w}$ and $\tilde{z}$.
    Let us first calculate the variance of $(\tilde{w}h - \tilde{z})$.
    By It\^o's isometry, we have
    \begingroup
    \allowdisplaybreaks
    \begin{align}
        \E[(\tilde{w}h - \tilde{z})^2]
        = \E\left[\left( \int_0^h s dB_s \right)^2 \right]
        = \E\left[\int_0^h s^2 ds\right] = \frac{1}{3}h^3.
    \end{align}
    \endgroup
    The correlation between $\tilde{w}$ and $\tilde{z}$
    is similarly evaluated by It\^o's isometry (see~\citep[Proof of Lemma~3.1.5]{oksendal2013stochastic}) as follows,
    \begingroup
    \allowdisplaybreaks
    \begin{align}
        \E[\tilde{w} \tilde{z}]
            = \E\left[B_h\int_0^h s dB_s \right]
            = \E\left[\left(\int_0^h dB_s\right) \left(\int_0^h s dB_s \right)\right]
             = \E\left[\int_0^h s ds \right]
             = \frac{1}{2} h^2.
    \end{align}
    \endgroup
    Using the above variance and covariance,
    we can calculate the variance of $\tilde{z}$ as follows,
    \begingroup
    \allowdisplaybreaks
    \begin{align}
        \E[\tilde{z}^2]
            = \E[(\tilde{w}h - \tilde{z})^2 - h^2\tilde{w}^2 + 2h\tilde{w}\tilde{z}]
            = \frac{1}{3}h^3 - h^2 \cdot h + 2h\cdot\frac{h^2}{2}
            = \frac{1}{3} h^3.
    \end{align}
    \endgroup

    We need to find random variables $\tilde{w}, \tilde{z}$ that satisfy the requirements for (co)variances
    that $\E[\tilde{w}^2] = h$, $\E[\tilde{w}\tilde{z}] = h^2/2$ and $\E[\tilde{z}^2] = h^3/3$,
    and we can easily verify that the following ones do,
    \begin{equation}
        \begin{bmatrix}
        \tilde{w}\\
        \tilde{z}
        \end{bmatrix}
        =
        \begin{bmatrix}
        \sqrt{h} & 0 \\
        {h\sqrt{h}}/{2} & {h\sqrt{h}}/{2\sqrt{3}}
        \end{bmatrix}
        \begin{bmatrix}
        u_1 \\ u_2
        \end{bmatrix}
        \label{label:064}
    \end{equation}
    where $u_1, u_2 \stackrel{\text{i.i.d.}}{\sim} \gN(0, 1)$.
    Let us compute the covariance matrix just to be sure.
    \begingroup
    \allowdisplaybreaks
    \begin{align}
            \E\left[\begin{bmatrix}
                \tilde{w}\\
                \tilde{z}
            \end{bmatrix}
            \begin{bmatrix}
                \tilde{w} &
                \tilde{z}
            \end{bmatrix}
            \right]
        &=
            \E\bigg[
            \begin{bmatrix}
            \sqrt{h} & 0 \\
            {h\sqrt{h}}/{2} & {h\sqrt{h}}/{2\sqrt{3}}
            \end{bmatrix}
            \begin{bmatrix}
            u_1 \\ u_2
            \end{bmatrix}
            \begin{bmatrix}
            u_1 & u_2
            \end{bmatrix}
            \begin{bmatrix}
            \sqrt{h} & {h\sqrt{h}}/{2} \\
            0 & {h\sqrt{h}}/{2\sqrt{3}}
            \end{bmatrix} \bigg] \nonumber \\
        &=
            \begin{bmatrix}
            \sqrt{h} & 0 \\
            {h\sqrt{h}}/{2} & {h\sqrt{h}}/{2\sqrt{3}}
            \end{bmatrix}
            \E\left[
                \begin{bmatrix}
                u_1^2 & u_1u_2 \\
                u_1u_2 & u_2^2
                \end{bmatrix}
            \right]
            \begin{bmatrix}
            \sqrt{h} & {h\sqrt{h}}/{2} \\
            0 & {h\sqrt{h}}/{2\sqrt{3}}
            \end{bmatrix} \nonumber \\
        &=
            \begin{bmatrix}
            \sqrt{h} & 0 \\
            {h\sqrt{h}}/{2} & {h\sqrt{h}}/{2\sqrt{3}}
            \end{bmatrix}
            \begin{bmatrix}
            \sqrt{h} & {h\sqrt{h}}/{2} \\
            0 & {h\sqrt{h}}/{2\sqrt{3}}
            \end{bmatrix} \nonumber \\
        &=
            \begin{bmatrix}
                h & h^2/2 \\
                h^2/2 & h^3/3
            \end{bmatrix}.
    \end{align}
    \endgroup
\vfill
\newpage
    \subsection{Diffusion Process and Fokker-Planck Equation}\label{label:065}
        \paragraph{Derivation of Fokker-Planck Equation}
        In this section, let us derive the Fokker-Planck equation
        \begin{equation}
            \diff{t} p(\rvx_t, t)
                = -\nabla_{\rvx_t} \left(f(\rvx_t, t) p(\rvx_t, t)\right)
                + \frac{1}{2} g(t)^2 \nabla_{\rvx_t}^2 p(\rvx_t, t).
        \end{equation}
        from the It\^o SDE,
        \begin{equation}
            d \rvx_t = f(\rvx_t, t) dt + g(t) d \mathbf{B}_t.
        \end{equation}
        For simplicity, we consider a 1-dim case here.
        The following approach using the Fourier methods (characteristic function) will be easy and intuitive.
        See also e.g.~\citep[\S~5]{cox2017theory},\citep[\S~15]{karlin1981second},\citep[\S~6]{shreve2004stochastic} and \citep[\S~5]{sarkka2019applied}.

        Let us consider an infinitesimal time step $h$. Then $x_{t+h}$ is written as follows,
        \begin{equation}
            x_{t+h} = x_t + w_t, \quad w_t \sim \mathcal{N}(f(x_t, t) h, g(t)^2h).
        \end{equation}
        Let us consider the
        characteristic function $\phi_{x}(\omega) \coloneqq \mathbb{E}_{p(x)} [ \exp{i \omega x} ]$ (where $i = \sqrt{-1}$),
        which is the Fourier transform of the density function $p(x)$.
        Because of the convolution theorem $\phi_{x + y}(\omega) = \phi_{x}(\omega) \cdot \phi_{y}(\omega)$,
        the characteristic functions of $p(x_{t+h}, t+h)$, $p(x_t, t)$, and $p(w_t)$ have the following relation,
        \begin{equation}
            \phi_{x_{t+h}}(\omega) = \phi_{w_t}(\omega) \phi_{x_t}(\omega).
        \end{equation}
        It is easily shown that the characteristic function of the above Gaussian is given by
        \begin{equation}
            \phi_{w_t}(\omega)
                = \exp\left( i \omega f(x_t, t) h - \frac{1}{2} g(t)^2 h \omega^2 \right).
        \end{equation}
        Expanding the r.h.s.\ up to the first order terms of $h$, we have
        \begin{equation}
            \phi_{w_t}(\omega)
                = 1
                + i \omega f(x_t, t) h
                - \frac{1}{2} g(t)^2 h \omega^2
                + O(h^2).
        \end{equation}
        Thus we obtain
        \begin{equation}
            \frac{\phi_{x_{t+h}}(\omega) - \phi_{x_{t}}(\omega)}{h} =
            \left(
                -(-i \omega) f(x_t, t)
                + (-i\omega)^2 \frac{g(t)^2}{2}
            \right)\phi_{t}(\omega) + O(h).
        \end{equation}
        When $h\to 0$,
        \begin{equation}
            \diff{t} \phi_{x_{t}}(\omega) =
                -(-i \omega) f(x_t, t) \phi_{t}(\omega)
                + (-i\omega)^2 \frac{g(t)^2}{2}  \phi_{t}(\omega).
            \end{equation}
        Since $(-i\omega)$ in the Fourier domain corresponds to the spatial derivative $\diff{x_t}$ in the real domain
        \footnote{
            Integral by parts and $p(-\infty) = p(\infty) = 0$. Formally writing,
            \begin{equation}
                \mathbb{E}_{\frac{d}{dx} p(x)} [ e^{i \omega x} ]
                = \int_{-\infty}^{\infty} e^{i \omega x} \frac{d}{dx} p(x) dx
                = -\int_{-\infty}^{\infty} i\omega e^{i \omega x} p(x) dx
                = (-i\omega) \mathbb{E}_{p(x)}[e^{i\omega x}].\nonumber
            \end{equation}
        },
        it translates as follows,
        \begin{equation}
            \diff{t} p(x_t, t) =
                -\diff{x_t} (f(x_t, t) p(x_t, t))
                + \diff[2]{x_t} \frac{g(t)^2}{2} p(x_t, t),
        \end{equation}
        and thus we have obtained the Fokker-Planck equation.
        In particular, if $f=0$, this equation is called the heat equation, which was also first developed by Fourier.

        \paragraph{Example: Overdamped Langevin Dynamics}
        When the drift term is the gradient of a potential function $U(\cdot)$,
        the SDE is often called the overdampled Langevin equation\footnote{
            The Langevin equation should actually be the following equation system, which is called the underdamped Langevin equation,
            \begin{equation}
                \dot\rvx_t = \rvv_t, \quad
                M\dot\rvv_t = -\gamma \rvv_t -\nabla_{\rvx_t} U(\rvx_t) + \sqrt{2 D} \frac{d\mathbf{B}_t}{dt},\nonumber
            \end{equation}
            where $\rvv_t$ is the velocity (momentum) variable,
            $M$ is the mass of particle,
            and $\gamma$ is a constant called \textit{friction} or \textit{viscosity} coefficient.
            In this case, the energy function should also be modified as $E = \frac{1}{2}M \|\rvv\|^2 + U(\rvx).$
            Assuming that the mass is very small
            compared to the friction,
            the derivative of momentum $M \dot \rvv_t$ can be ignored
            (i.e.,
                if the force $F$ is constant,
                the ODE $M\dot v = -\gamma v + F$ has the solution $v = C e^{-t\gamma/M} + F/\gamma$,
                and the velocity immediately converges to $F/\gamma$ if $M \ll \gamma $
            )
            and
            the overdamped equation is obtained.
        },
        \begin{equation}
            d\rvx_t = -\nabla_{\rvx_t} U(\rvx_t) dt + \sqrt{2 D} d\mathbf{B}_t,
        \end{equation}
        where $D$ is a scalar constant.
        Its associated Fokker-Planck equation is
        \begin{align}
            \diff{t} p(\rvx_t, t)
            &= \nabla_{\rvx_t} \cdot
                \left(
                    \nabla_{\rvx_t} U(\rvx_t) \cdot p(\rvx_t, t) + \nabla_{\rvx_t} D p(\rvx_t, t)
                \right) \nonumber \\
            &= \nabla_{\rvx_t} \cdot
            \left[
                \nabla_{\rvx_t}
                \left(U(\rvx_t) + D \log p(\rvx_t, t) \right)\cdot
                p(\rvx_t, t)
            \right].
        \end{align}
        If $U(\rvx_t) + D \log p(\rvx_t, t)$ is constant, the r.h.s.\ will be zero, and therefore, $\diff{t}p(\rvx_t, t)=0.$
        That is,
        \begin{equation}
            p(\rvx) = \frac{1}{Z} e^{- U(\rvx) / D}, \quad \text{~where~} Z = \int_{\mathbb{R}^d} e^{- U(\rvx) / D} d\rvx
        \end{equation}
        is the stationary solution of the FPE, and it does no longer evolve over time.
        Therefore, it is expected that the particles obeying the Langevin equation will eventually follow this Boltzmann distribution.

\begin{figure}[!t]
            \centering
            \includegraphics[width=0.8\linewidth]{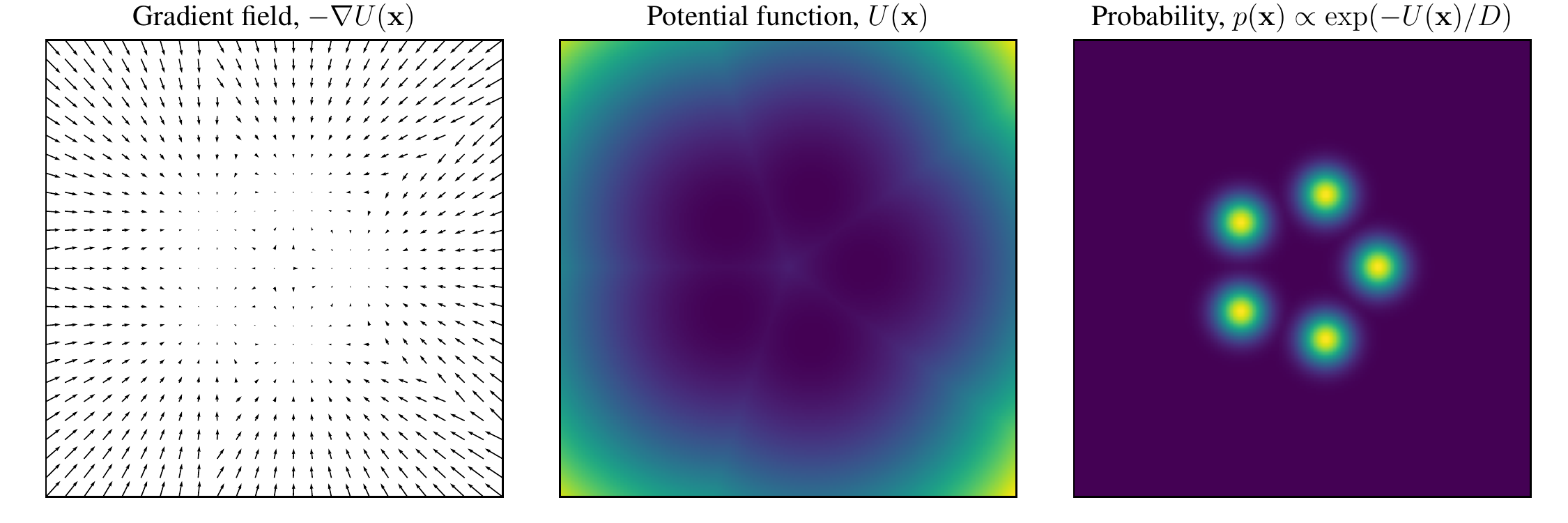} \\
            \captionof{figure}{
                Force field (potential gradient, score function) $-\nabla U(\rvx)$,
                potential function $U(\rvx)$, and the Boltzmann distribution $p(\rvx)=\frac{1}{Z}e^{-U(\rvx)/D}$.
                The scalar potential $U(\rvx)$ is given by \eqref{label:066}
            }
            \label{label:067}
            \begin{tabular}{m{\linewidth}}
                \includegraphics[width=\linewidth]{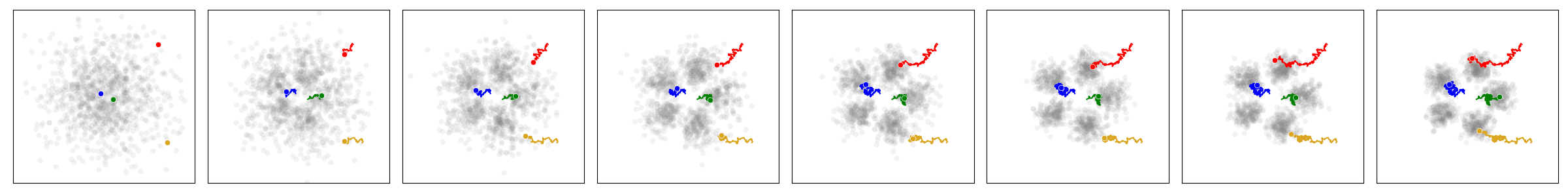}\\
                \includegraphics[width=\linewidth]{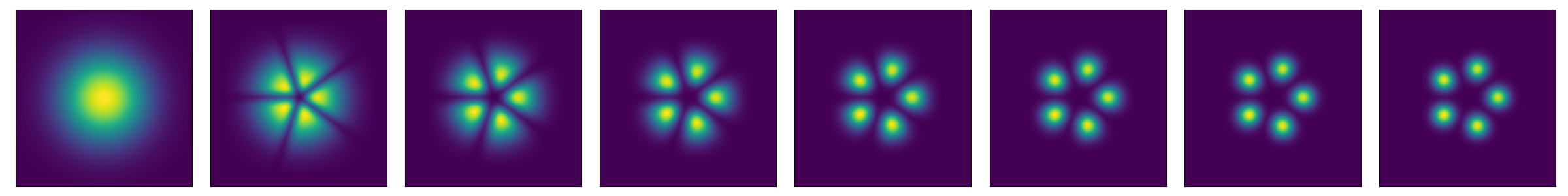}\\
            \end{tabular}
            \captionof{figure}{
                Numerical simulation of Langevin and Fokker-Planck equations.
                (Top: microscopic particle-wise dynamics)
                Time evolution of 1000 random samples $\{\rvx_t^{(i)}\}_{i=1}^{1000}$ obeying the Langevin equation.
                Paths of four particular samples are shown in color.
                The time evolves from left to right.
                (Bottom: macroscopic population dynamics)
                Time evolution of the density function $p(\rvx_, t)$ obeying the Fokker-Planck equation.
            }
            \label{label:068}
        \end{figure}

        Let us compare the microscopic dynamics of each particle obeying the Langevin equation
        with the macroscopic dynamics of the population obeying the FPE
        using a 2-dim toy model.
        The energy function we use here is the following Gaussian mixture model,
        \begin{equation}
            U(x,y) = - \log \sum_{k=1}^5 \exp
                \left[- \frac{(x - \cos \frac{2 k \pi}{5})^2 + (y - \sin \frac{2 k \pi}{5})^2}{2\sigma^2}\right],
                \label{label:066}
        \end{equation}
        where $\sigma = 0.1$.
        The diffusion parameter was $D = 5$.
        \figref{label:067} shows the
        force field $-\nabla U(\rvx)$ where $\rvx = (x, y)$,
        potential function $U(\rvx)$,
        and the Boltzmann distribution $p(\rvx)=\frac{1}{Z}e^{-U(\rvx)/D}$.

        \figref{label:068}
        shows the time evolution of Langevin and Fokker-Planck equations,
        where the initial density was $p(\rvx_0,0) = \mathcal{N}(\bm{0}, \mathbf{I})$.
        The time step size was $h = 5 \times 10^{-5}$,
        and the figures are plotted every 50 steps.
        These figures will be helpful to intuitively understand that the SDE (Langevin equation)
        and the FPE are describing the same phenomenon from different perspectives:
        individual description or population description.
    \vfill
    \newpage

    \subsection{Derivation of the Probability Flow ODE}\label{label:069}
    For convenience, the derivation of the PF-ODE is briefly described here.
    For more general case details, please refer to the original paper~\citep{song2020score}.

    Firstly, let us consider an SDE,
    \begin{equation}
        dx_t = f(x_t, t) dt + g(t) dB_t.
    \end{equation}
    The associated FPE is,
    \begin{equation}
        \diff{t}p(x_t, t)
        = \diff{x_t} \left(f(x_t, t) p(x_t, t) \right)
        - \frac{1}{2}g(t)^2\diff[2]{x_t} p(x_t, t).
    \end{equation}
    Here, let us forcibly incorporate the diffusion term (i.e., the second-order derivative) into the drift term (first-order derivative).
    \begin{align}
    \text{r.h.s.}
    &= \diff{x_t} \Big[
        f(x_t, t) p(x_t, t)
        - \frac{1}{2}g(t)^2 \diff{x_t} p(x_t, t)
    \Big] \nonumber \\
    &=\diff{x_t} \Big[
        \Big(
            f(x_t, t)
            - \frac{1}{2}g(t)^2 \frac{1}{p(x_t, t)} \diff{x_t} p(x_t, t )
        \Big)p(x_t, t)
    \Big] \nonumber \\
    &=\diff{x_t} \Big[
        \Big(
            \underbrace{
                f(x_t, t)
                - \frac{1}{2}g(t)^2 \diff{x_t} \log p(x_t, t)
            }_{\eqqcolon \fdet(x_t, t)}
        \Big)p(x_t, t)
    \Big].
    \end{align}
    Now we have obtained a `FPE' of the form,
    \begin{equation}
        \diff{t}p(x_t, t) = \diff{x_t} \left(\fdet(x_t, t) p(x_t, t)\right) - \cancel{\diff[2]{x_t} \left(0 \cdot  p(x_t, t)\right)},
    \end{equation}
    and its associated `SDE',
    \begin{equation}
        dx_t = \fdet(x_t, t) dt + \cancel{0 \cdot dB_t}.
    \end{equation}
    Although this `SDE' does not give the same particle-wise dynamics as the original SDE,
    its macroscopic dynamics of population is exactly the same as the original one,
    since the associated FPE has not changed from the original one at all.
    That is, the density evolves from $p(x_0, 0)$ to $p(x_T, T)$ exactly the same way.

    Since the obtained `SDE' is actually a deterministic ODE,
    its time-reversal is simply obtained by flipping the sign,
    i.e.,
    \begin{equation}
        dx_t = (- \fdet(x_t, t)) (-dt).
    \end{equation}
    Thus the Probability Flow ODE is obtained.
    If the terminal random variables are drawn from $p(x_T, T)$,
    then the particles obeying this ODE will reconstruct the initial density $p(x_0, 0)$ as a population.

    Note that the vector field of the form $\mathbf{J} \coloneqq \rho \rvv$ is often referred to as the \textit{flux},
    particularly in physical contexts e.g.\ fluid dynamics,
    where $\rho$ is density and $\rvv$ is velocity,
    and the PDE of the form
    \begin{equation}
        \diff{t} \rho + \nabla \cdot \mathbf{J} = 0
    \end{equation}
    is often called the \textit{continuity equation}, which is closely related to the conservation laws.
    In the present case, $\mathbf{J} \coloneqq p \fdet$ is understood as the flux,
    and therefore, $\fdet$ could be understood as a sort of velocity field.

    \vfill
    \newpage
    \subsection{On the Convergence of Numerical SDE Schemes.}\label{label:070}
        Let us introduce two convergence concepts which are commonly used in numerical SDE studies.
        See also \citep[\S~3.3, \S~3.4]{kloeden1994numerical}
        \begin{dfn}[Strong Convergence]
            Let $\tilde{\rvx}_t$ be the path at the continuous limit $h \to 0$,
            and $\rvx_t$ be the discretized numerical path,
            computed by a numerical scheme with the step size $h>0$.
            Then, it is said the numerical scheme has the strong order of convergence $\gamma$
            if the following inequality holds for a certain constant $K_\text{s} > 0$,
            \begin{equation}
                \E[|\rvx_t - \tilde{\rvx}_t|] \le K_\text{s} h^\gamma.
            \end{equation}
        \end{dfn}
        \begin{dfn}[Weak Convergence]
            Similarly,
            it is said that the scheme has the weak order of convergence $\beta $,
            if the following inequality holds for any test functions $\phi(\cdot)$ in a certain class of functions,
            and a certain constant $K_\text{w} > 0$,
            \begin{equation}
                |\E[\phi(\rvx_t)] - \E[\phi(\tilde{\rvx}_t)]| \le K_\text{w} h^\beta .
            \end{equation}
        \end{dfn}

        It is known that the Euler-Maruyama scheme has the strong convergence of order $\gamma = 0.5$,
        and weak order of $\beta  = 1$, in general.
        However, for more specific cases including the diffusion generative models
        that the diffusion coefficient $g(t)$ is not dependent on $\rvx_t$,
        the Euler-Maruyama scheme has a little better strong convergence of order $\gamma = 1$.

        The strong convergence is concerned with the precision of the path,
        while the weak convergence is with the precision of the moments.
        In our case, we are not much interested in whether a data $\hat{\rvx}_0$ generated using a finite $h > 0$
        approximates the continuous limit $\tilde{\rvx}_0, (h \to 0)$ driven by the same Brownian motion.
        Instead, we are more interested in whether the density $p(\hat{\rvx}_0)$ of the samples generated with a finite step size $h>0$
        approximates the ideal density $p (\tilde{\rvx}_0), (h \to 0)$
        which is supposed to approximate the true density $p(\rvx_0)$.
        In this sense, the concept of strong convergence is not much important for us,
        but the weak convergence would be sufficient.

    \vfill
    \newpage
    \subsection{Runge-Kutta methods}\label{label:071}
        Let us briefly introduce the derivation of Runge-Kutta methods
        for a 1-dim ODE $\dot x = f(x, t)$.
        We are interested in deriving the following derivative-free formula
        \begin{equation}
            x_{t+h} = x_t + \sum_{i=1}^n h b_i k_i + o(h^p),
            \quad\text{where~~}
            k_i = f(x_t + \sum_{j<i} h a_{ij} k_j, t + h c_i), \quad n \ge p.
        \end{equation}
        The array of coefficients are often shown in the following form,
        which is called the Butcher tableau.
        \begin{equation}
            \begin{array}{c|ccccc}
                0    \\
                c_2    & a_{21}\\
                c_3    & a_{31} & a_{32} \\
                \vdots & \vdots & \vdots & \ddots \\
                c_n    & a_{n1} & a_{n2} & \cdots & a_{n,n-1}\\
                \hline
                       & b_1    & b_2    & \cdots & b_{n-1} & b_n
            \end{array}
        \end{equation}
        Now the problem is how to design each value in this tableau.
        First, for simplicity, let us consider the following 2nd order case,
        \begin{align}
            x_{t+h} &= x_t + hb_1k_1 + hb_2k_2 + o(h^2)\nonumber \\
            k_1 &= f(x_t, t)\nonumber \\
            k_2 &= f(x_t + h a_{21} k_1, t + h c_2),\nonumber
        \end{align}
        or,
        \begin{equation}
        x_{t+h} = x_t + hb_1f(x_t, t) + hb_2 f(x_t + h a_{21} f(x_t, t), t + h c_2) + o(h^2). \label{label:072}
        \end{equation}
        Noting that the Taylor expansions of the l.h.s.\ and the third term in r.h.s.\ are written as follows,
        \begin{align}
            \text{(l.h.s.)}
            &= x_t + h f(x_t, t) + \frac{h^2}{2} (\dot{f} + f f') (x_t, t) + o(h^2),\\
            \text{(third term of r.h.s.)}
            &=
                hb_2 \left(f(x_t, t) + h a_{21} f(x_t, t) f'(x_t, t) + hc_2 \dot{f}(x_t, t) + o(h)\right),
        \end{align}
        we obtain the following relation.
        \begin{multline}
            x_t + h f(x_t, t) + \frac{h^2}{2} (\dot{f} + f f') (x_t, t) = \\
            x_t + hb_1f(x_t, t) + hb_2\left(f(x_t, t) + h a_{21} f(x_t, t) f'(x_t, t) + hc_2 \dot{f}(x_t, t)\right) + o(h^2).
        \end{multline}
        Comparing both sides,
        we can find that
        the derivatives in both sides can be eliminated if the following equation are satisfied,
        \begin{align}
            h f(x_t, t) &= hb_1f(x_t, t) + hb_2f(x_t, t),\\
            \frac{h^2}{2} (\dot{f} + f f') (x_t, t) &= hb_2\left(h a_{21} f(x_t, t) f'(x_t, t) + hc_2 \dot{f}(x_t, t)\right).
        \end{align}
        By simplifying the above equations, we obtain the following relations of coefficients.
        \begin{gather}
            b_1 + b_2 = 1,
            \quad
            b_2 a_{21} = \frac{1}{2},
            \quad
            b_2 c_2 = \frac{1}{2},
        \end{gather}
        and thus we obtain the following Butcher tableau.
        \begin{equation}
            \begin{array}{c|cc}
                0    \\
                c_2 & a_{21} \\
                \hline
                    & b_1 & b_2
            \end{array}
            =
            \begin{array}{c|cc}
                0    \\
                c_2 & c_2 \\
                \hline
                    & 1 - 1/(2c_2) & 1/(2c_2)
            \end{array}
        \end{equation}
        (When $c_2 = 1$, the method is particularly called Heun's method.)
        Thus we have confirmed that the second order Taylor expansion of $x_{t+h}$ is expressed
        as \eqref{label:072},
        which is independent of any derivatives of $f$.

        We can similarly consider higher-order methods.
        In the 3rd-order case, following relations are automatically obtained after a little effort of
        symbolic math programming using e.g.\ SymPy,
        \begin{gather}
            b_1 + b_2 + b_3 = 1, \quad
            b_2c_2 + b_3c_3 = \frac{1}{2}, \quad
            b_2c_2^2 + b_3c_3^2 = \frac{1}{3}, \quad
            a_{32}a_{21}b_3 = \frac{1}{6}, \quad
            a_{32}b_3c_2 = \frac{1}{6},
            \nonumber \\
            a_{21}b_2 + (a_{31} + a_{32})b_3 = \frac{1}{2}, \quad
            a_{21}^2b_2 + (a_{32} + a_{31})^2b_3 = \frac{1}{3},\quad
            a_{21}b_2c_2 + (a_{31} + a_{32})b_3c_3 = \frac{1}{3}.\nonumber
        \end{gather}
        By solving this equation system using SymPy, the following Butcher tableau is obtained.
        \begin{equation}
            \begin{array}{c|ccc}
                0    \\
                c_2 & a_{21} \\
                c_3 & a_{31} & a_{32} \\
                \hline
                    & b_1 & b_2 & b_3
            \end{array}
            =
            \begin{array}{c|ccc}
                0    \\
                c_2 (\ne 2/3) & c_2 \\
                c_3 (\ne c_2) & \frac{c_3 (3c_2^2 - 3c_2 + c_3)}{c_2(3c_2-2)}& \frac{c_3(c_2-c_3)}{c_2(3c_2-2)} \\
                \hline
                    & 1 -\frac{1}{2c_2}-\frac{1}{2c_3}+\frac{1}{3c_2c_3} & \frac{2 - 3c_3}{6c_2(c_2-c_3)} & \frac{3c_2-2}{6c_3(c_2-c_3)}
            \end{array}
        \end{equation}
        It will naturally become much more complicated when we consider higher-order methods,
        though it can be simpler if substituting specific values.
        In the 4th-order case,
        when the Butcher tableau is written as follows,
        the method is particularly called the Classical 4th-order Runge-Kutta (RK4) method.
        \begin{equation}
            \begin{array}{c|cccc}
                0    \\
                c_2 & a_{21} \\
                c_3 & a_{31} & a_{32} \\
                c_4 & a_{41} & a_{42} & a_{43}\\
                \hline
                    & b_1 & b_2 & b_3 & b_4
            \end{array}
            =
            \begin{array}{c|cccc}
                0    \\
                1/2 & 1/2 \\
                1/2 & 0 & 1/2 \\
                1 & 0 & 0 & 1 \\
                \hline
                    & 1/6 & 1/3 & 1/3 & 1/6
            \end{array}
        \end{equation}

\begin{figure}[H]
            \centering
            \includegraphics[width=0.8\linewidth]{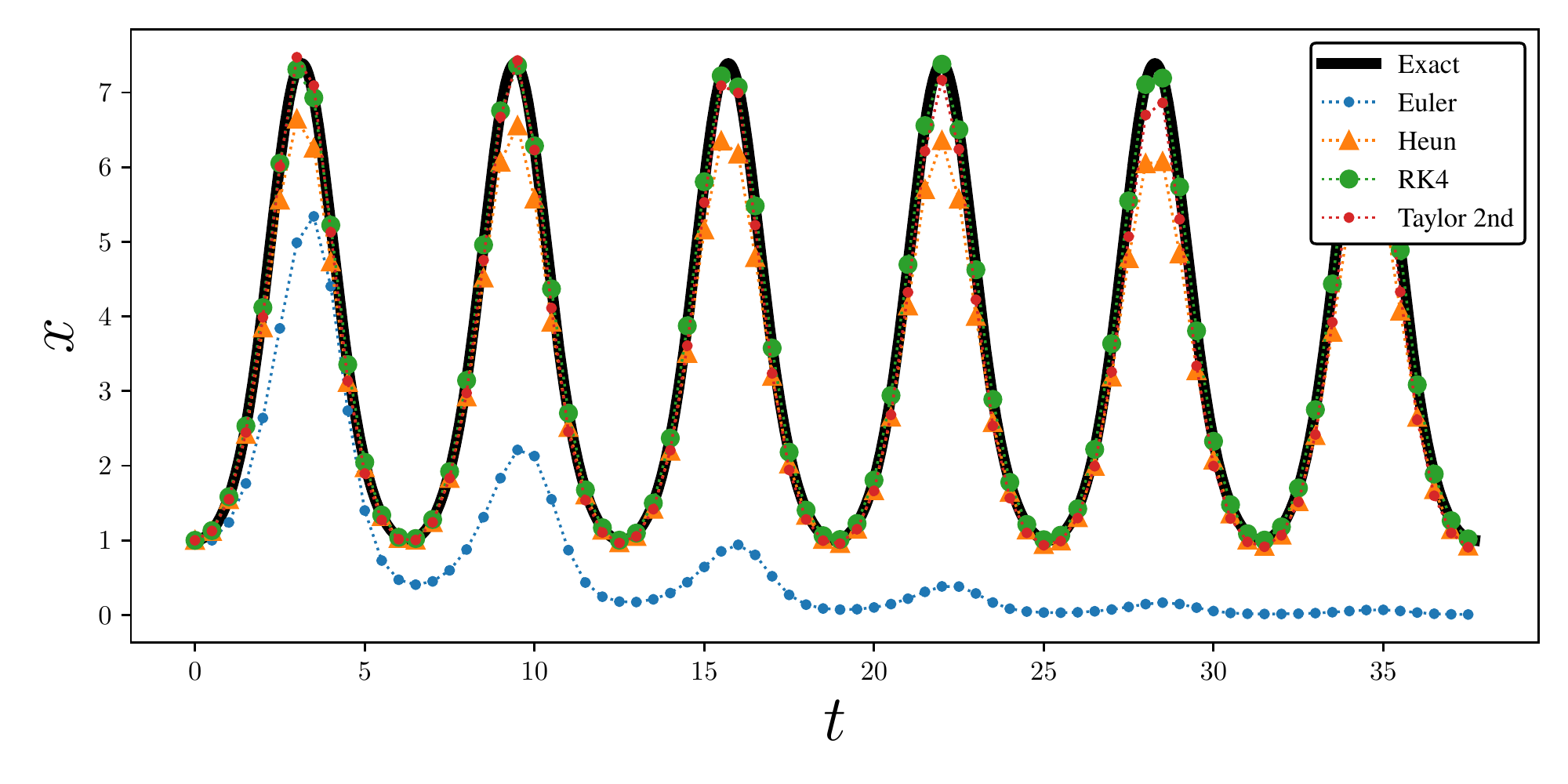}
            \caption{Comparison of numerical solutions of the ODE $\dot{x} = x \sin t, x_0 = 1$.}\label{label:073}
        \end{figure}

        \paragraph{Numerical Example}
        Let us consider a toy example $\dot{x} = x \sin t, x_0 = 1$.
        The exact solution is $x_t = e^{1-\cos t}$.
        \figref{label:073} compares the numerical solutions
        of the Euler, Heun, Classical RK4 and Taylor 2nd methods,
        where the step size is $h = 0.5$.
        It is clearly observed that
        the Euler method immediately deviates significantly from the exact solution
        but higher-order methods follow it for longer periods;
        while 2nd-order methods (Heun and Taylor 2nd) gradually deviate from the exact solution,
        the Classical 4th-order Runge-Kutta method more finely approximates the exact solution.
        Note that the Taylor 2nd method is given as follows,
        \begin{equation}
            x_{t+h} = x_t + h (x_t \sin t) + \frac{h^2}{2} \underbrace{(x_t \cos t + x_t \sin^2t)}_{\text{derivative of~} x \sin t}.
        \end{equation}
        In this case the computation of the derivative $(\diff{t} + x \sin t \diff{x})(x\sin t)$ is tractable.
        However, it will be infeasible if this term is a little more complicated.
        The Runge-Kutta method is advantageous if the derivatives are not easy to compute.

\vfill
\newpage
\section{Pseudocode}\label{label:074}
\subsection{Quasi-Taylor sampler}
\begin{minipage}{\textwidth}
The following pseudocode shows the proposed Quasi-Taylor sampler.

\begin{algorithm}[H]
    \newcommand{\MYOUTPUT}{\item[\textbf{Output:}]}
    \newcommand{\MYINIT}{\item[\textbf{Constants:}]}
    \newcommand{\MYBEGIN}{\item[\textbf{Begin}]}
    \newcommand{\MYEND}{\item[\textbf{End}]}
    \newcommand{\mycom}[1]{\hfill$\sharp$~{#1}}
    \caption{Quasi-Taylor Sampling Scheme with Ideal Derivatives}
    {\small
    \begin{algorithmic}
        \REQUIRE
            \STATE Trained neural network model $\scorefn(\rvx_t, t, \rvc)$
            \mycom{The conditioning information $\rvc$ is optional.}
            \STATE Data size $d > 0$ \quad $\cdots$\textit{Int}
            \STATE Total time $T > 0$\quad $\cdots$\textit{Float}
            \STATE Number of steps $N > 0$\quad $\cdots$\textit{Int}
            \STATE Initial and terminal noise levels $\nu_0, \nu_T \in (0, 1)$ \quad $\cdots$\textit{Float}
            \STATE (Optional) Conditioning information $\rvc$
            \STATE Step size schedule $h_1, h_2, \cdots, h_N$ \quad $\cdots$ \textit{List of Float}
        \MYINIT
            \STATE $\displaystyle\vphantom{\int}
                A \gets \frac{2 \sqrt{\nu_0}}{1 - \sqrt{\nu_0}},
                \quad
                k \gets \frac{1}{T} \left(\log\frac{2\sqrt{\nu_T} }{1 - \sqrt{\nu_T}} - \log A\right)$
                \mycom{\eqref{label:033}, \secref{label:056}}
        \MYBEGIN
            \STATE $\displaystyle\vphantom{\int}
                \rvx \sim \gN(\bm{0}_d, \mathbf{I}_d)$
            \mycom{Draw a $d$-dimensional Gaussian noise with unit variance}

            \FOR{$n = 1$ \textbf{to} $N$}

                \STATE
                \STATE{\textbf{[Step size and time parameter]}}
                \STATE $\displaystyle\vphantom{\int}
                    h \gets h_n$
                \STATE $\displaystyle\vphantom{\int}
                    t \gets T - \sum_{i=1}^{n-1} h_i$
                \STATE
                \STATE{\textbf{[Compute the Noise Level]}}
                    \mycom{\eqref{label:032}, \secref{label:055}}
                \STATE $\displaystyle\vphantom{\int}
                    \lambda \gets \log (1 + A e^{k t}),
                    \quad
                    \displaystyle \dot{\lambda} \gets \frac{Ake^{kt}}{Ae^{kt} + 1},
                    \quad
                    \displaystyle \ddot{\lambda} \gets \frac{Ak^2e^{kt}}{(Ae^{kt} + 1)^2},
                    \quad
                    \displaystyle \dddot{\lambda} \gets \frac{-Ak^3e^{kt}(Ae^{kt}-1)}{(Ae^{kt} + 1)^3}.$
                \STATE
                    $\displaystyle\vphantom{\int}
                    \nu \gets \tanh^2 \frac{\lambda}{2}.
                    $
                \STATE
                    $\displaystyle\vphantom{\int}
                    \beta \gets \dot\lambda \tanh \frac{\lambda}{2},
                    \quad
                    \dot\beta \gets \ddot\lambda \tanh \frac{\lambda}{2}
                        + \frac{\dot\lambda^2}{2\cosh^2\frac{\lambda}{2}},
                    \quad
                    \ddot\beta \gets -\frac{\dot\lambda^3 \sinh\frac{\lambda}{2}}{2\cosh^3\frac{\lambda}{2}}
                    + \dddot\lambda \tanh\frac{\lambda}{2}
                    + \frac{3\dot\lambda\ddot\lambda}{\cosh^2\frac{\lambda}{2}}
                    $

                \STATE~
                \STATE{\textbf{[Compute Coefficients]}}
                \STATE $\displaystyle\vphantom{\int}$
                    Compute $\rho^\flat$ using \eqref{label:026}
                \STATE $\displaystyle\vphantom{\int}$
                    Compute $\mu^\flat$ using \eqref{label:027}

                \STATE~
                \STATE{\textbf{[Update Data]}}
                \STATE $\displaystyle\vphantom{\int}
                    \rvx \gets \rho^\flat \rvx + \mu^\flat \scorefn(\rvx, t, \rvc)/\sqrt{\nu}$
                    \mycom{\eqref{label:025}}
                \STATE (Optional) Clip outliers of $\rvx$ so that e.g.\ $-1 \le \rvx \le 1$.
            \ENDFOR
        \MYEND
        \MYOUTPUT $\rvx$
    \end{algorithmic}
    }
\end{algorithm}
\end{minipage}

Note that, $\rho_\flat$ and $\mu_\flat$, as well as $\nu, \lambda, \beta$ and their derivatives,
are only dependent on $T, \{h_i\}, \nu_0$ and $\nu_T$.
Therefore, they can be pre-computed before the actual synthesis.

\vfill
\newpage

\subsection{Quasi-It\^o-Taylor sampler}
\begin{minipage}{\textwidth}
The following pseudocode shows the proposed Quasi-It\^o-Taylor sampler.

\begin{algorithm}[H]
    \newcommand{\MYOUTPUT}{\item[\textbf{Output:}]}
    \newcommand{\MYINIT}{\item[\textbf{Constants:}]}
    \newcommand{\MYBEGIN}{\item[\textbf{Begin}]}
    \newcommand{\MYEND}{\item[\textbf{End}]}
    \newcommand{\mycom}[1]{\hfill$\sharp$~{#1}}
    \caption{Quasi-It\^o-Taylor Sampling Scheme with Ideal Derivatives}
    {\small
    \begin{algorithmic}
        \REQUIRE
            \STATE Trained neural network model $\scorefn(\rvx_t, t, \rvc)$
            \mycom{The conditioning information $\rvc$ is optional.}
            \STATE Data size $d > 0$ \quad $\cdots$\textit{Int}
            \STATE Total time $T > 0$\quad $\cdots$\textit{Float}
            \STATE Number of steps $N > 0$\quad $\cdots$\textit{Int}
            \STATE Initial and terminal noise levels $\nu_0, \nu_T \in (0, 1)$ \quad $\cdots$\textit{Float}
            \STATE (Optional) Conditioning information $\rvc$
            \STATE Step size schedule $h_1, h_2, \cdots, h_N$ \quad $\cdots$ \textit{List of Float}
        \MYINIT
            \STATE $\displaystyle\vphantom{\int}
                A \gets \frac{2 \sqrt{\nu_0}}{1 - \sqrt{\nu_0}},
                \quad
                k \gets \frac{1}{T} \left(\log\frac{2\sqrt{\nu_T} }{1 - \sqrt{\nu_T}} - \log A\right)$
                \mycom{\eqref{label:033}, \secref{label:056}}
        \MYBEGIN
            \STATE $\displaystyle\vphantom{\int}
                \rvx \sim \gN(\bm{0}_d, \mathbf{I}_d)$
            \mycom{Draw a $d$-dimensional Gaussian noise with unit variance}

            \FOR{$n = 1$ \textbf{to} $N$}

                \STATE
                \STATE{\textbf{[Step size and time parameter]}}
                \STATE $\displaystyle\vphantom{\int}
                    h \gets h_n$
                \STATE $\displaystyle\vphantom{\int}
                    t \gets T - \sum_{i=1}^{n-1} h_i$
                \STATE
                \STATE{\textbf{[Compute the Noise Level]}}
                    \mycom{\eqref{label:032}, \secref{label:055}}
                \STATE $\displaystyle\vphantom{\int}
                    \lambda \gets \log (1 + A e^{k t}),
                    \quad
                    \displaystyle \dot{\lambda} \gets \frac{Ake^{kt}}{Ae^{kt} + 1},
                    \quad
                    \displaystyle \ddot{\lambda} \gets \frac{Ak^2e^{kt}}{(Ae^{kt} + 1)^2}.$
                \STATE
                    $\displaystyle\vphantom{\int}
                    \nu \gets \tanh^2 \frac{\lambda}{2}.
                    $
                \STATE
                    $\displaystyle\vphantom{\int}
                    \beta \gets \dot\lambda \tanh \frac{\lambda}{2},
                    \quad
                    \dot\beta \gets \ddot\lambda \tanh \frac{\lambda}{2}
                        + \frac{\dot\lambda^2}{2\cosh^2\frac{\lambda}{2}}.
                    $

                \STATE~
                \STATE{\textbf{[Compute Coefficients]}}
                \STATE $\displaystyle\vphantom{\int}$
                    Compute $\rho^\sharp$ using \eqref{label:030}
                \STATE $\displaystyle\vphantom{\int}$
                    Compute $\mu^\sharp$ using \eqref{label:030}

                \STATE~
                \STATE{\textbf{[Draw a Correlated Driving Noise]}}
                \IF{$n = N$}
                    \STATE $\displaystyle\vphantom{\int}
                    \rvn^\sharp = \bm{0}_d$ \mycom{No noise is injected at the final step}
                \ELSE
                    \STATE $\displaystyle\vphantom{\int}
                        \rvw \sim \gN(\bm{0}_d, \mathbf{I}_d),
                        \quad \rvu \sim \gN(\bm{0}_d, \mathbf{I}_d),
                        \quad \rvz = \frac{1}{2}\rvw + \frac{1}{2\sqrt{3}}\rvu$
                        \mycom{See \eqref{label:064} and Theorem~\ref{label:018}}
                    \STATE $\displaystyle\vphantom{\int}$
                        Compute $\rvn^\sharp$ using \eqref{label:031}
                        and $\rvw, \rvz$ above.
                \ENDIF

                \STATE~
                \STATE{\textbf{[Update Data]}}
                \STATE $\displaystyle\vphantom{\int}
                    \rvx \gets \rho^\sharp \rvx + \mu^\sharp \scorefn(\rvx, t, \rvc)/\sqrt{\nu} + \rvn^\sharp$
                    \mycom{\eqref{label:029}}
                \STATE (Optional) Clip outliers of $\rvx$ so that e.g.\ $-1 \le \rvx \le 1$.
            \ENDFOR
        \MYEND
        \MYOUTPUT $\rvx$
    \end{algorithmic}
    }
\end{algorithm}
\end{minipage}
\vfill
\newpage

\subsection{Training}
\begin{minipage}{\textwidth}
The following pseudocode shows an example for the training of diffusion-based generative models.

\begin{algorithm}[H]
    \newcommand{\MYOUTPUT}{\item[\textbf{Output:}]}
    \newcommand{\MYINIT}{\item[\textbf{Constants:}]}
    \newcommand{\MYBEGIN}{\item[\textbf{Begin}]}
    \newcommand{\MYEND}{\item[\textbf{End}]}
    \newcommand{\mycom}[1]{\hfill$\sharp$~{#1}}
    \caption{Training of Diffusion models}
    {\small
    \begin{algorithmic}
        \REQUIRE
            \STATE Data size $d > 0$ \quad $\cdots$\textit{Int}
            \STATE Training Data $\mathcal{D} = \{\rvx_0^{(i)}, \textbf{c}^{(i)}\}$, where $\rvx_0^{(i)} \in \mathbb{R}^d$
            \mycom{The conditioning information $\rvc$ is optional.}
            \STATE Neural network model $\scorefn(\rvx_t, t, \rvc)$, parameterized by $\theta$.
            \STATE Neural network optimizer (\textit{Adam}) and its parameters (e.g. learning rate)
            \STATE Total time $T > 0$ \quad $\cdots$\textit{Float}
            \mycom{Not necessarily the same as the one for the synthesis.}
            \STATE Initial and terminal noise levels $\nu_0, \nu_T \in (0, 1)$
            \mycom{Not necessarily the same as the one for the synthesis.}
            \STATE Batch size $b>0$  \quad $\cdots$\textit{Int}
        \MYINIT
            \STATE $\displaystyle
                A \gets \frac{2 \sqrt{\nu_0}}{1 - \sqrt{\nu_0}},
                \quad
                k \gets \frac{1}{T} \left(\log\frac{2\sqrt{\nu_T} }{1 - \sqrt{\nu_T}} - \log A\right)$
                \mycom{\eqref{label:033}, \secref{label:056}}
        \MYBEGIN
            \FOR{sufficiently many times until convergence}
                \STATE{\textbf{[Draw a Batch]}}
                \STATE \textit{batch} $\gets$ [~]
                \mycom{Empty list}
                \FOR{$b$ times (batch size)}
                    \STATE $
                        (\rvx_0, \rvc) \sim \mathcal{D}$
                    \mycom{Draw a data from the set of training data; the conditioning $\rvc$ is optional.}
                    \STATE ~
                    \STATE $\displaystyle\vphantom{\int}
                        t \sim \text{Uniform}(0, T)$
                    \mycom{Draw a time parameter $t$ from the uniform distribution.}
                    \STATE $\displaystyle\vphantom{\int}
                        \lambda \gets \log (1 + A e^{k t})$
                        \mycom{\secref{label:055}}
                    \STATE $\displaystyle\vphantom{\int}
                        \beta \gets \dot\lambda \tanh \frac{\lambda}{2},
                        \nu \gets \tanh^2 \frac{\lambda}{2}$
                        \mycom{\eqref{label:032}, \secref{label:055}}
                    \STATE ~
                    \STATE $
                        \rvw \sim \gN(\bm{0}_d, \mathbf{I}_d)$
                    \mycom{Draw a $d$-dimensional Gaussian noise with unit variance.}
                    \STATE ~
                    \STATE $
                    \displaystyle
                        \textit{SNR}' \gets \frac{(1 - \nu)\beta}{\nu^2}$
                    \mycom{Compute the SNR-based weight. See \eqref{label:075} and \citep{kingma2021variational}}
                    \STATE ~
                    \STATE Append the tuple $(\rvx_0, \rvc, t, \nu, \rvw, \textit{SNR}')$ to \textit{batch}
                \ENDFOR

                \STATE~
                \STATE{\textbf{[Forward Computation]}}
                \STATE $
                    ([\rvx_0], [\rvc], [t], [\nu], [\rvw], [\textit{SNR}']) \gets \textit{batch}$
                \STATE $
                    [\mathbf{s}] \gets \scorefn([\sqrt{1 - \nu}]\odot[\rvx_0] + [\sqrt{\nu}]\odot[\rvw], [t], [\rvc])$
                    \mycom{See also \eqref{label:011}.}
                \STATE~
                \STATE{\textbf{[Back-propagation]}}
                \STATE $
                    \mathcal{L} \gets \E_\text{batch} \left[ [\textit{SNR}'] \times \|[\rvw] - [\mathbf{s}]\|_2^2\right]$
                \mycom{SNR-weighted Loss~\citep{kingma2021variational}. See also \eqref{label:011}.}
                \STATE Compute $\nabla_{\theta} \mathcal{L}$ by the back-propagation.
                \STATE $
                    \theta \gets \textit{Adam}(\bm{\theta}, \nabla_{\theta} \mathcal{L})$
                \mycom{Update the parameters of $\scorefn(\cdot, \cdot, \cdot)$.}
            \ENDFOR
        \MYEND
        \MYOUTPUT The network parameter $\theta$.
    \end{algorithmic}
    }
\end{algorithm}
\end{minipage}
\paragraph{On the SNR weight}
In our case, the SNR is given by $\textit{SNR}(t) = (1-\nu_t) / \nu_t$.
The SNR-based weight is defined by its negative time derivative, so
\begin{equation}
    \textit{SNR}'
    = - \frac{d}{dt} \frac{1 - \nu_t}{\nu_t}
    = \frac{\dot\nu_t}{\nu_t^2}
    = \frac{(1 - \nu_t)\beta_t}{\nu_t^2}.
    \label{label:075}
\end{equation}
See also \eqref{label:051}.

\newpage
\section{Additional Results of Image Synthesis}\label{label:076}
\subsection{Comparison of the Step-size Schedule (Constant vs Exponential)}
    \begin{minipage}{\textwidth}
        \centering
        \begin{table}[H]
            \centering
            \begin{tabular}{m{2.5cm}m{0.7\linewidth}}
                Euler &
                \includegraphics[width=0.95\linewidth]{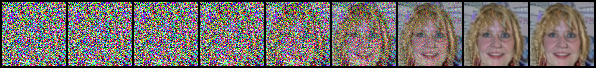}\\
                DDIM &
                \includegraphics[width=0.95\linewidth]{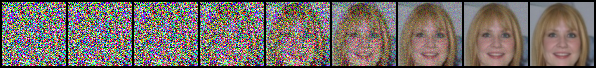}\\
                Heun &
                \includegraphics[width=0.95\linewidth]{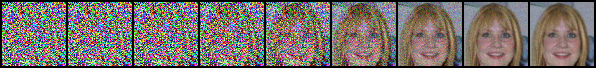}\\
                RK4 &
                \includegraphics[width=0.95\linewidth]{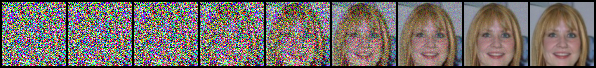}\\
                Taylor (2nd) &
                \includegraphics[width=0.95\linewidth]{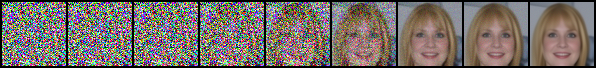}\\
                Taylor (3rd) &
                \includegraphics[width=0.95\linewidth]{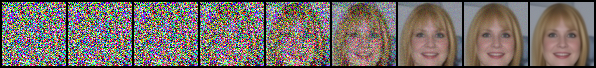}\\
                Euler-Maruyama &
                \includegraphics[width=0.95\linewidth]{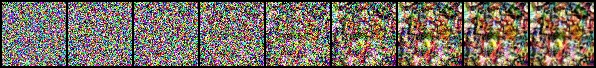}\\
                It\^o-Taylor &
                \includegraphics[width=0.95\linewidth]{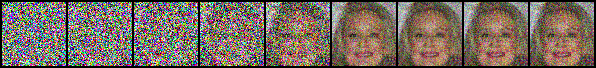}\\
                \\
                Euler &
                \includegraphics[width=0.95\linewidth]{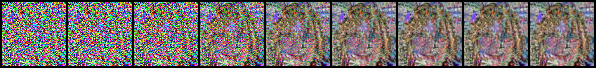}\\
                DDIM &
                \includegraphics[width=0.95\linewidth]{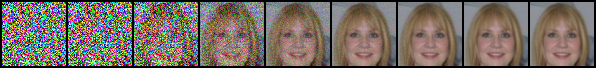}\\
                Heun &
                \includegraphics[width=0.95\linewidth]{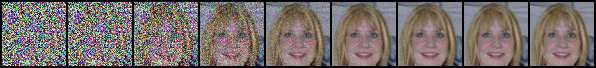}\\
                RK4 &
                \includegraphics[width=0.95\linewidth]{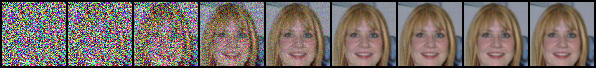}\\
                Taylor (2nd) &
                \includegraphics[width=0.95\linewidth]{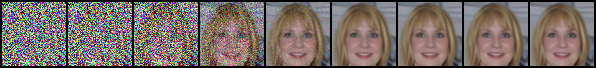}\\
                Taylor (3rd) &
                \includegraphics[width=0.95\linewidth]{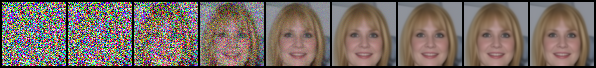}\\
                Euler-Maruyama &
                \includegraphics[width=0.95\linewidth]{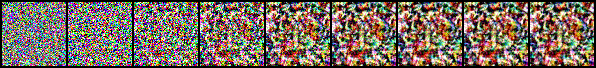}\\
                It\^o-Taylor &
                \includegraphics[width=0.95\linewidth]{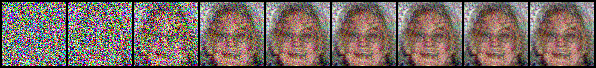}\\
            \end{tabular}
            \captionof{figure}{
                Comparison of the image synthesis process using different step-size schedules.
                above: constant schedule, below: exponential schedule.
                The number of refinement steps is $N = 8$.
                The synthesis noise schedule was condition (ii).
            }
        \end{table}
    \end{minipage}

\vfill
\newpage
\subsection{Other Sampling Examples}
    \begin{minipage}{\textwidth}
        \begin{table}[H]
            \begin{tabular}{m{2.5cm}m{0.7\linewidth}}
                Euler &
                \includegraphics[width=0.95\linewidth]{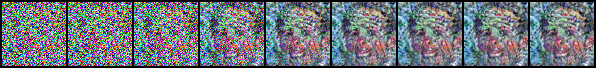}\\
                DDIM &
                \includegraphics[width=0.95\linewidth]{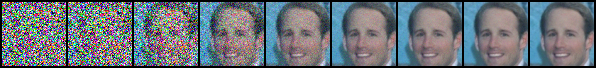}\\
                Heun &
                \includegraphics[width=0.95\linewidth]{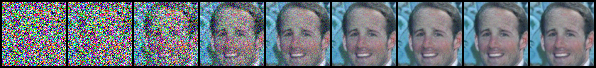}\\
                RK4 &
                \includegraphics[width=0.95\linewidth]{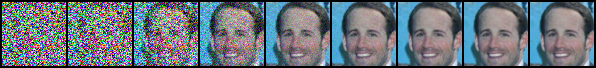}\\
                Taylor (2nd) &
                \includegraphics[width=0.95\linewidth]{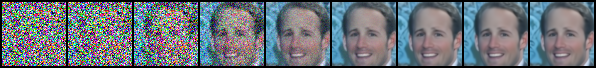}\\
                Taylor (3rd) &
                \includegraphics[width=0.95\linewidth]{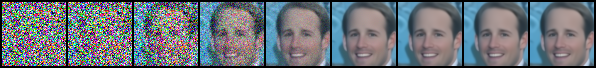}\\
                Euler-Maruyama &
                \includegraphics[width=0.95\linewidth]{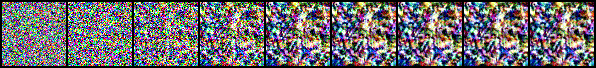}\\
                It\^o-Taylor &
                \includegraphics[width=0.95\linewidth]{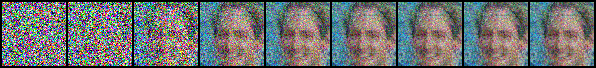}\\
                \\
                Euler &
                \includegraphics[width=0.95\linewidth]{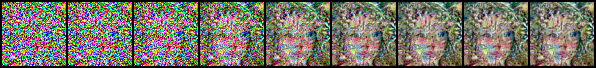}\\
                DDIM &
                \includegraphics[width=0.95\linewidth]{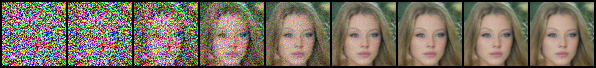}\\
                Heun &
                \includegraphics[width=0.95\linewidth]{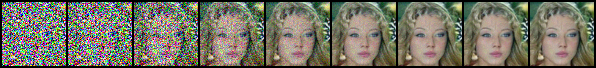}\\
                RK4 &
                \includegraphics[width=0.95\linewidth]{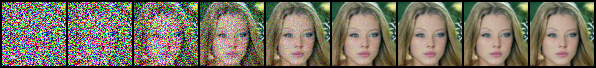}\\
                Taylor (2nd) &
                \includegraphics[width=0.95\linewidth]{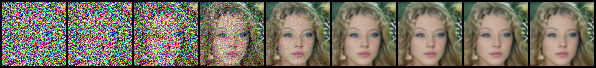}\\
                Taylor (3rd) &
                \includegraphics[width=0.95\linewidth]{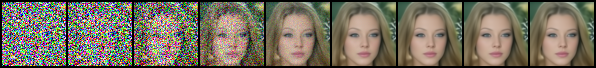}\\
                Euler-Maruyama &
                \includegraphics[width=0.95\linewidth]{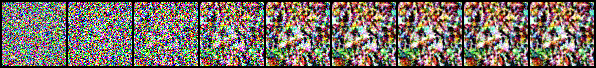}\\
                It\^o-Taylor &
                \includegraphics[width=0.95\linewidth]{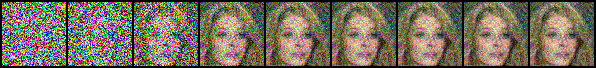}\\
            \end{tabular}
            \caption{
                Comparison of the image synthesis process.
                The dataset is CelebA ($64 \times 64$).
                All the conditions but the sampling algorithm are the same including the random seed.
                The number of refinement steps is $N = 8$.
            }
        \end{table}
    \end{minipage}
    \vfill
    \newpage

\vfill
\newpage
\subsection{Random Samples}
    \begin{figure}[H]
        \centering
            \begin{subfigure}[t]{0.24\textwidth}
                \centering
                \includegraphics[width=0.95\linewidth]{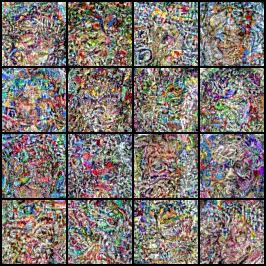}
                \caption{Euler}
            \end{subfigure}
            \begin{subfigure}[t]{0.24\textwidth}
                \centering
                \includegraphics[width=0.95\linewidth]{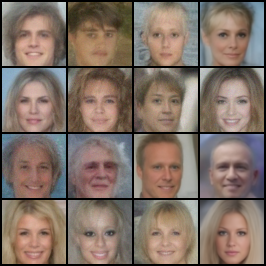}
                \caption{DDIM}
            \end{subfigure}
            \begin{subfigure}[t]{0.24\textwidth}
                \centering
                \includegraphics[width=0.95\linewidth]{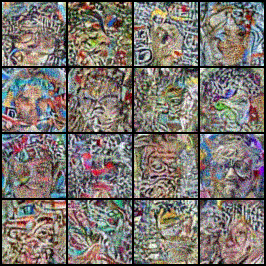}
                \caption{Heun (NFE=2)}
            \end{subfigure}
            \begin{subfigure}[t]{0.24\textwidth}
                \centering
                \includegraphics[width=0.95\linewidth]{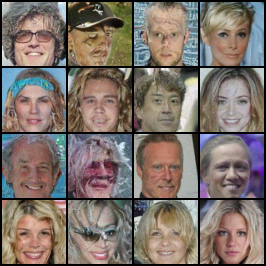}
                \caption{RK4 (NFE=4)}
            \end{subfigure}
            \begin{subfigure}[t]{0.24\textwidth}
                \centering
                \includegraphics[width=0.95\linewidth]{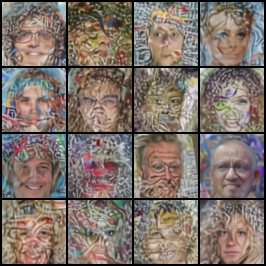}
                \caption{\textbf{Taylor (2nd)}}
            \end{subfigure}
            \begin{subfigure}[t]{0.24\textwidth}
                \centering
                \includegraphics[width=0.95\linewidth]{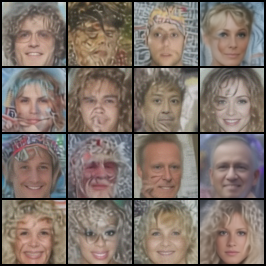}
                \caption{\textbf{Taylor (3rd)}}
            \end{subfigure}
            \begin{subfigure}[t]{0.24\textwidth}
                \centering
                \includegraphics[width=0.95\linewidth]{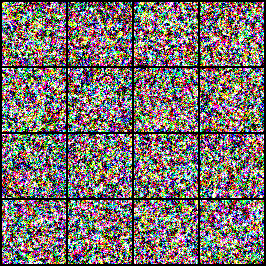}
                \caption{Euler-Maruyama}
            \end{subfigure}
            \begin{subfigure}[t]{0.24\textwidth}
                \centering
                \includegraphics[width=0.95\linewidth]{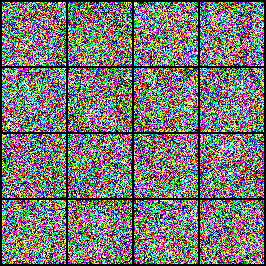}
                \caption{\textbf{It\^o-Taylor}}
            \end{subfigure}
        \caption{
            CelebA ($64\times 64$) synthesis samples. \underline{$N=4$}. Noise schedule was (ii).
        }
        \vspace{2cm}
        \centering
            \begin{subfigure}[t]{0.24\textwidth}
                \centering
                \includegraphics[width=0.95\linewidth]{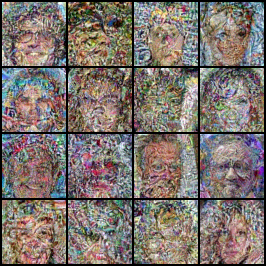}
                \caption{Euler}
            \end{subfigure}
            \begin{subfigure}[t]{0.24\textwidth}
                \centering
                \includegraphics[width=0.95\linewidth]{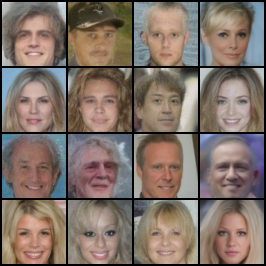}
                \caption{DDIM}
            \end{subfigure}
            \begin{subfigure}[t]{0.24\textwidth}
                \centering
                \includegraphics[width=0.95\linewidth]{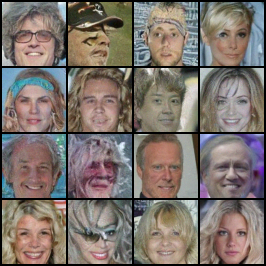}
                \caption{Heun (NFE=2)}
            \end{subfigure}
            \begin{subfigure}[t]{0.24\textwidth}
                \centering
                \includegraphics[width=0.95\linewidth]{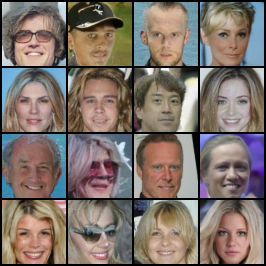}
                \caption{RK4 (NFE=4)}
            \end{subfigure}
            \begin{subfigure}[t]{0.24\textwidth}
                \centering
                \includegraphics[width=0.95\linewidth]{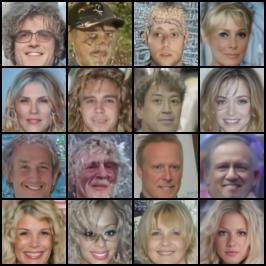}
                \caption{\textbf{Taylor (2nd)}}
            \end{subfigure}
            \begin{subfigure}[t]{0.24\textwidth}
                \centering
                \includegraphics[width=0.95\linewidth]{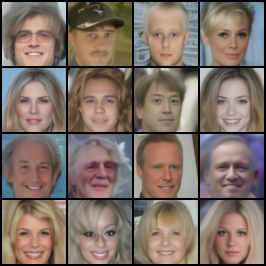}
                \caption{\textbf{Taylor (3rd)}}
            \end{subfigure}
            \begin{subfigure}[t]{0.24\textwidth}
                \centering
                \includegraphics[width=0.95\linewidth]{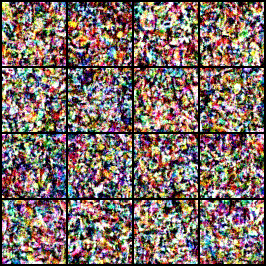}
                \caption{Euler-Maruyama}
            \end{subfigure}
            \begin{subfigure}[t]{0.24\textwidth}
                \centering
                \includegraphics[width=0.95\linewidth]{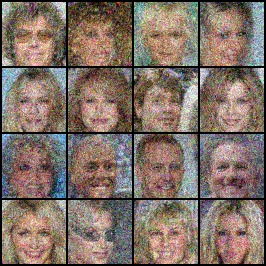}
                \caption{\textbf{It\^o-Taylor}}
            \end{subfigure}
        \caption{
            CelebA ($64\times 64$) synthesis samples. \underline{$N=8$}. Noise schedule was (ii).
        }
    \end{figure}
    \vfill
    \newpage

    \begin{figure}[H]
        \centering
            \begin{subfigure}[t]{0.24\textwidth}
                \centering
                \includegraphics[width=0.95\linewidth]{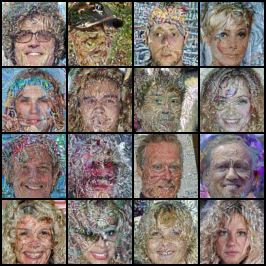}
                \caption{Euler}
            \end{subfigure}
            \begin{subfigure}[t]{0.24\textwidth}
                \centering
                \includegraphics[width=0.95\linewidth]{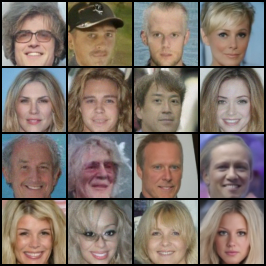}
                \caption{DDIM}
            \end{subfigure}
            \begin{subfigure}[t]{0.24\textwidth}
                \centering
                \includegraphics[width=0.95\linewidth]{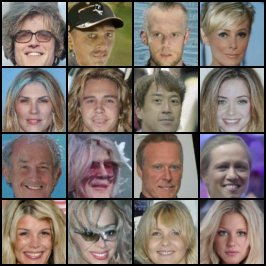}
                \caption{Heun (NFE=2)}
            \end{subfigure}
            \begin{subfigure}[t]{0.24\textwidth}
                \centering
                \includegraphics[width=0.95\linewidth]{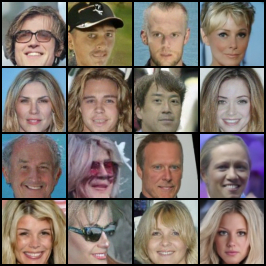}
                \caption{RK4 (NFE=4)}
            \end{subfigure}
            \begin{subfigure}[t]{0.24\textwidth}
                \centering
                \includegraphics[width=0.95\linewidth]{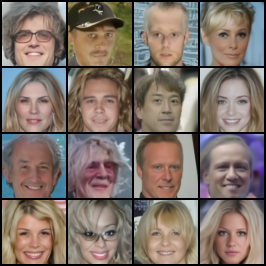}
                \caption{\textbf{Taylor (2nd)}}
            \end{subfigure}
            \begin{subfigure}[t]{0.24\textwidth}
                \centering
                \includegraphics[width=0.95\linewidth]{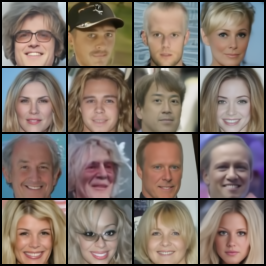}
                \caption{\textbf{Taylor (3rd)}}
            \end{subfigure}
            \begin{subfigure}[t]{0.24\textwidth}
                \centering
                \includegraphics[width=0.95\linewidth]{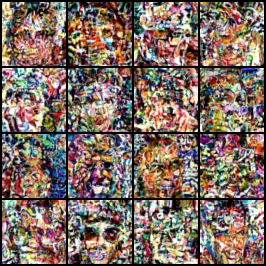}
                \caption{Euler-Maruyama}
            \end{subfigure}
            \begin{subfigure}[t]{0.24\textwidth}
                \centering
                \includegraphics[width=0.95\linewidth]{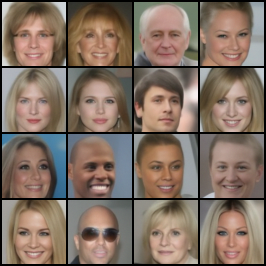}
                \caption{\textbf{It\^o-Taylor}}
            \end{subfigure}
        \caption{
            CelebA ($64\times 64$) synthesis samples. \underline{$N=16$}. Noise schedule was (ii).
        }
        \vspace{2cm}
        \centering
            \begin{subfigure}[t]{0.24\textwidth}
                \centering
                \includegraphics[width=0.95\linewidth]{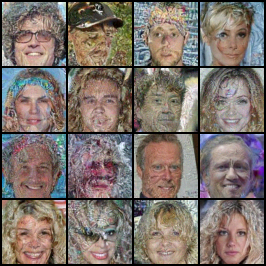}
                \caption{Euler}
            \end{subfigure}
            \begin{subfigure}[t]{0.24\textwidth}
                \centering
                \includegraphics[width=0.95\linewidth]{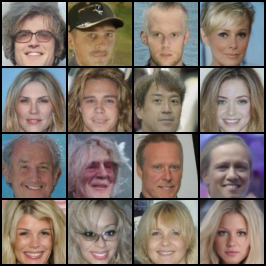}
                \caption{DDIM}
            \end{subfigure}
            \begin{subfigure}[t]{0.24\textwidth}
                \centering
                \includegraphics[width=0.95\linewidth]{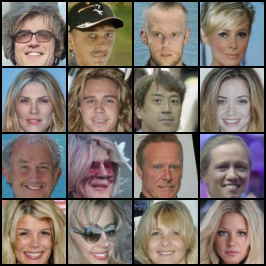}
                \caption{Heun (NFE=2)}
            \end{subfigure}
            \begin{subfigure}[t]{0.24\textwidth}
                \centering
                \includegraphics[width=0.95\linewidth]{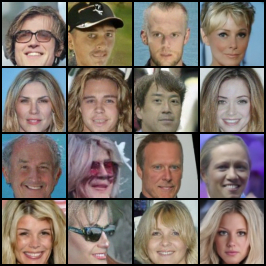}
                \caption{RK4 (NFE=4)}
            \end{subfigure}
            \begin{subfigure}[t]{0.24\textwidth}
                \centering
                \includegraphics[width=0.95\linewidth]{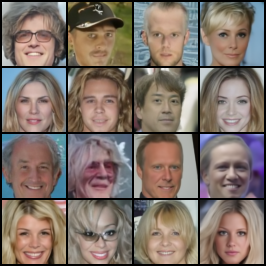}
                \caption{\textbf{Taylor (2nd)}}
            \end{subfigure}
            \begin{subfigure}[t]{0.24\textwidth}
                \centering
                \includegraphics[width=0.95\linewidth]{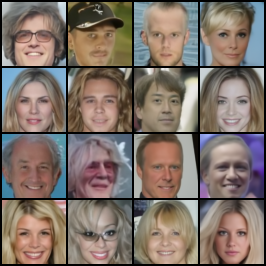}
                \caption{\textbf{Taylor (3rd)}}
            \end{subfigure}
            \begin{subfigure}[t]{0.24\textwidth}
                \centering
                \includegraphics[width=0.95\linewidth]{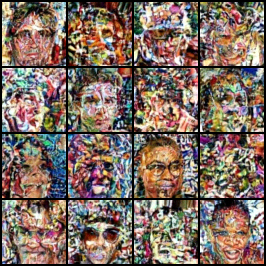}
                \caption{Euler-Maruyama}
            \end{subfigure}
            \begin{subfigure}[t]{0.24\textwidth}
                \centering
                \includegraphics[width=0.95\linewidth]{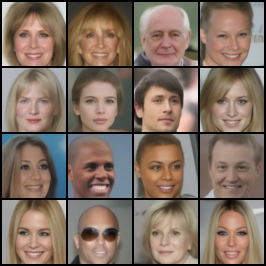}
                \caption{\textbf{It\^o-Taylor}}
            \end{subfigure}
        \caption{
            CelebA ($64\times 64$) synthesis samples. \underline{$N=20$}. Noise schedule was (ii).
        }
    \end{figure}
    \vfill
    \newpage

    \begin{figure}[H]
        \centering
            \begin{subfigure}[t]{0.24\textwidth}
                \centering
                \includegraphics[width=0.95\linewidth]{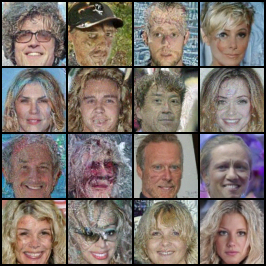}
                \caption{Euler}
            \end{subfigure}
            \begin{subfigure}[t]{0.24\textwidth}
                \centering
                \includegraphics[width=0.95\linewidth]{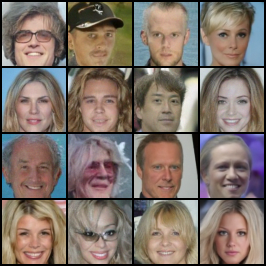}
                \caption{DDIM}
            \end{subfigure}
            \begin{subfigure}[t]{0.24\textwidth}
                \centering
                \includegraphics[width=0.95\linewidth]{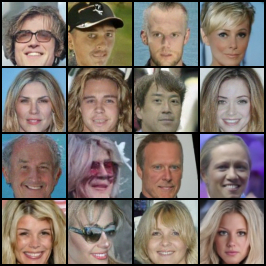}
                \caption{Heun (NFE=2)}
            \end{subfigure}
            \begin{subfigure}[t]{0.24\textwidth}
                \centering
                \includegraphics[width=0.95\linewidth]{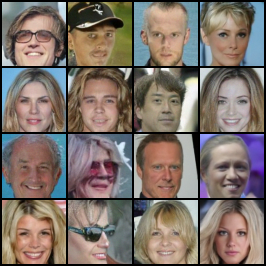}
                \caption{RK4 (NFE=4)}
            \end{subfigure}
            \begin{subfigure}[t]{0.24\textwidth}
                \centering
                \includegraphics[width=0.95\linewidth]{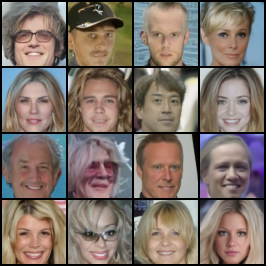}
                \caption{\textbf{Taylor (2nd)}}
            \end{subfigure}
            \begin{subfigure}[t]{0.24\textwidth}
                \centering
                \includegraphics[width=0.95\linewidth]{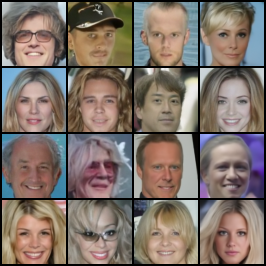}
                \caption{\textbf{Taylor (3rd)}}
            \end{subfigure}
            \begin{subfigure}[t]{0.24\textwidth}
                \centering
                \includegraphics[width=0.95\linewidth]{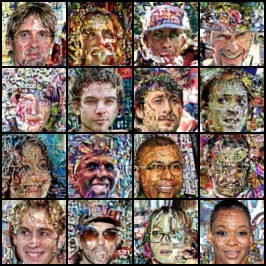}
                \caption{Euler-Maruyama}
            \end{subfigure}
            \begin{subfigure}[t]{0.24\textwidth}
                \centering
                \includegraphics[width=0.95\linewidth]{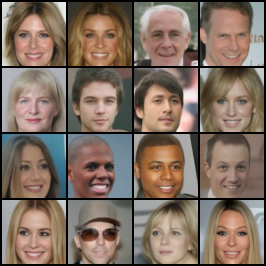}
                \caption{\textbf{It\^o-Taylor}}
            \end{subfigure}
        \caption{
            CelebA ($64\times 64$) synthesis samples. \underline{$N=30$}. Noise schedule was (ii).
        }
        \vspace{2cm}
        \centering
            \begin{subfigure}[t]{0.24\textwidth}
                \centering
                \includegraphics[width=0.95\linewidth]{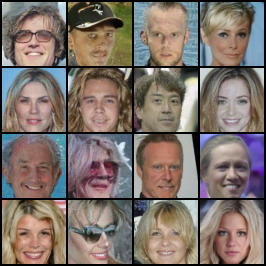}
                \caption{Euler}
            \end{subfigure}
            \begin{subfigure}[t]{0.24\textwidth}
                \centering
                \includegraphics[width=0.95\linewidth]{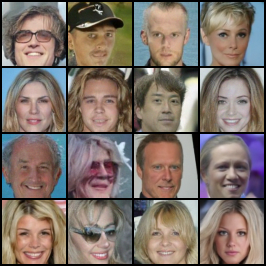}
                \caption{DDIM}
            \end{subfigure}
            \begin{subfigure}[t]{0.24\textwidth}
                \centering
                \includegraphics[width=0.95\linewidth]{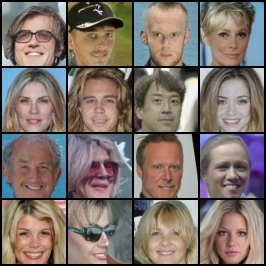}
                \caption{Heun (NFE=2)}
            \end{subfigure}
            \begin{subfigure}[t]{0.24\textwidth}
                \centering
                \includegraphics[width=0.95\linewidth]{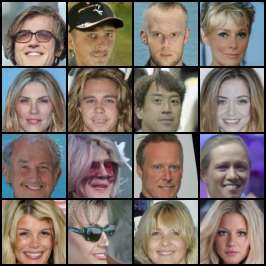}
                \caption{RK4 (NFE=4)}
            \end{subfigure}
            \begin{subfigure}[t]{0.24\textwidth}
                \centering
                \includegraphics[width=0.95\linewidth]{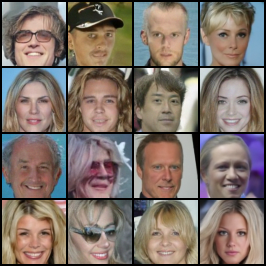}
                \caption{\textbf{Taylor (2nd)}}
            \end{subfigure}
            \begin{subfigure}[t]{0.24\textwidth}
                \centering
                \includegraphics[width=0.95\linewidth]{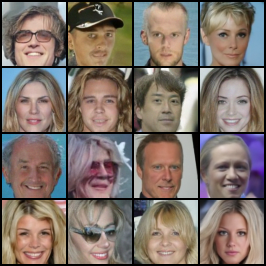}
                \caption{\textbf{Taylor (3rd)}}
            \end{subfigure}
            \begin{subfigure}[t]{0.24\textwidth}
                \centering
                \includegraphics[width=0.95\linewidth]{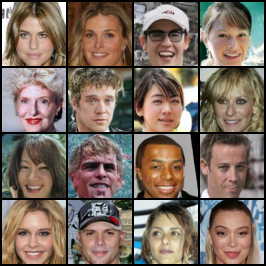}
                \caption{Euler-Maruyama}
            \end{subfigure}
            \begin{subfigure}[t]{0.24\textwidth}
                \centering
                \includegraphics[width=0.95\linewidth]{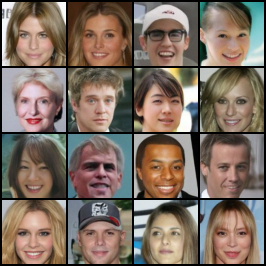}
                \caption{\textbf{It\^o-Taylor}}
            \end{subfigure}
        \caption{
            CelebA ($64\times 64$) synthesis samples. \underline{$N=100$}. Noise schedule was (ii).
        }\label{label:077}
    \end{figure}
    \vfill
    \newpage

    \begin{figure}[H]
        \centering
            \begin{subfigure}[t]{0.24\textwidth}
                \centering
                \includegraphics[width=0.95\linewidth]{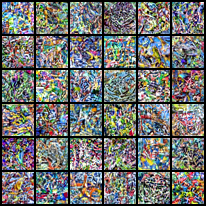}
                \caption{Euler}
            \end{subfigure}
            \begin{subfigure}[t]{0.24\textwidth}
                \centering
                \includegraphics[width=0.95\linewidth]{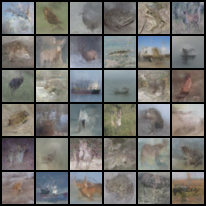}
                \caption{DDIM}
            \end{subfigure}
            \begin{subfigure}[t]{0.24\textwidth}
                \centering
                \includegraphics[width=0.95\linewidth]{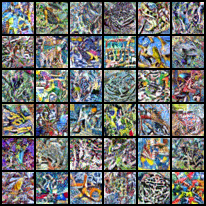}
                \caption{Heun (NFE=2)}
            \end{subfigure}
            \begin{subfigure}[t]{0.24\textwidth}
                \centering
                \includegraphics[width=0.95\linewidth]{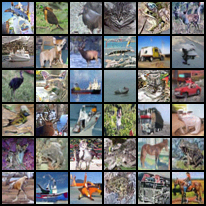}
                \caption{RK4 (NFE=4)}
            \end{subfigure}
            \begin{subfigure}[t]{0.24\textwidth}
                \centering
                \includegraphics[width=0.95\linewidth]{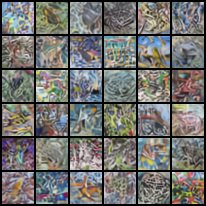}
                \caption{\textbf{Taylor (2nd)}}
            \end{subfigure}
            \begin{subfigure}[t]{0.24\textwidth}
                \centering
                \includegraphics[width=0.95\linewidth]{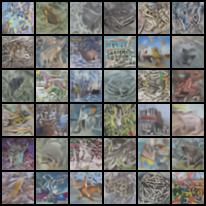}
                \caption{\textbf{Taylor (3rd)}}
            \end{subfigure}
            \begin{subfigure}[t]{0.24\textwidth}
                \centering
                \includegraphics[width=0.95\linewidth]{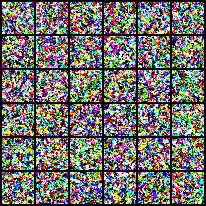}
                \caption{Euler-Maruyama}
            \end{subfigure}
            \begin{subfigure}[t]{0.24\textwidth}
                \centering
                \includegraphics[width=0.95\linewidth]{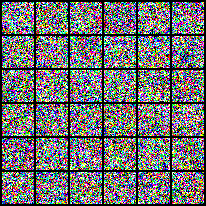}
                \caption{\textbf{It\^o-Taylor}}
            \end{subfigure}
        \caption{
            CIFAR-10($32\times 32$) synthesis samples. \underline{$N=4$}. Noise schedule was (ii).
        }
        \vspace{2cm}
        \centering
            \begin{subfigure}[t]{0.24\textwidth}
                \centering
                \includegraphics[width=0.95\linewidth]{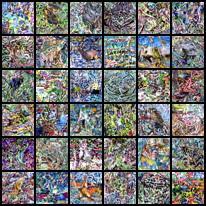}
                \caption{Euler}
            \end{subfigure}
            \begin{subfigure}[t]{0.24\textwidth}
                \centering
                \includegraphics[width=0.95\linewidth]{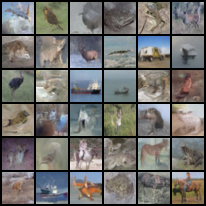}
                \caption{DDIM}
            \end{subfigure}
            \begin{subfigure}[t]{0.24\textwidth}
                \centering
                \includegraphics[width=0.95\linewidth]{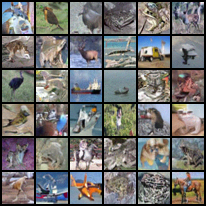}
                \caption{Heun (NFE=2)}
            \end{subfigure}
            \begin{subfigure}[t]{0.24\textwidth}
                \centering
                \includegraphics[width=0.95\linewidth]{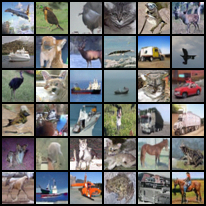}
                \caption{RK4 (NFE=4)}
            \end{subfigure}
            \begin{subfigure}[t]{0.24\textwidth}
                \centering
                \includegraphics[width=0.95\linewidth]{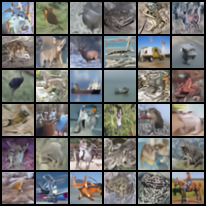}
                \caption{\textbf{Taylor (2nd)}}
            \end{subfigure}
            \begin{subfigure}[t]{0.24\textwidth}
                \centering
                \includegraphics[width=0.95\linewidth]{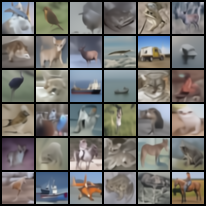}
                \caption{\textbf{Taylor (3rd)}}
            \end{subfigure}
            \begin{subfigure}[t]{0.24\textwidth}
                \centering
                \includegraphics[width=0.95\linewidth]{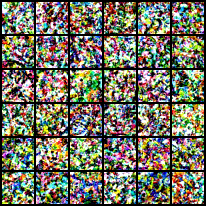}
                \caption{Euler-Maruyama}
            \end{subfigure}
            \begin{subfigure}[t]{0.24\textwidth}
                \centering
                \includegraphics[width=0.95\linewidth]{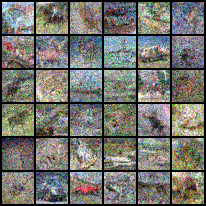}
                \caption{\textbf{It\^o-Taylor}}
            \end{subfigure}
        \caption{
            CIFAR-10 ($32\times 32$) synthesis samples. \underline{$N=8$}. Noise schedule was (ii).
        }
    \end{figure}
    \vfill
    \newpage

    \begin{figure}[H]
        \centering
            \begin{subfigure}[t]{0.24\textwidth}
                \centering
                \includegraphics[width=0.95\linewidth]{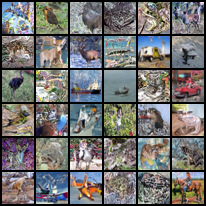}
                \caption{Euler}
            \end{subfigure}
            \begin{subfigure}[t]{0.24\textwidth}
                \centering
                \includegraphics[width=0.95\linewidth]{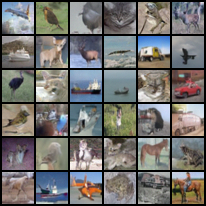}
                \caption{DDIM}
            \end{subfigure}
            \begin{subfigure}[t]{0.24\textwidth}
                \centering
                \includegraphics[width=0.95\linewidth]{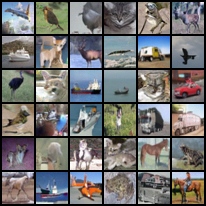}
                \caption{Heun (NFE=2)}
            \end{subfigure}
            \begin{subfigure}[t]{0.24\textwidth}
                \centering
                \includegraphics[width=0.95\linewidth]{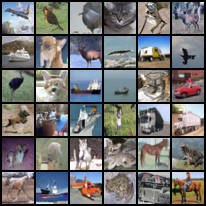}
                \caption{RK4 (NFE=4)}
            \end{subfigure}
            \begin{subfigure}[t]{0.24\textwidth}
                \centering
                \includegraphics[width=0.95\linewidth]{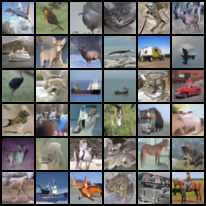}
                \caption{\textbf{Taylor (2nd)}}
            \end{subfigure}
            \begin{subfigure}[t]{0.24\textwidth}
                \centering
                \includegraphics[width=0.95\linewidth]{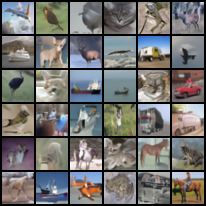}
                \caption{\textbf{Taylor (3rd)}}
            \end{subfigure}
            \begin{subfigure}[t]{0.24\textwidth}
                \centering
                \includegraphics[width=0.95\linewidth]{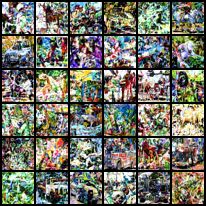}
                \caption{Euler-Maruyama}
            \end{subfigure}
            \begin{subfigure}[t]{0.24\textwidth}
                \centering
                \includegraphics[width=0.95\linewidth]{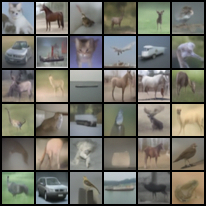}
                \caption{\textbf{It\^o-Taylor}}
            \end{subfigure}
        \caption{
            CIFAR-10 ($32\times 32$) synthesis samples. \underline{$N=20$}. Noise schedule was (ii).
        }
        \vspace{2cm}
        \centering
            \begin{subfigure}[t]{0.24\textwidth}
                \centering
                \includegraphics[width=0.95\linewidth]{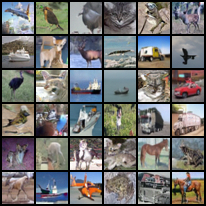}
                \caption{Euler}
            \end{subfigure}
            \begin{subfigure}[t]{0.24\textwidth}
                \centering
                \includegraphics[width=0.95\linewidth]{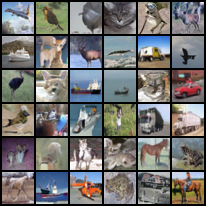}
                \caption{DDIM}
            \end{subfigure}
            \begin{subfigure}[t]{0.24\textwidth}
                \centering
                \includegraphics[width=0.95\linewidth]{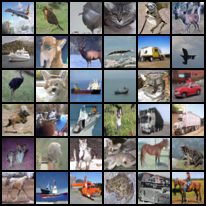}
                \caption{Heun (NFE=2)}
            \end{subfigure}
            \begin{subfigure}[t]{0.24\textwidth}
                \centering
                \includegraphics[width=0.95\linewidth]{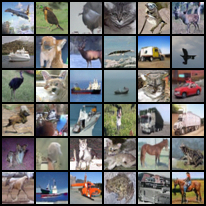}
                \caption{RK4 (NFE=4)}
            \end{subfigure}
            \begin{subfigure}[t]{0.24\textwidth}
                \centering
                \includegraphics[width=0.95\linewidth]{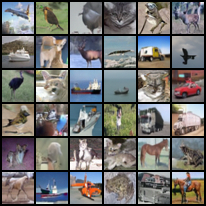}
                \caption{\textbf{Taylor (2nd)}}
            \end{subfigure}
            \begin{subfigure}[t]{0.24\textwidth}
                \centering
                \includegraphics[width=0.95\linewidth]{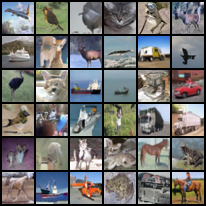}
                \caption{\textbf{Taylor (3rd)}}
            \end{subfigure}
            \begin{subfigure}[t]{0.24\textwidth}
                \centering
                \includegraphics[width=0.95\linewidth]{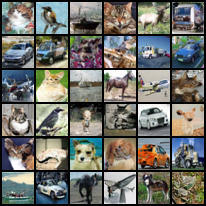}
                \caption{Euler-Maruyama}
            \end{subfigure}
            \begin{subfigure}[t]{0.24\textwidth}
                \centering
                \includegraphics[width=0.95\linewidth]{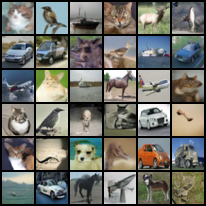}
                \caption{\textbf{It\^o-Taylor}}
            \end{subfigure}
        \caption{
            CIFAR-10 ($32\times 32$) synthesis samples. \underline{$N=100$}. Noise schedule was (ii).
        }
    \end{figure}
    \vfill
    \newpage

\end{document}